\documentclass[10pt,twocolumn]{article} 
\usepackage{simpleConference}
\usepackage{times}
\usepackage{graphicx}
\usepackage{amssymb}
\usepackage{epstopdf}
\usepackage{url}

\usepackage{simpleConference}
\usepackage{times}
\usepackage{graphicx}
\usepackage{amssymb}
\usepackage{url}
\usepackage{array}
\usepackage{longtable}
\usepackage{booktabs}
\usepackage{float} 
\usepackage{amsmath}
\usepackage{makecell}
\usepackage{multirow}
\usepackage{colortbl}
\usepackage{rotating}
\definecolor{xred}{RGB}{200,0,0} 
\definecolor{xgreen}{RGB}{0,200,0} 
\usepackage{supertabular}
\usepackage[dvipsnames]{xcolor}
\usepackage{tikz}
\usetikzlibrary{positioning, arrows.meta}
\definecolor{newcolor}{rgb}{.8,.349,.1}

\usepackage{float}
\usepackage[authoryear]{natbib}
\usepackage{authblk}

\usepackage[colorlinks=true]{hyperref}

\begin{document}

\title{From Pixels to Polygons: A Survey of Deep Learning Approaches for Medical Image-to-Mesh Reconstruction}

\author{
Fengming Lin$^{1,2}$, 
Arezoo Zakeri$^{1,3}$, 
Yidan Xue$^{1,3}$, 
Michael MacRaild$^{1,2}$, 
Haoran Dou$^{1,2}$, 
Zherui Zhou$^{1,2}$, 
Ziwei Zou$^{1}$, 
Ali Sarrami-Foroushani$^{1,3}$, 
Jinming Duan$^{1,3}$, 
Alejandro F. Frangi$^{1,2,3,4,5,6}$\thanks{Corresponding author: alejandro.frangi@manchester.ac.uk}
}

\affil[1]{Centre for Computational Imaging and Modelling in Medicine (CIMIM), The Christabel Pankhurst Institute, The University of Manchester, Manchester, UK}
\affil[2]{Department of Computer Science, School of Engineering, University of Manchester, Manchester, UK}
\affil[3]{Division of Informatics, Imaging and Data Sciences, University of Manchester, Manchester, UK}
\affil[4]{NIHR Manchester Biomedical Research Centre, Manchester Academic Health Sciences Centre, University of Manchester, Manchester, UK}
\affil[5]{Department of Cardiovascular Sciences, KU Leuven, Leuven, Belgium}
\affil[6]{Department of Electrical Engineering (ESAT), KU Leuven, Leuven, Belgium}

\maketitle
\thispagestyle{empty}

\begin{abstract}
Deep learning-based medical image-to-mesh reconstruction has rapidly evolved, enabling the transformation of medical imaging data into three-dimensional mesh models that are critical in computational medicine and in silico trials for advancing our understanding of disease mechanisms, and diagnostic and therapeutic techniques in modern medicine. This survey systematically categorizes existing approaches into four main categories: template models, statistical models, generative models, and implicit models. Each category is analysed in detail, examining their methodological foundations, strengths, limitations, and applicability to different anatomical structures and imaging modalities. We provide an extensive evaluation of these methods across various anatomical applications, from cardiac imaging to neurological studies, supported by quantitative comparisons using standard metrics. Additionally, we compile and analyze major public datasets available for medical mesh reconstruction tasks and discuss commonly used evaluation metrics and loss functions. The survey identifies current challenges in the field, including requirements for topological correctness, geometric accuracy, and multi-modality integration. Finally, we present promising future research directions in this domain. This systematic review aims to serve as a comprehensive reference for researchers and practitioners in medical image analysis and computational medicine.
\end{abstract}

\section{Introduction}
\label{sec:introduction}

\begin{figure*}[!ht]
    \centering
    \includegraphics[width=\linewidth]{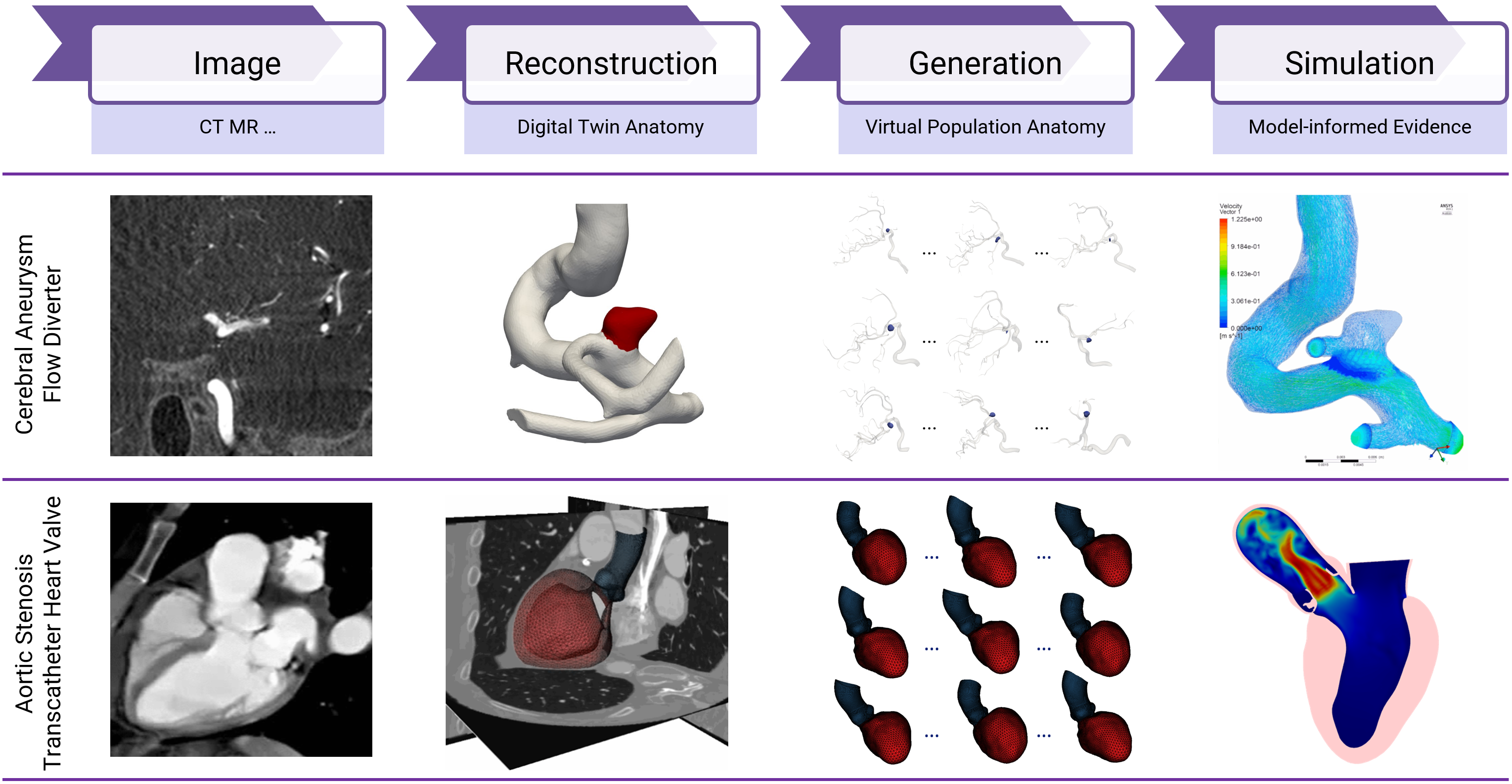}
    \caption{Pipeline of in-silico trials, consisting of four key stages: image acquisition, reconstruction, generation, and simulation. Medical imaging modalities such as CT and MR scans provide anatomical data, which are then converted into digital twin anatomy in the reconstruction stage. The generation phase creates virtual population anatomy to account for variability, while the simulation stage performs computational analyses to derive model-informed evidence. Two examples are shown: (1) Cerebral aneurysm with flow diverter (\cite{sarrami2021silico, macraild2024off, lin2023high}), where vascular imaging is used to reconstruct the aneurysm, generate a virtual population, and simulate blood flow changes after device implantation; and (2) Aortic Stenosis with transcatheter heart valve  (\cite{pak2023patient, ozturk2025ai}), where CT images are used to reconstruct the heart, generate a virtual cohort, and simulate hemodynamic effects of THV implantation.}
    \label{fig:IST} 
\end{figure*}

\begin{figure}[!ht]
    \centering
    \includegraphics[width=1.05\linewidth]{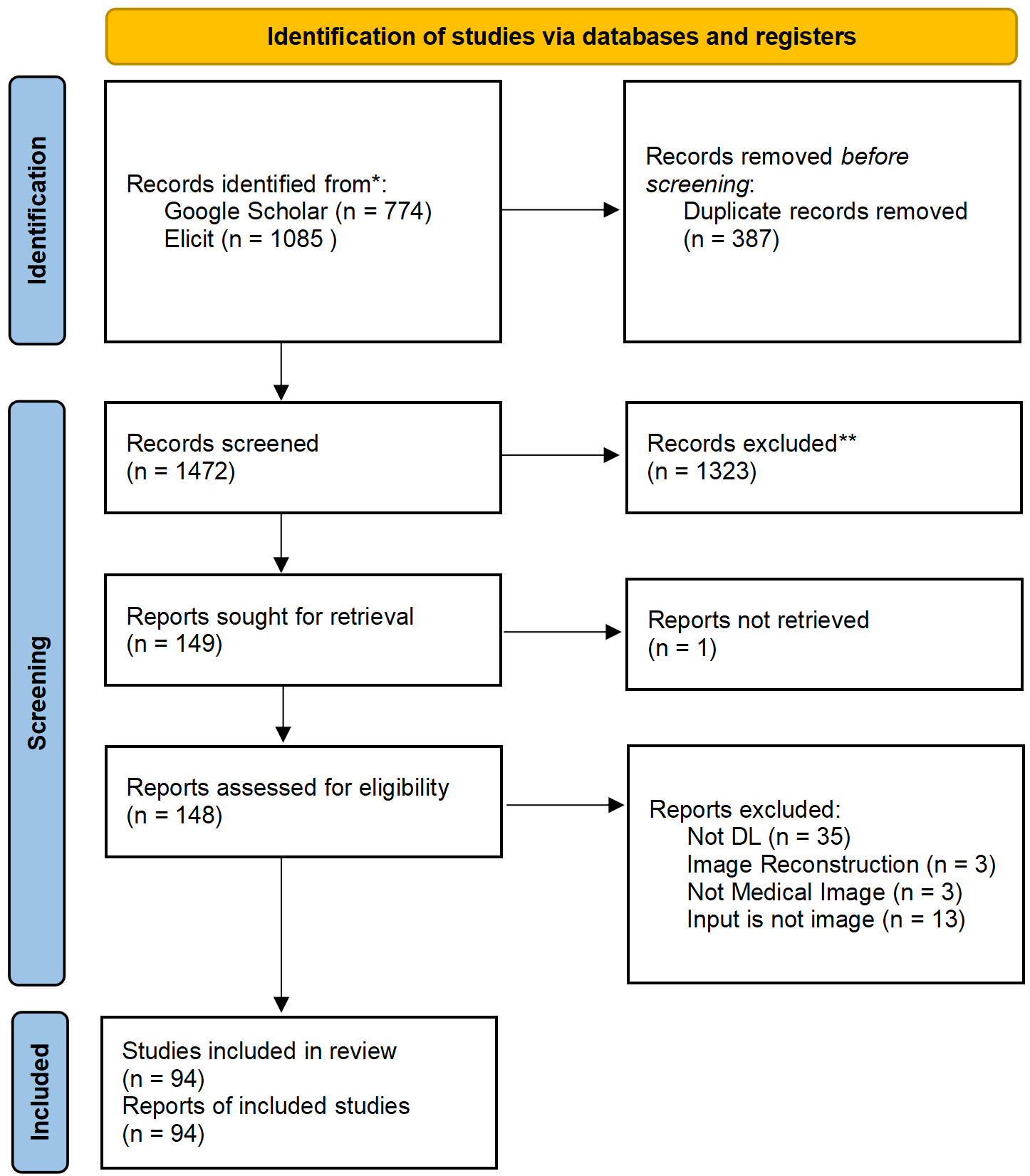}
    \caption{PRISMA flowchart summarizing the systematic review process. (PRISMA registration number: CRD420250655291)}
    \label{fig:prisma}
\end{figure}

\subsection{Background}
In the fields of computational medicine and in silico trials, image-to-mesh reconstruction is critical for advancing our understanding of disease mechanisms, and diagnostic and therapeutic techniques in modern medicine.for advancing diagnostic and therapeutic techniques. From anatomical visualization to computational simulations \cite{sarrami2021silico}, the ability to accurately reconstruct three-dimensional (3D) representations of organs and tissues has transformed modern medicine. 
A mesh (\cite{lorensen1998marching}) is a collection of vertices, edges, and faces that define the shape of a 3D object, commonly used to represent anatomical structures in computational modelling.
With the advent of deep learning, traditional methods relying on manual segmentation are increasingly complemented or replaced by data-driven approaches. 
Deep learning (\cite{shen2017deep}) has enabled the direct mapping of medical images
to anatomical representations like meshes, bypassing intermediate steps and offering enhanced precision.

\subsection{Motivation}
The concept of in-silico trials, shown in Fig.~\ref{fig:IST}, has emerged as a groundbreaking paradigm in computational medicine. In-silico trials use computer simulations to model the behaviour and response of biological systems under various conditions, providing a safe, ethical, and cost-effective alternative to traditional in-vivo or in-vitro experiments (\cite{frangi2001quantitative, walcott2016flow}). Central to this approach is the accurate generation of 3D anatomical meshes from medical images, which serve as the foundation for simulations such as hemodynamic studies, structural analysis, and virtual device testing.

Image-to-mesh reconstruction is pivotal for in-silico trials as it bridges the gap between imaging modalities and computational analysis. Through the reconstruction of patient-specific meshes, this technology helps overcome challenges related to variability in anatomy, complex geometries, and the need for simulation-ready models. Consequently, advancements in image-to-mesh techniques play a key role in enabling personalized medicine and precision healthcare.

Traditional image-to-mesh reconstruction methods typically follow a two-step approach: segmentation first, followed by post-processing or registration. However, these segmentation-based methods heavily depend on the accuracy of segmentation, which is inherently limited by image resolution and segmentation performance. Errors in segmentation can significantly impact downstream tasks, such as simulations, leading to reduced accuracy and reliability. To overcome these limitations, the new generation of image-to-mesh reconstruction methods adopts an end-to-end approach, directly generating meshes from images without relying on an intermediate segmentation step.

\subsection{Scope}

This paper focuses on deep learning-based methods for end-to-end image-to-mesh reconstruction, offering a structured taxonomy and in-depth analysis of methodologies, loss functions, evaluation metrics, and their applications in different anatomies. 

We categorize existing methods into four primary groups: template models, statistical shape models, generative models, and implicit models. These methodologies further branch into twelve subcategories, capturing variations in processing pipelines and feature representations (see Fig.~\ref{fig:taxonomy}). These approaches take various medical imaging modalities as input, including computed tomography (CT), magnetic resonance (MR), and ultrasound (US), to generate anatomical meshes tailored for advancing diagnostic and therapeutic techniques.

In addition to methodological classification, we systematically analyze loss functions and evaluation metrics from two perspectives. First, we group them based on their design principles into shape similarity, regularization, and function-driven constraints. Shape similarity measures include distance-based metrics and consistency checks, while regularization encompasses smoothness, topological constraints, and curvature preservation. Function-driven evaluation further considers clinical and simulation-specific constraints. Second, we examine these metrics concerning their computational representations, such as vertex-based, voxel-based, normal vector-based, and implicit function-based formulations.

Furthermore, we conduct a meta-analysis, applying these evaluation metrics to different anatomical structures and comparing the performance of various methodologies. This comparative assessment provides insights into the effectiveness of different techniques across diverse anatomical regions.

Lastly, we discuss the strengths and limitations of each approach, offering guidance for selecting the most suitable methodology for specific medical applications. By identifying current challenges and opportunities, we aim to assist researchers in refining and advancing image-to-mesh reconstruction frameworks.

\subsection{Contributions}
This work makes the following contributions:
\begin{itemize}

    \item We present a systematic review of end-to-end medical image-to-mesh reconstruction, offering a detailed analysis of its methodologies and performance. We establish a systematic taxonomy, categorizing these methods into four main types: statistical shape models, template models, generative models, and implicit models. Additionally, we further divide these four categories into twelve subcategories based on their processing pipelines and feature representations.
    
    \item A summary and classification of loss functions and evaluation metrics used in these methods, along with a meta-analysis of experimental results on different anatomies to assess the performance of different method categories.
    \item We systematically review and curate publicly available datasets in this field, covering multiple anatomical structures and imaging modalities. This dataset collection provides a structured resource for researchers working on image-to-mesh reconstruction and in-silico trial simulation.
    \item Identification of open challenges and future research directions, emphasizing the need for higher-fidelity training methods, topology constraints and multi-model fusion.

\end{itemize}

\subsection{PRISMA Framework}
The systematic review conducted in this paper adheres to the PRISMA (Preferred Reporting Items for Systematic Reviews and Meta-Analyses) guidelines \cite{page2021prisma}. A detailed flowchart summarizing the literature selection process is provided in Fig.~\ref{fig:prisma}, ensuring transparency and reproducibility in the research methodology.

\section{Methods Taxonomy}

\begin{figure*}[!h]
    \centering
    \includegraphics[width=0.95\linewidth]{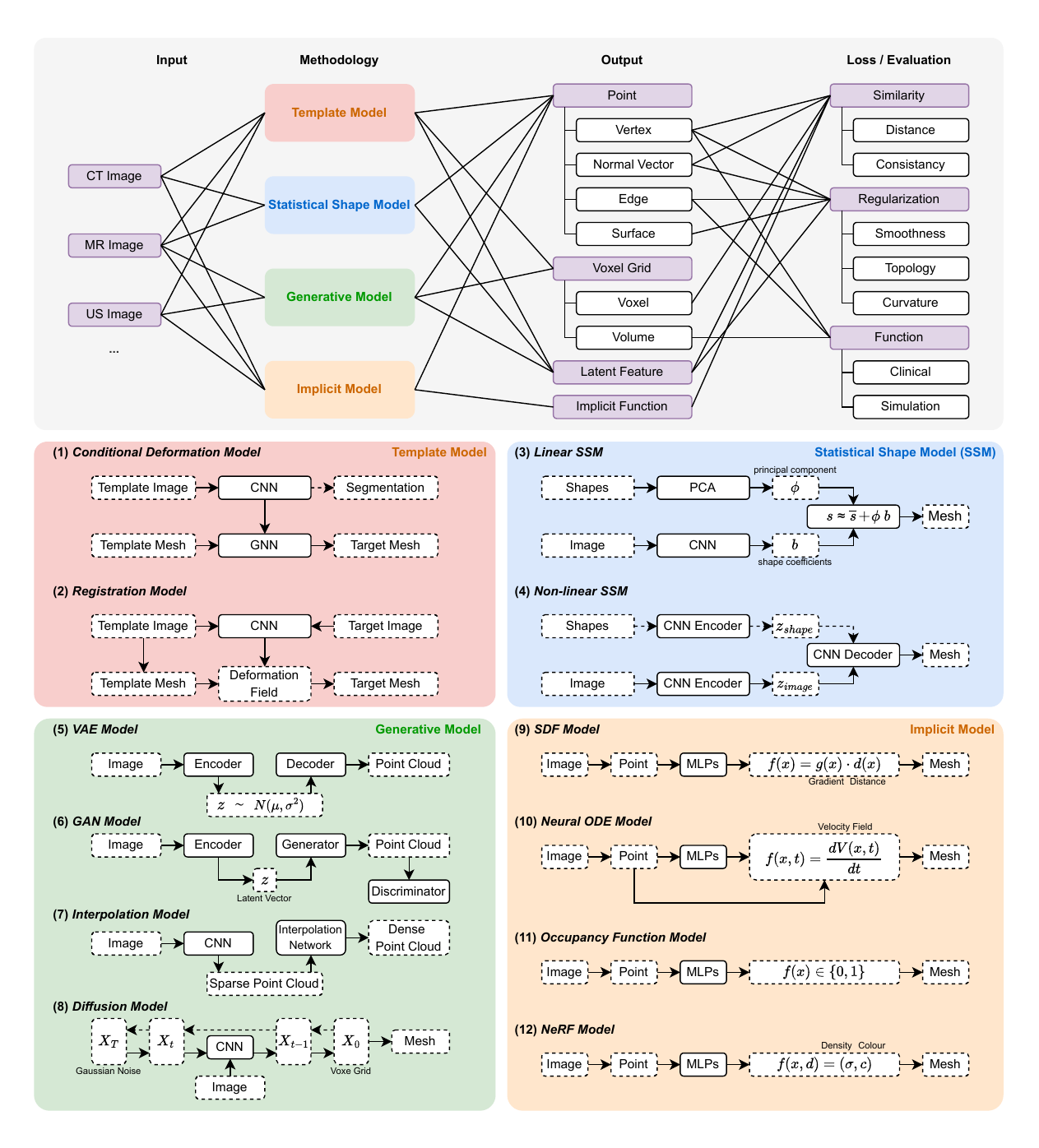}
    \caption{Survey structure, categorizing deep learning-based medical surface reconstruction into four paradigms: template models, statistical shape models (SSM), generative models, and implicit models. Dashed boxes represent variables, while solid boxes represent modules that process these variables using algorithms such as convolutional neural networks (CNN) and graph neural networks (GNN). Solid arrows indicate essential processes, whereas dashed arrows denote optional processes.}
    \label{fig:taxonomy} 
\end{figure*}

Medical image-to-mesh reconstruction methods, shown in Fig.~\ref{fig:taxonomy}, can be broadly categorized into four primary types: statistical shape models, template models, generative models, and implicit models. Each of these methodologies addresses the challenge of reconstructing mesh structures from imaging data through distinct approaches, reflecting differences in theoretical foundations and application scopes. This classification not only highlights the diversity of techniques but also underscores the evolving landscape of mesh reconstruction driven by advancements in deep learning and computational geometry.  

Template models represent a class of approaches that focus on transforming an initial template mesh to fit the anatomy shape derived from input medical images. These models rely heavily on iterative adjustment of vertex positions, allowing for gradual refinement that results in anatomically consistent meshes.

Statistical shape models offer a data-driven perspective, leveraging a large repository of anatomical meshes to construct probabilistic shape representations. By capturing the principal modes of variation within the dataset, these models enable accurate predictions of new mesh structures that conform to the general anatomical shape while preserving individual variations. 

Generative models aim to synthesize mesh structures directly from input data without relying on pre-existing templates. By learning the underlying distribution of anatomical shapes, these models can generate complex surfaces, providing a flexible solution for mesh reconstruction tasks across various imaging modalities.  

Implicit models diverge from the other categories by representing surfaces as continuous functions, allowing for smooth and high-resolution mesh extraction. These methods bypass the need for explicit mesh representations during training, instead learning implicit functions that define surface boundaries indirectly.  

The output representations across these methodologies can take different forms, ranging from discrete point clouds to implicit functions and meshes. 
Ultimately, all these representations can be converted into meshes.
Furthermore, assessing these models depends on similarity metrics that measure the degree of alignment between the reconstructed meshes and the ground truth in terms of point, voxel grid, latent feature, and implicit function. 
Regularization techniques are also frequently used to ensure smoothness, topology, and curvature in case of unrealistic deformations, contributing to robustness and generalizability.  

This taxonomy serves as a framework for understanding the broad spectrum of approaches in medical image-to-mesh reconstruction. By organizing methods according to their processing pipelines and feature representations, this taxonomy provides a structured overview that facilitates a deeper exploration of each technique in subsequent sections.

\section{Template Model}

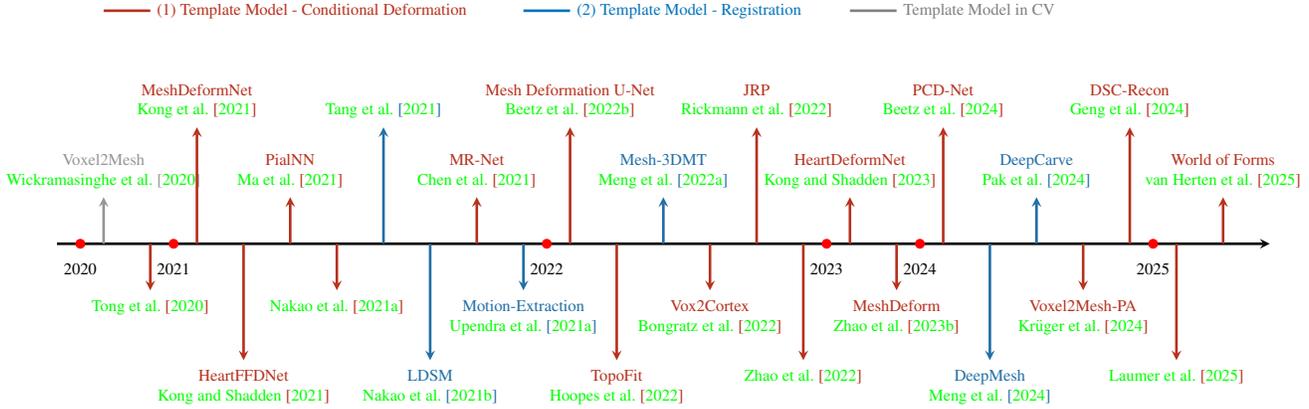
\begin{figure*}
    \centering
    \resizebox{\textwidth}{!}{ 
    \begin{tikzpicture}
        \draw[ultra thick, -stealth, black] (-1,0) -- (25,0) node[right] {};

        \foreach \x/\year in {-0.5/2020, 1.5/2021, 9.5/2022, 15.5/2023, 17.5/2024, 22.5/2025} {
            \fill[red] (\x,0) circle (3pt);
            \node[below] at (\x,-0.3) {\year}; 
        }
        \draw[Gray, ultra thick, -stealth] (0.0,0) -- (0.0, 1.0) node[above] {\makecell[c]{Voxel2Mesh \\ \cite{wickramasinghe2020voxel2mesh}} };
        \draw[BrickRed, ultra thick, -stealth] (1.0,0) -- (1.0, -1.0) node[below] {\makecell[c]{\cite{Fei2020X} } };
        \draw[BrickRed, ultra thick, -stealth] (2.0,0) -- (2.0,2.5) node[above] {\makecell[c]{MeshDeformNet \\  \cite{Fanwei2021Deep}} };
        \draw[BrickRed, ultra thick, -stealth] (3,0) -- (3,-2.5) node[below] {\makecell[c]{HeartFFDNet \\ \cite{Fanwei2021Whole}} };
        \draw[BrickRed, ultra thick, -stealth] (4,0) -- (4,1.0) node[above] {\makecell[c]{PialNN \\ \cite{Qiang2021PialNN}} };
        \draw[BrickRed, ultra thick, -stealth] (5,0) -- (5,-1.0) node[below] {\makecell[c]{\cite{M2021Image}} };\
        \draw[RoyalBlue!80!black, ultra thick, -stealth] (6,0) -- (6,2.5) node[above] {\makecell[c]{\cite{Songyuan2021CNN}} };\
        \draw[RoyalBlue!80!black, ultra thick, -stealth] (7,0) -- (7,-2.5) node[below] {\makecell[c]{LDSM \\ \cite{M2019Statistical}} };
        \draw[BrickRed, ultra thick, -stealth] (8,0) -- (8,1.0) node[above] {\makecell[c]{MR-Net \\ \cite{Xiang2021Shape}} };\
        \draw[RoyalBlue!80!black, ultra thick, -stealth] (9,0) -- (9,-1) node[below] {\makecell[c]{Motion-Extraction \\ \cite{R2021Motion} } };\
        \draw[BrickRed, ultra thick, -stealth] (10,0) -- (10,2.5) node[above] {\makecell[c]{Mesh Deformation U-Net \\ \cite{M2022Reconstructing}} };
        \draw[BrickRed, ultra thick, -stealth] (11,0) -- (11,-2.5) node[below] {\makecell[c]{TopoFit \\ \cite{Andrew2022TopoFit} } };
        \draw[RoyalBlue!80!black, ultra thick, -stealth] (12,0) -- (12,1) node[above] {\makecell[c]{Mesh-3DMT \\ \cite{meng2022mesh} } };  
        \draw[BrickRed, ultra thick, -stealth] (13,0) -- (13,-1.0) node[below] {\makecell[c]{Vox2Cortex \\ \cite{Fabian2022Vox2Cortex} } };
        \draw[BrickRed, ultra thick, -stealth] (14,0) -- (14,2.5) node[above] {\makecell[c]{JRP \\ \cite{Rickmann2022Joint} } };
        \draw[BrickRed, ultra thick, -stealth] (15,0) -- (15,-2.5) node[below] {\makecell[c]{\cite{Jingliang2022Segmentation} } };
        \draw[BrickRed, ultra thick, -stealth] (16,0) -- (16,1) node[above] {\makecell[c]{HeartDeformNet \\ \cite{Fanwei2022Learning}} };  
        
        \draw[BrickRed, ultra thick, -stealth] (17,0) -- (17,-1) node[below] {\makecell[c]{MeshDeform \\ \cite{Junjie2023MeshDeform} } };
        
        \draw[BrickRed, ultra thick, -stealth] (18,0) -- (18,2.5) node[above] {\makecell[c]{PCD-Net \\ \cite{M2023modelling} } };
        \draw[RoyalBlue!80!black, ultra thick, -stealth] (19,0) -- (19,-2.5) node[below] {\makecell[c]{DeepMesh \\ \cite{Qingjie2023DeepMesh} } };
        
        \draw[RoyalBlue!80!black, ultra thick, -stealth] (20,0) -- (20,1.0) node[above] {\makecell[c]{DeepCarve \\ \cite{pak2023patient}} }; 
        
        \draw[BrickRed, ultra thick, -stealth] (21.0,0) -- (21.0,-1.0) node[below] {\makecell[c]{Voxel2Mesh-PA \\ \cite{Nina2023Deep}} };
        \draw[BrickRed, ultra thick, -stealth] (22.0,0) -- (22.0,2.5) node[above] {\makecell[c]{DSC-Recon \\ \cite{geng2024dsc} } };
        \draw[BrickRed, ultra thick, -stealth] (23.0,0) -- (23.0,-2.5) node[below] {\makecell[c]{\cite{laumer20252d} } };
        \draw[BrickRed, ultra thick, -stealth] (24.0,0) -- (24.0,1) node[above] {\makecell[c]{World of Forms \\ \cite{van2025world} } };

        \begin{scope}[shift={(0,5)}]
            \draw[ultra thick, BrickRed] (0,0) -- (1,0) node[right] {(1) Template Model - Conditional Deformation};
            \draw[ultra thick, RoyalBlue] (9,0) -- (10,0) node[right] {(2) Template Model - Registration};
            \draw[ultra thick, gray] (16,0) -- (17,0) node[right] {Template Model in CV};
        \end{scope}

    \end{tikzpicture}
    } 
    \caption{\textcolor{black}{Chronological} overview of representative template models for medical image-to-mesh reconstruction.}
\end{figure*}

\begin{table*}[!ht]
\centering
\caption{Template Models for Medical Image-to-Mesh Reconstruction
}
\resizebox{\textwidth}{!}{\begin{tabular}{>{\centering\arraybackslash}m{4cm} >{\centering\arraybackslash}m{4cm} >{\centering\arraybackslash}m{3cm} >{\centering\arraybackslash}m{3cm} >{\centering\arraybackslash}m{2cm} >{\centering\arraybackslash}m{2cm} >{\centering\arraybackslash}m{6cm}}
\hline
\hline
\textbf{Name} & \textbf{Method}& \textbf{Input} & \textbf{Output}& \textbf{Anatomy}& \textbf{Modality} & \textbf{Dataset} \\
\hline \hline

\makecell[c]{MeshDeformNet \\ \cite{Fanwei2021Deep}} & \makecell[c]{Conditional Deformation} & \makecell[c]{3D Image} & \makecell[c]{Mesh} & \makecell[c]{Heart} & \makecell[c]{CT/MR} & \makecell[c]{MMWHS, OrCaScore, SLAWT, \\ LASC, Private Data} \\ \hline

\makecell[c]{HeartDeformNet \\ \cite{Fanwei2022Learning}} & \makecell[c]{Conditional Deformation} & \makecell[c]{3D Image} & \makecell[c]{Simulation-ready \\ Mesh} & \makecell[c]{Heart} & \makecell[c]{CT/MR} & \makecell[c]{MMWHS, OrCaScore, LASC} \\ \hline
\makecell[c]{3DAngioNet \\ \cite{K20233D}} & \makecell[c]{Conditional Deformation} & \makecell[c]{2D Image} & \makecell[c]{Mesh} & \makecell[c]{Coronary \\ Arteries} & \makecell[c]{X-ray \\ Angiographic} & \makecell[c]{Private Data \\ (Barts Health NHS Trust)} \\ \hline

\makecell[c]{Mesh-3DMT \\ \cite{meng2022mesh}} & Registration & \makecell[c]{2D CMR images \\ (SAX+LAX)}& \makecell[c]{Dynamic Mesh } & Heart & CMR & UK Biobank \\ \hline

\makecell[c]{DeepMesh \\ \cite{Qingjie2023DeepMesh}} & Registration & \makecell[c]{2D CMR Images \\ (SAX, 2CH, 4CH)} & \makecell[c]{Dynamic Mesh } & Heart & CMR & UK Biobank \\ \hline
\makecell[c]{PCD-Net \\ \cite{M2023modelling}} & Conditional Deformation & Point cloud at ED/ES phase & Point Cloud at ES/ED & \makecell[c]{Heart \\ (LV/RV)} & CMR & UK Biobank \\ \hline 

\makecell[c]{Vox2Cortex \\ \cite{Fabian2022Vox2Cortex}} & Conditional Deformation & 3D Image & Surface Mesh & Brain Cortex & MRI-T1 & ADNI, OASIS-1, TRT \\ \hline
\makecell[c]{MeshDeform \\ \cite{Junjie2023MeshDeform}} & Conditional Deformation & MRI-T1w/T2w & Surface Mesh & Brain Cortex  & MR & HCP \\ \hline
LDSM \cite{M2019Statistical} & Registration & 4D Contours & Tumor position and deformation & Abdominal Tumor & CT & Private Data (25) \\ \hline

\makecell[c]{Mesh Deformation U-Net \\ \cite{M2022Reconstructing}} & Conditional Deformation & 2D Image & Mesh & Heart (LV/RV) & MR & UK Biobank, Private Data (250) \\ \hline

\makecell[c]{HeartFFDNet \\ \cite{Fanwei2021Whole}} & Conditional Deformation & 3D Image & Surface Mesh & Heart & CT & MMWHS, orCalScore, SLAWT \\  \hline
\makecell[c]{TopoFit \\ \cite{Andrew2022TopoFit}} & Conditional Deformation & 3D Image & Surface Mesh & Brain Cortex & MR & OASIS, IXI, MCIC, Buckner40 \\ \hline
\makecell[c]{PialNN \\ \cite{Qiang2021PialNN}}& Conditional Deformation & 3D Image & Surface Mesh & Brain Cortex & MR & HCP (300) \\ \hline
\makecell[c]{JRP \\ \cite{Rickmann2022Joint}}& Conditional Deformation & 3D Image & Multi-class Surface Mesh & Brain Cortex & MR & OASIS-1 \\ \hline
\makecell[c]{- \\ \cite{Jingliang2022Segmentation}} & Conditional Deformation & 3D Image & Surface Mesh & Aortic Dissection & CTA & WAD (35) \\ \hline
\makecell[c]{ - \\\cite{Fei2020X}} & Conditional Deformation & Single 2D Image & Surface mesh & Liver & CT & Private Data \\ \hline
\makecell[c]{ - \\ \cite{M2021Image}} & Conditional Deformation & 2D Projection Image & Surface Mesh & Abdominal Organs & CT & Private Data (474) \\ \hline
\makecell[c]{ - \\ \cite{Songyuan2021CNN}} & Registration & 2D Image & 3D Surface Mesh & Spine & US & Private Data (10) \\ \hline 
\makecell[c]{Voxel2Mesh-PA \\ \cite{Nina2023Deep}} & Conditional Deformation & 3D Image & Surface mesh & Pulmonary & CT & Private Data (58) \\ \hline

\makecell[c]{MR-Net \\ \cite{Xiang2021Shape}} & Conditional Deformation & 2D Contour & Surface Mesh & Bi-ventricular heart & MR & UK Biobank \\ \hline
\makecell[c]{DSC-Recon \\ \cite{geng2024dsc}} & Conditional Deformation & 3D Image & Dynamic Mesh & Liver, Lung & CT/X-ray & 4D Preoperative CT (20 lung, 45 liver) + 13 Intraoperative CT/X-ray pairs \\ \hline
\makecell[c]{Motion-Extraction \\ \cite{R2021Motion}} & Registration & 4D Image & Dynamic Mesh & Right Ventricle & CMR & ACDC \\ \hline 

\makecell[c]{ - \\ \cite{laumer20252d}} & Conditional Deformation & 2D Image & Surface Mesh & \makecell[c]{Heart \\ (LV)} & Echocardiographic & 4D SSM (458), SPUM-ACS (458)\\ \hline  

\makecell[c]{ GHD-DVS  \\ \cite{luo2024explicit}} & Registration & 3D Image & Mesh & \makecell[c]{Heart \\ (LV)} & CT/MR & MMWHS, CCT48, ACDC, UK Biobank, MITEA  \\ \hline 

\end{tabular}%
}

\end{table*}

A template model is a type of image-to-mesh reconstruction method that operates by deforming a template mesh. The core idea is to start with an initial template mesh and iteratively deform it to match the input medical image. This approach is particularly suitable for medical images with relatively simple topology, such as the heart or abdominal organs (not vessels), as it leverages prior anatomical knowledge to produce physically plausible meshes. Deformation models can be divided into two main categories: {conditioned deformation models} and {registration models}. 
A conditional deformation model extracts deformation information through a segmentation pipeline to guide the deformation of a template. In contrast, a registration model computes a deformation field to direct the template’s transformation.
Although these categories differ in how the template deformation is achieved, their shared goal is to generate accurate structure meshes through learned deformations.

\subsection{Conditioned Deformation Methods}

\begin{figure}[!ht]
    \centering
    \includegraphics[width=\linewidth]{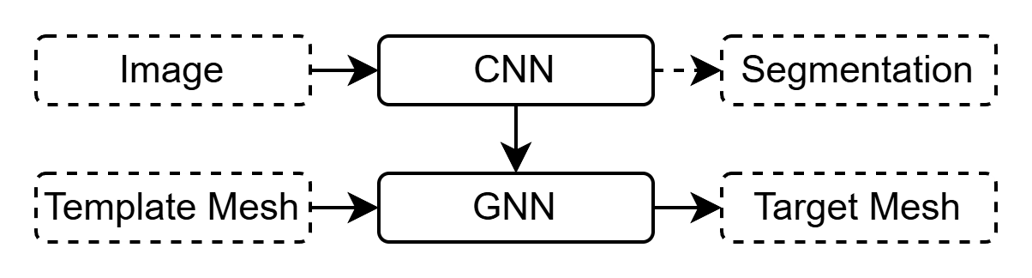}
    \caption{Schematic of conditioned deformation methods. The framework consists of two pipelines: one leveraging a CNN for deformation feature extraction and the other employing a GNN for direct template mesh deformation. During feature propagation, the CNN pipeline transfers learned features from the image to the GNN pipeline. Dashed boxes represent variables, while solid boxes denote operations. Solid arrows indicate mandatory connections, and dashed arrows represent optional connections.}
    \label{fig:DM1}
\end{figure}

As shown in Fig.~\ref{fig:DM1}, conditional deformation models employ graph convolution networks to iteratively deform the template mesh. Typically, a CNN extracts features from the input medical image, which are then passed to the GNN to guide the vertex-wise deformation of the template. Since GNNs are well-suited for handling non-Euclidean data such as meshes and point clouds, they are highly effective for template-based deformation tasks. The advantage of this method lies in its ability to preserve the topological structure of the original template while producing meshes that conform to the medical image.

\cite{Fanwei2021Deep, Fanwei2021Whole} introduced an additional free-form deformation (FFD) module in the GNN to enhance smoothness.
Specifically, the method \cite{Fanwei2021Whole} employs a deep learning model and template deformation technique to generate whole heart meshes matching input image data by predicting displacements of multi-resolution control point grids. The method first processes input images through an image encoding module then samples image features via a feature sampling module, uses a deep FFD module to predict control point displacements for template mesh deformation, and employs a segmentation module for additional supervision.
The method \cite{Fanwei2021Deep} uses a graph convolutional neural network to directly predict whole heart surface meshes from volumetric CT and MR image data, reconstructing multiple anatomical structures by deforming a predefined mesh template, and can reconstruct whole heart dynamics from 4D imaging data.

Based on \cite{Fanwei2021Deep} and \cite{Fanwei2021Whole}, \cite{Fanwei2022Learning} used biharmonic coordinates to further improve the topological quality of the reconstructed mesh, enabling the reconstruction of simulation-suitable meshes.
Specifically, the method uses deep learning techniques to construct simulation-suitable heart models from 3D patient image data (CT and MRI) by deforming a whole heart template, generating anatomically and temporally consistent geometries.

\cite{M2021Image, nakao2022image} trained an additional deformation map with the CNN, which directly acts on the template. This method proposes an Image-to-Graph Convolutional Network (IGCN+) framework that integrates an image generative network and graph convolutional network to reconstruct 3D organ shapes from a single two-dimensional (2D) projection image, which is particularly effective for low-contrast organs. The method first uses an image generative network to create displacement maps from 2D projection images, then employs a graph convolutional network to learn vertex features and generate the final 3D deformed mesh.

Same with the structure in Fig.~\ref{fig:DM1}, the input used by \cite{Xiang2021Shape} is not an image but a sparse and incomplete point cloud. Therefore, format transfer is performed both before and after feeding it into the CNN. Features are extracted from both the point cloud and the converted voxel, and these features are then combined.
Specifically, this method presents MR-Net, a deep learning architecture for 3D shape reconstruction from sparse and incomplete point clouds. The method consists of a feature extraction module and a deformation module. The feature extraction module extracts features from input contours through point cloud feature extraction and 3D CNN, while the deformation module uses three GNN blocks to gradually deform a template mesh under the guidance of extracted features. The method innovatively designs a point-to-point mapping mechanism to address the challenge of feature mapping between unstructured data.

\cite{Andrew2022TopoFit} introduced TopoFit, a fast cortical reconstruction method that deforms a topologically correct template mesh to fit the target anatomy. To ensure local structural accuracy and maintain topological consistency across different scales, the approach incorporates manifold distance loss and manifold regularization loss as additional constraints. This allows for the direct reconstruction of a topologically correct surface without requiring post-processing to fix topology errors. The model integrates image and graph convolutions with an efficient symmetric distance loss to learn deformation fields. It first extracts features from input images using a convolutional network, then progressively deforms the template mesh through multiple graph convolution blocks. Each block samples image features and predicts local deformations, ultimately producing an anatomically accurate mesh.

\cite{Qiang2021PialNN} proposed PialNN, this model did not use GNN as is used in a typical template model in Fig.~\ref{fig:DM1}, but instead directly employed multilayer perception (MLP) layers in the PialNN model to fuse image features and point features.
Experiments demonstrated that the performance of GNN and MLP in brain pial surface reconstruction is similar in terms of accuracy, but MLP significantly reduces computational demand compared to GCN.
Specifically, PialNN is a deep learning framework that deforms an initial white matter surface to a pial surface through a sequence of learned deformation blocks. Each block incorporates vertex features and local MR features using local convolutional operations to capture multi-scale information. The method employs three deformation blocks, each extracting point features and local MR features to predict vertex displacements for surface deformation, followed by Laplacian smoothing for optimization.

\cite{Rickmann2022Joint} and \cite{Fabian2022Vox2Cortex} improve the template to a multi-class template and calculated reconstruction loss separately for each class.
Specifically, \cite{Rickmann2022Joint} propose CSR, a graph classification branch and a novel 3D reconstruction loss approach to enhance template-deformation algorithms, achieving joint reconstruction and atlas-based parcellation of cortical surfaces. The method first deforms template meshes using CSR networks, then achieves parcellation either through a graph classification network or class-based reconstruction loss, with end-to-end training.
\cite{Fabian2022Vox2Cortex} propose Vox2Cortex, this algorithm employs geometric deep neural networks, combining convolutional and graph convolutional networks, to directly reconstruct high-resolution 3D meshes of cortical surfaces from MR scans by deforming an initial template mesh. The method introduces a curvature-weighted Chamfer loss function for better accuracy in highly folded regions.

Similarly, \cite{Jingliang2022Segmentation} tackle mesh quality issues by incorporating a homogeneity-optimized stepwise mesh regression module, which refines the output mesh to suppress folding and uneven distribution problems. Their morphology-constrained stepwise deep mesh regression (MSMR) method segments the true lumen of aortic dissection by first generating an initial mesh based on centerlines, then estimating offsets using deep feature encoders and graph convolutional decoders, and finally deforming the mesh to the target surface through multiple regression steps. Both approaches leverage deep learning for template-based mesh deformation, with Vox2Cortex prioritizing cortical surface reconstruction and MSMR refining anatomical meshes to improve structural integrity. Their shared emphasis on regularization strategies, curvature-aware loss in Vox2Cortex and homogeneity optimization in MSMR, highlights the growing role of geometric deep learning in accurate, topology-preserving anatomical mesh reconstruction.

3D shape reconstruction from 2D projections using a template-based approach is performed by \cite{Fei2020X} in the X-ray2Shape model. 
X-ray2Shape is a deep learning framework combining GNN and CNN that reconstructs a 3D liver mesh from a single 2D X-ray projection image. The method learns mesh deformation from a mean template using deep features computed from individual projection images. The method first extracts features from digitally reconstructed radiograph (DRR) images using CNN, then concatenates these features with 3D vertex coordinates and feeds them into a GNN network for mesh deformation, optimizing the reconstruction using MSE loss and Laplacian loss. 
Similarly to the method in \cite{Fei2020X}, \cite{Yifan2019DeepOrganNet} proposed DeepOrganNet, which uses 2D projection images as input and incorporates the FFD method.
Specifically, DeepOrganNet employs a deep neural network to reconstruct 3D/4D lung models from single-view 2D medical images, learning smooth deformation fields from multiple templates and extracting latent descriptors to generate high-fidelity lung model meshes.
The method first encodes the input image using MobileNets \cite{howard2017mobilenets} to obtain feature descriptors, then reconstructs left and right lungs through two independent branches, each learning optimal template selection and smooth deformation based on FFD, finally combining them through a spatial arrangement module to generate the final two-lung model.

\cite{K20233D} added an additional CNN branch to the typical structure for template-based models in Fig.~\ref{fig:DM1} to extract the centerline, which is then used to reconstruct the template surface mesh of coronary vessels.
\cite{K20233D} propose 3DAngioNet, this method combines an EfficientB3-UNet \cite{tan2019efficientnet} segmentation network with graph convolutional networks for deformation to reconstruct 3D coronary vessels from bi-plane angiography images, capable of handling bifurcated vessels.

Based on Fig.~\ref{fig:DM1}, \cite{Junjie2023MeshDeform} sampled the output of the feature by the encoder part of the CNN into the GNN for reconstruction. On the contrary, \cite{Nina2023Deep} sampled the features from the decoder part of the CNN into the GCN. Moreover, the output of the CNN does not include a segmentation task as a constraint in \cite{Nina2023Deep}.
Specifically, MeshDeform \cite{Junjie2023MeshDeform} is an end-to-end deep learning network for reconstructing subcortical structure surfaces from brain MR images. The method employs a 3D U-Net encoder to extract multi-scale features and combines them with a graph convolutional network to predict subcortical surfaces by deforming spherical mesh templates. The network contains three mesh deformation blocks, each utilizing graph convolutions to predict vertex displacements. This approach can efficiently and accurately reconstruct subcortical surfaces while preserving spherical topology.
Meanwhile, \cite{Nina2023Deep} present a deep learning approach for automatic pulmonary artery surface mesh generation from CT images. Based on the Voxel2Mesh proposed by \cite{wickramasinghe2020voxel2mesh} algorithm, it combines voxel encoder-decoder and mesh decoder to deform a prototype mesh (sphere) into the target mesh. The key innovations include a centerline coverage loss to facilitate branching structure formation and vertex classification for defining inlets and outlets. The method can automatically generate high-quality meshes suitable for hemodynamic simulation while significantly reducing manual interaction time.

\cite{geng2024dsc} enriched the framework structure of Fig.~\ref{fig:DM1} by reconstructing 4D dynamic meshes from different modalities. The method is divided into three steps. In the first step, CT images are segmented to obtain meshes at two time points, and interpolation is performed to estimate meshes at intermediate time points, which are used as templates. In the second step, enhanced projection images are obtained from the CT volume as input and, together with the mesh templates, fed into the framework in Fig.~\ref{fig:DM1} to obtain the reconstructed meshes. In the third step, during inference, a domain adaptation network trained for X-ray to CT style conversion is applied, and the converted images are used as input for reconstruction.  

Instead of directly inputting a template mesh into the GNN, \cite{M2022Reconstructing, M2022Mesh, M2023modelling} first obtained a point cloud contour through image segmentation. They then combined the contour with the template mesh to create a misaligned mesh, which was subsequently aligned using the GNN to generate a high-quality mesh.

GNN-based registration methods like \cite{hansen2021graphregnet} are an alternative variant of template methods designed to address the issue of misaligned meshes during reconstruction. In this approach, a rough, misaligned mesh is first generated before being fed into the GNN. The GNN then learns to refine and correct the misalignment, gradually deforming the input mesh to align it with the target shape.  
By introducing this initial misalignment, the network is encouraged to develop a stronger understanding of local and global surface features, enhancing its ability to handle complex deformations. Alignment-based methods are particularly effective in scenarios where the input data is incomplete or noisy, as the GNN can iteratively improve mesh quality and ensure accurate surface reconstruction. This process ultimately reduces errors in the final mesh, yielding smoother and more topologically consistent results.

\cite{M2022Reconstructing} propose a Mesh Deformation U-Net deep learning method that combines spectral graph convolutions and mesh sampling operations to reconstruct 3D cardiac surface meshes directly on mesh data through multi-scale feature learning, using a template mesh as a prior to correct misalignment in multi-view MR slices.
The method follows a 4-step pipeline where each MR slice is first segmented using CNNs, the segmentation contours are then placed in 3D space as point clouds, a template mesh is approximately fitted to the sparse contours, and finally the Mesh Deformation U-Net corrects any slice misalignment to output the reconstructed 3D cardiac shape.
Based on \cite{M2022Reconstructing}, \cite{M2022Mesh} applied GNN not for correcting misaligned meshes but to model deformations between different phases of the cardiac cycle. The method utilizes two separate networks to predict cardiac contraction and relaxation: one network predicts the end-systolic (ES) shape from end-diastolic (ED) input, while the other predicts the ED shape from ES input. Both networks leverage spectral graph convolutions for feature extraction and employ a multi-scale encoder-decoder architecture to process mesh data, enabling accurate and efficient reconstruction of cardiac dynamics.

\cite{M2023modelling} builds upon their previous work \cite{M2022Reconstructing, M2022Mesh}, enabling the transformation between end-diastole (ED) and end-systole (ES) after reconstruction. Additionally, the method can perform tasks such as disease classification based on latent features.
This point cloud deformation network (PCD-Net) is a deep learning approach for modelling 3D cardiac deformation during contraction and relaxation. The method employs an encoder-decoder architecture that enables efficient multi-scale feature learning directly on multi-class 3D point cloud representations of cardiac anatomy. The network takes a point cloud representation of the heart at one phase (end-systolic or end-diastolic) as input and predicts the shape at the other phase. 

\cite{van2025world} does not extract deformation features from a CNN-based segmentation branch. Instead, it directly samples image intensities along rays cast from a central point to each vertex of a template mesh. These 1D ray signals are then encoded using a 1D CNN or a masked autoencoder to obtain per-vertex features, which are subsequently processed by a GNN to predict mesh deformations. By leveraging geometric priors such as spherical or tubular templates, the method reformulates mesh prediction as a radial regression task along fixed ray directions, enabling topologically consistent surface meshing with strong performance even in low-data regimes.

\cite{laumer20252d} also adopt a template-based deformation approach; however, unlike typical methods, the template mesh is not directly used as an input during forward propagation. Instead, the template (mean mesh from a statistical shape model) is utilized to construct the structure of the GNN decoder by generating multi-resolution mesh representations and computing the corresponding upsampling and adjacency matrices. This design fixes the decoder’s topology and enables the network to progressively decode a full-resolution left ventricular mesh from CNN-extracted features.

\subsection{Template-based Registration Methods}

Registration-based methods directly learn the mapping from medical images to deformation fields. Typically, a CNN extracts features from two different medical images, predicting a deformation field that deforms the initial template into the target mesh. The key to this method is the accuracy and smoothness of the deformation field, which can capture both local and global deformation information. This makes it particularly suitable for tasks involving multi-time point data or registration-based mesh generation. Deformation Field methods are flexible and efficient, making them ideal for large-scale medical image processing and deformation tasks.

\begin{figure}[!ht]
    \centering
    \includegraphics[width=\linewidth]{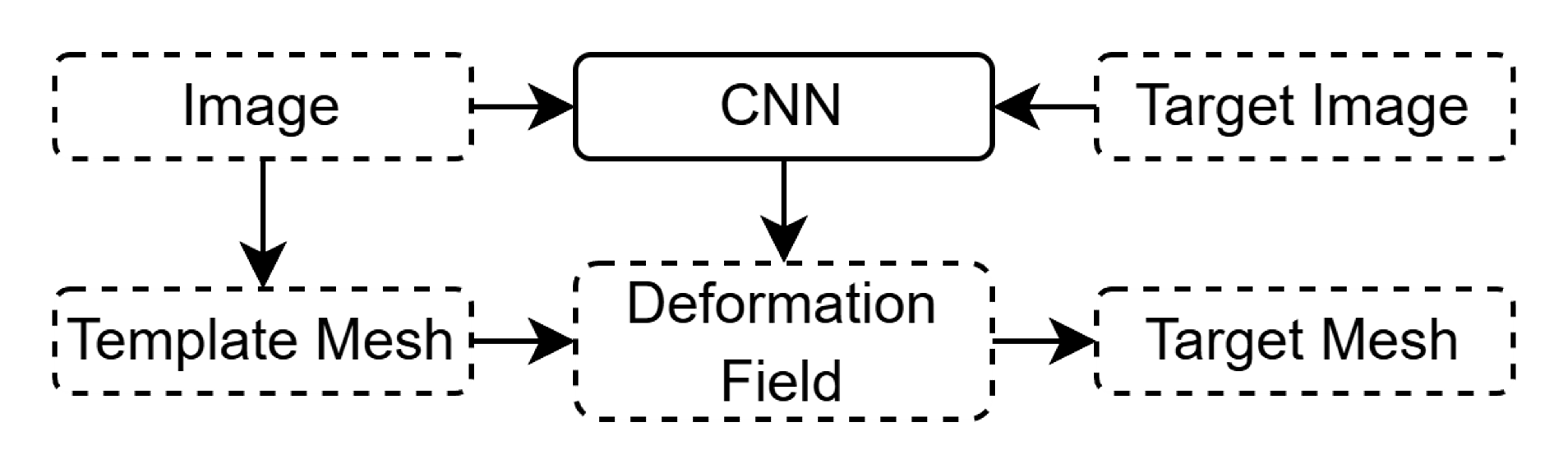}
    \caption{Schematic of template-based registration methods. The process begins by generating a template mesh from the input image using segmentation or registration techniques. The image and template image are then passed into a CNN to compute the deformation field, which is subsequently applied to the template mesh to generate the target mesh.}
    \label{fig:DM4 deformation field}
\end{figure}

\cite{meng2022mulvimotion} proposed MulViMotion, a shape-aware 3D myocardial motion tracking framework that estimates 3D heart motion from multi-view 2D cine CMR. The method integrates SAX and LAX views using a hybrid 2D/3D deep learning network. FeatureNet extracts motion and shape features, while MotionNet predicts dense 3D motion fields. A shape regularization module enforces consistency between predicted 3D myocardial edge maps and ground-truth 2D contours. Trained in a weakly supervised manner.

Based on Fig.~\ref{fig:DM4 deformation field}, \cite{meng2022mesh} used a differentiable mesh-to-image rasterizer to directly supervise the mesh with voxel data.
This method uses a deep learning model and differentiable mesh-to-image rasterizer to estimate 3D heart mesh motion from 2D short- and long-axis cardiac MR images, enabling quantitative assessment of cardiac function by tracking the motion of each vertex in the 3D mesh.
In a subsequent study, \cite{Qingjie2023DeepMesh} used a deep learning framework with a differentiable mesh-to-image rasterizer to reconstruct 3D heart meshes and estimate their motion from 2D cardiac MR images, leveraging 2D shape information from multiple views for 3D reconstruction and motion estimation. Notably, this paper performs bi-directional representation transformations between discrete and continuous domains: converting from continuous mesh representations to discrete voxel-based probability maps through differentiable rasterization, and sampling from discrete voxel-based motion fields to obtain continuous vertex-wise displacements via a grid sampling mechanism.

\cite{pak2023patient} employs a registration-based strategy, DeepCarve, where a 3D CNN is used to predict a space-deforming field that is applied to a structured template volumetric mesh. The deformation field is sampled via trilinear interpolation to generate displacement vectors for each template point, enabling smooth and accurate transformation. To achieve label-efficient training, the model relies only on minimal surface mesh labels and jointly optimizes for spatial accuracy and volumetric mesh quality using a combination of isotropic and anisotropic energy terms. The resulting patient-specific heart meshes—including the ventricles, aorta, and valve leaflets—achieve high spatial fidelity and element quality. It can be directly used for downstream finite element biomechanical simulations.

Aligned with the structure in Fig.~\ref{fig:DM4 deformation field}, 
\cite{R2021Motion} used a CNN to learn the deformation field between multi-phase images, which is then applied to deform the mesh.
This is a deep learning-based deformable registration framework for extracting right ventricle motion from 4D cardiac cine MRI. It first uses CondenseUNet \cite{hasan2020condenseunet} for segmentation, then employs VoxelMorph \cite{balakrishnan2019voxelmorph} for image registration, and finally generates geometric models for all cardiac cycle frames by propagating the end-diastole isosurface mesh.

\cite{Songyuan2021CNN} used a traditional registration method \cite{chui2003new} to learn the deformation field between local images (ultrasound, US) and global images (CT). The learned series of deformation fields (DF) is then applied to mesh deformation.
This is a CNN-based method for reconstructing 3D spine surfaces from untracked freehand ultrasound images. The method combines a U-net deep learning network for spine surface detection with registration-based scan path estimation to achieve reconstruction from 2D ultrasound images to 3D surfaces. 
The paper converts detected surface points to volumetric images, then applies Gaussian filtering and thresholding to generate the final surface reconstruction.

\section{Statistical Shape Model} 

Statistical shape models (SSMs) \cite{cootes1995active} are widely used in medical image analysis and computer vision to describe and analyze the variability of anatomical structures. By representing shapes in a low-dimensional space, SSMs enable accurate shape reconstruction and efficient analysis. Traditional SSMs rely on techniques such as principal component analysis (PCA) or singular value decomposition (SVD) to extract the primary modes of variation. In contrast, deep learning-based SSMs harness neural networks to model more complex, non-linear shape deformations directly from images or point clouds.

Traditional SSMs \cite{heimann2009statistical} encompass various approaches, including active shape models (ASM), point distribution models (PDM), point set models (PSM), and template models. Among these, PDM is widely applied in medical image-to-mesh reconstruction tasks \cite{Guillermo2018Weighted, S2018Assessment}.

PDM \cite{cootes1995active} is a classical method for statistical shape modelling that represents a shape using a set of landmarks. By applying statistical techniques such as PCA, PDM analyzes the distribution and variability of these landmarks across different samples. The core concept of PDM is to model and compress the primary modes of shape variation, allowing complex shapes to be represented by a small number of parameters. This not only simplifies shape generation and matching but also facilitates efficient shape analysis and reconstruction.

Although traditional SSMs are effective in capturing shape variability, they are inherently limited to modelling linear shape variations. This constraint makes it challenging to accurately represent highly complex or non-linear deformations that are often encountered in real-world anatomical structures. To overcome these limitations, deep learning-based SSMs have emerged as a powerful alternative. By leveraging neural networks, these methods can model intricate, non-linear shape variations directly from image data, significantly enhancing the flexibility and accuracy of shape reconstruction.

\begin{figure*}
    \centering
    \resizebox{0.8\textwidth}{!}{ 
    \begin{tikzpicture}
        \draw[ultra thick, -stealth, black] (0,0) -- (15,0) node[right] {};

        \foreach \x/\year in {} {
            \fill[red] (\x,0) circle (3pt);
            \node[below] at (\x,-0.3) {\year}; 
        }

        \draw[BrickRed, ultra thick, -stealth] (2,0) -- (2,1) node[above] {\makecell[c]{\\ \cite{attar2019high} } };
        
        \draw[BrickRed, ultra thick, -stealth] (3,0) -- (3,-1) node[below] {\makecell[c]{3DCSP-DNN \\ \cite{attar20193d} } };
        \draw[orange, ultra thick, -stealth] (4,0) -- (4,2.5) node[above] {\makecell[c]{\cite{K2020Artificial}} };
        \draw[orange, ultra thick, -stealth] (5,0) -- (5,-2.5) node[below] {\makecell[c]{\cite{Abhirup2021completely} } };
        \draw[BrickRed, ultra thick, -stealth] (6,0) -- (6,1) node[above] {\makecell[c]{Neural-LV-Geometry \\ \cite{Lukasz2021Neural} } };
        \draw[BrickRed, ultra thick, -stealth] (7,0) -- (7,-1) node[below] {\makecell[c]{\cite{Abhirup2022Automated} } };
        \draw[BrickRed, ultra thick, -stealth] (8,0) -- (8,2.5) node[above] {\makecell[c]{MCSI-Net \\ \cite{xia2022automatic} } };
        \draw[orange, ultra thick, -stealth] (9,0) -- (9,-2.5) node[below] {\makecell[c]{DISSM \\ \cite{raju2022deep}} };
        \draw[BrickRed, ultra thick, -stealth] (10,0) -- (10,1) node[above] {\makecell[c]{4D Myocardium \\ \cite{Xiaohan20234D}} };
        \draw[orange, ultra thick, -stealth] (11,0) -- (11,-1) node[below] {\makecell[c]{\cite{Sachin2023deep} } };
        \draw[RoyalBlue, ultra thick, -stealth] (12,0) -- (12,2.5) node[above] {\makecell[c]{Progressive DeepSSM \\ \cite{Aziz2023Progressive} } };
        \draw[RoyalBlue, ultra thick, -stealth] (13,0) -- (13,-2.5) node[below] {\makecell[c]{DeepSSM \\ \cite{Riddhish2021DeepSSM}} };

        \begin{scope}[shift={(0,4)}]

            \draw[ultra thick, BrickRed] (1,0) -- (2,0) node[right] {(3) Linear SSM};
            \draw[ultra thick, RoyalBlue] (6,0) -- (7,0) node[right] {(4) Non-linear SSM};
            \draw[ultra thick, orange] (11,0) -- (12,0) node[right] {Hybrid SSM};
        \end{scope}

    \end{tikzpicture}
    } 
    \caption{\textcolor{black}{Chronological} overview of representative deep learning-based statistical shape models for medical image-to-mesh reconstruction.}
\end{figure*}
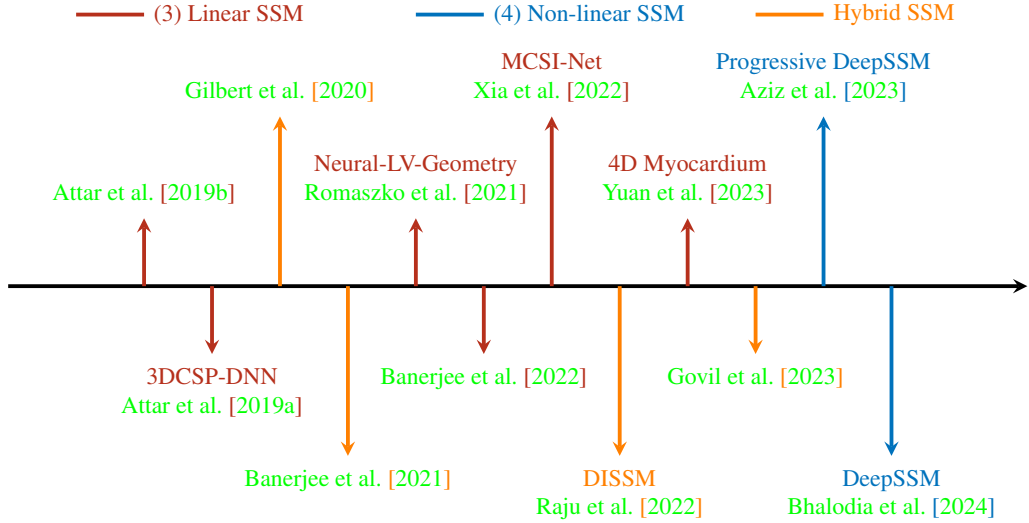

\begin{table*}[!ht]
\centering
\caption{Statistical Shape Models for Medical Image-to-Mesh Reconstruction
}
\resizebox{\textwidth}{!}{
\begin{tabular}{>{\centering\arraybackslash}m{4cm} >{\centering\arraybackslash}m{4cm} >{\centering\arraybackslash}m{3cm} >{\centering\arraybackslash}m{3cm} >{\centering\arraybackslash}m{2cm} >{\centering\arraybackslash}m{2cm} >{\centering\arraybackslash}m{6cm}}
\hline
\hline
\textbf{Name} & \textbf{Method}& \textbf{Input} & \textbf{Output}& \textbf{Anatomy}& \textbf{Modality} & \textbf{Dataset} \\ 
\hline \hline
\makecell[c]{ 3DCSP-DNN \\ \cite{attar20193d} }& Linear SSM & Multi-view image, metadata & Mesh & Heart& CMR  & UK Biobank  \\ \hline
\makecell[c]{MCSI-Net \\ \cite{xia2022automatic}} & Linear SSM & Multi-view image, metadata & Mesh & Heart& CMR  & UK Biobank \\ \hline 

\makecell[c]{DeepSSM \\ \cite{Riddhish2021DeepSSM}} & Non-linear SSM & 3D Image & Point Cloud & Craniosynostosis, LA, Femur, &CT, MR & Metopic (120 CT), LA (206 MRI), Femur (49 CT) \\ \hline

\makecell[c]{4D Myocardium \\ \cite{Xiaohan20234D}} & Hybrid SSM & Sparse point cloud of CMR slices & Dynamic Mesh & \makecell[c]{Heart \\ Myocardium}& CMR &Private data (55), ACDC (100)\\ \hline

\makecell[c]{Neural-LV-Geometry \\ \cite{Lukasz2021Neural}} & Linear SSM & 2D cine images (6 SA + 3 LA) & Dynamic Mesh & \makecell[c]{Heart \\ LV} & CMR & Private dataset (64 HV + 118 MI patients) \\ \hline

\makecell[c]{  \\ \cite{Abhirup2022Automated}} & Hybrid SSM & 2D CMR Slices & Whole-Heart Mesh & \makecell[c]{Heart \\ (4 Chamber)} & CMR & UK Biobank (30) + Private CT (134) \\ \hline

\makecell[c]{Progressive DeepSSM \\ \cite{Aziz2023Progressive}} & Non-linear SSM & 3D Image & Mesh & Femur, Left Atrium & CT, MR & Femur (59), Left Atrium (206) \\ \hline

\makecell[c]{WR-SSM \\ \cite{Guillermo2018Weighted}} & Linear SSM & 2D photo / 3D point clouds & Mesh & Breast & Photo/Point Cloud & 310 Point Clouds, 510 Photos \\ \hline

\makecell[c]{  \\\cite{K2020Artificial}} & Hybrid SSM & Cardiac MR & Surface/Volume Mesh & \makecell[c]{Heart \\ (4 Chamber)} & MR & MESA, UK Biobank \\ \hline 

\makecell[c]{  \\\cite{Sachin2023deep}}& Hybrid SSM & 3D Image & Mesh & \makecell[c]{Heart \\ (LV, RV)} & CMR & Private Data (111) \\  \hline

\makecell[c]{  \\ \cite{Abhirup2021completely}}& Hybrid SSM & 2D Image & Mesh & \makecell[c]{Heart \\ (LV)} & CMR, ECGI & Private Data (20) \\  \hline

\makecell[c]{DISSM \\ \cite{raju2022deep}} & Hybrid SSM & 3D Image & SDF & Liver & CT & MSDliver\\

\hline 
\end{tabular}%
}

\end{table*}

\subsection{Deep Learning-based Linear SSM}

\begin{figure}[!ht]
    \centering
    \includegraphics[width=\linewidth]{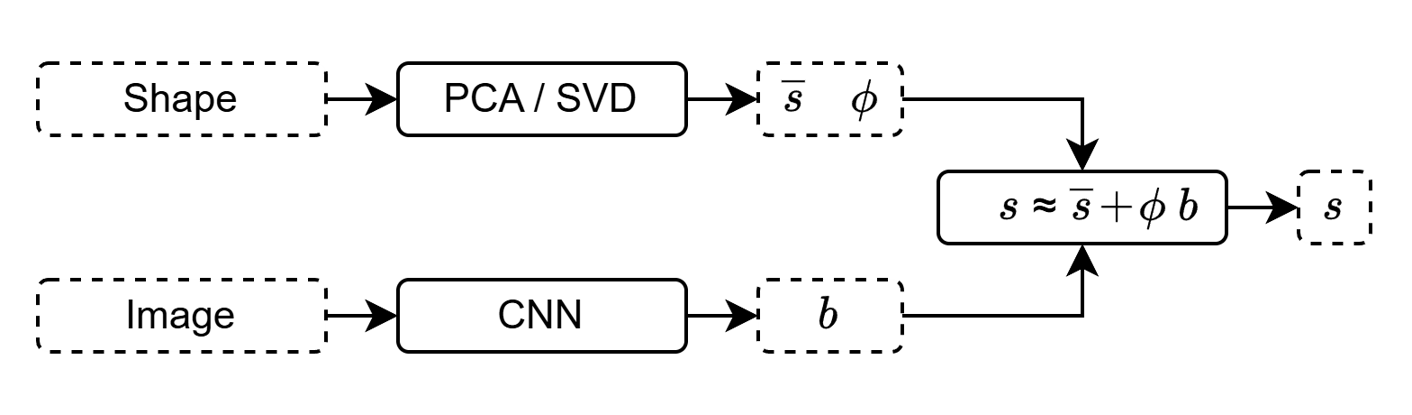}
    \caption{Schematic of Deep Learning-based Linear SSM. Dotted boxes represent variables. Solid boxes represent functions.} 
    \label{figssm1}
\end{figure}

Deep learning-based linear SSMs extend the traditional approach by automating the prediction of shape coefficients from input images with a deep learning encoder. As shown in Fig.~\ref{figssm1}, shape data undergoes PCA or SVD decomposition, resulting in the mean shape $\bar{s}$ and the principal component matrix $\phi$. A CNN predicts the shape coefficients $b$ from an image. The final shape $s$ is reconstructed by:
\[
s \approx \bar{s} + \phi b.
\]
This method automates the process of mapping from image to shape but is constrained by the linear nature of PCA.  

\cite{attar20193d}, \cite{xia2022automatic}, and \cite{Lukasz2021Neural} all integrated deep learning with statistical shape modelling techniques to achieve 3D cardiac shape reconstruction. \cite{attar20193d} utilized a deep neural network combining statistical shape modelling and convolutional neural networks, incorporating both cardiovascular magnetic resonance images and patient metadata as input for 3D cardiac shape prediction. 
Building upon this, \cite{xia2022automatic} proposed MCSI-Net. Except for input multi-view images, this method also incorporates metadata, which aids in predicting shape bias. It integrates their deep learning encoder with an SSM to reconstruct the four heart chambers from cardiac MR images. 
Likewise, \cite{Lukasz2021Neural} employ a deep learning pipeline that integrates image segmentation and PCA to generate a low-dimensional representation for reconstructing the 3D left ventricular geometry from cardiac MR images.

\subsection{Deep Learning-based Non-linear SSM}

\begin{figure}[!ht]
    \centering
    \includegraphics[width=\linewidth]{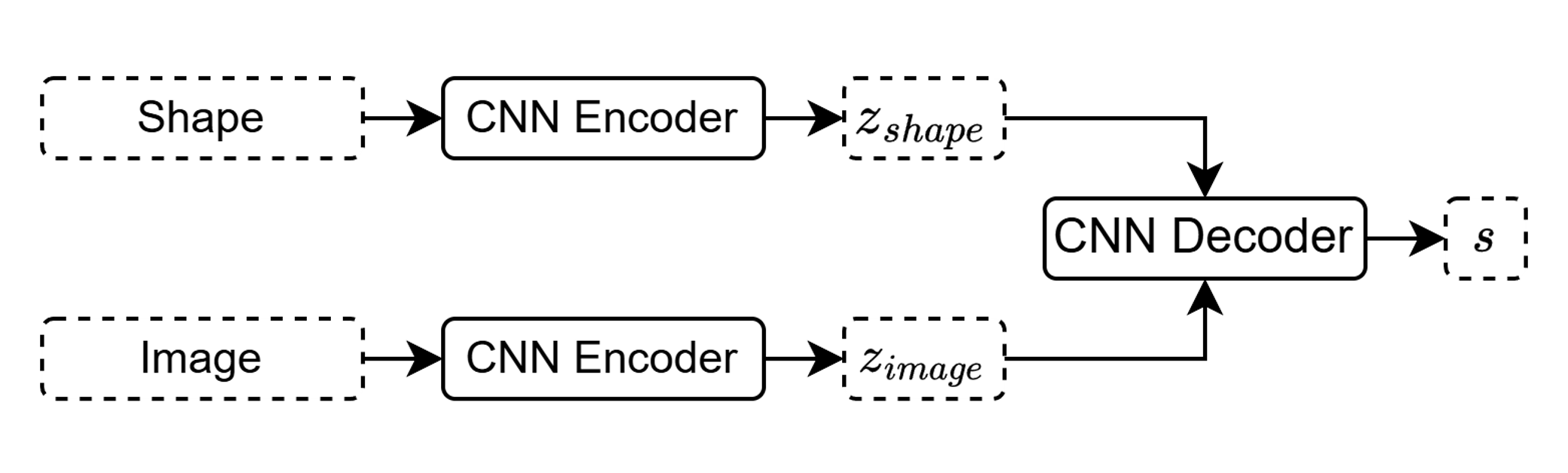}
    \caption{Schematic of Deep Learning-based Non-linear SSM.}
    \label{fig:ssm2}
\end{figure}

Deep learning-based non-linear SSM bypasses PCA by using neural networks to directly encode shapes into a non-linear latent space. As illustrated in Fig.~\ref{fig:ssm2}, shape and image data are processed by two separate CNN encoders, producing shape encoding $z_{shape}$ and image encoding $z_{image}$. These encodings are combined and decoded by a CNN decoder to reconstruct the final shape $s$. This method captures more complex, non-linear deformations and provides greater flexibility in modelling anatomical variability.

\cite{Riddhish2021DeepSSM} proposed DeepSSM, a deep learning framework that directly maps 3D images to low-dimensional shape descriptors, bypassing traditional manual preprocessing, and employs model-based data augmentation to address limited training data in shape modelling applications.

\cite{Aziz2023Progressive} presented Progressive DeepSSM, a training methodology for image-to-shape deep learning models. It employs a multi-scale progressive learning strategy that first learns coarse shape features and gradually progresses to more detailed features. The method incorporates segmentation-guided multi-task learning and deep supervision loss to ensure learning at each scale. The paper adopted a correspondence-based SSM approach by predicting correspondence points at different densities (starting from 256 points, progressively increasing to 512 and 1024 points) to represent shapes. 

The primary difference between linear and non-linear deep learning-based SSMs lies in their capacity to model shape variations. Linear models rely on PCA, limiting them to linear deformations, while non-linear models use neural networks to capture complex, non-linear shape deformations. Non-linear SSMs are more expressive and adaptable, making them suitable for advanced medical image analysis tasks that require highly flexible shape modelling.

\subsection{Hybrid SSM}

Some methods utilize deep learning in the reconstruction subprocess but do not directly apply it to the SSM itself. Instead, deep learning is employed to enhance other stages, such as segmentation, registration, or classification, which subsequently provides input to the SSM, rather than predicting the shape coefficients directly.

For example, \cite{Abhirup2022Automated} proposed a method for 3D whole-heart mesh reconstruction from limited 2D cardiac MR slices using an SSM. In their approach, deep learning is used to extract heart contours from the 2D slices. The SSM is then optimally fitted to these sparse heart contours in 3D space, generating an initial mesh representation. This initial mesh is further deformed to minimize the distance between the mesh and the heart contours, resulting in the final reconstructed shape.

\cite{Xiaohan20234D} used decoupled motion and shape models for 4D myocardium reconstruction, including a neural motion model for registration and an end-diastolic shape model for predicting sign distance function, pre-training the shape model with ED-space representation to guide motion model training, addressing data scarcity issues.

\cite{Sachin2023deep} used deep learning techniques for automated view classification, slice selection, phase selection, anatomical landmark location, and myocardial image segmentation to generate 3D biventricular cardiac shape models consistent with clinical workflows.
In this method, the SSM comprised of three integrated components: template modelling using a biventricular subdivision surface mesh, hybrid registration (affine transformation + diffeomorphic non-rigid registration + landmark registration) for shape fitting, and statistical analysis through PCA-based shape atlas construction and Z-score evaluation, with the first 20 PCA modes explaining $\sim$87\% of shape variations.

\cite{K2020Artificial} reviewed methods of combining statistical cardiac atlases with deep learning in three main approaches: direct use of shape parameters for machine learning, incorporation of shape priors into deep networks, and direct prediction of statistical shape parameters from images. The methods mentioned in this review are to build statistical atlases of cardiac anatomy and combine them with deep learning networks in three ways: direct use of shape parameters, adding shape constraints, or direct prediction of shape parameters.

\cite{Abhirup2021completely} used an SSM in the correction of misalignment during the reconstruction of the heart contour. This method corrects the misalignment of the cut through a three-step approach based on intensity, contours, and statistical shape models and finally generates 3D biventricular surface and volumetric meshes from the aligned contours.
This method highlights the integration of deep learning in other steps for reconstruction, leveraging segmentation or reconstruction to refine SSM inputs, rather than directly incorporating deep learning into the core SSM modelling process.

Combined with sign distance function representation, \cite{Xu2023Image2SSM} introduced Image2SSM, a deep learning-based statistical shape modelling approach that learns shape representations directly from medical image-segmentation pairs using radial basis functions (RBF). The method adaptively captures surface details and generates compact yet comprehensive shape representations without requiring the manual segmentation and correspondence optimization steps from traditional SSM workflows. The method first uses 3D medical images and their corresponding segmentations as training data, learning control points and their surface normals through a deep neural network. It then employs RBF interpolation to construct a continuous surface representation, while optimizing the network through multiple loss functions (surface loss, normal loss, correspondence loss, and sampling loss). Finally, it can directly predict PDMs from unsegmented images.

\section{Model Generative}

Generative models employ deep learning techniques to directly generate new shapes from data, particularly useful for synthesizing complex and diverse shapes. These methods are well-suited for unsupervised or semi-supervised learning, especially when data is scarce or shapes are intricate.

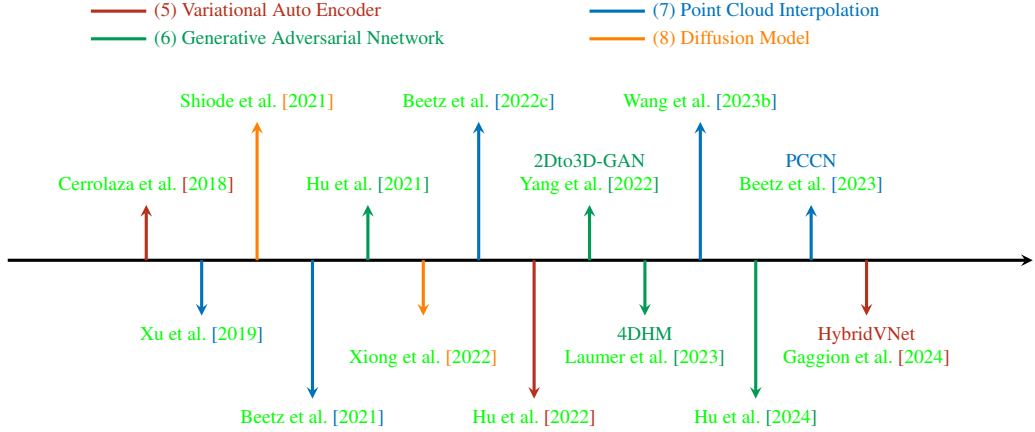
\begin{figure*}
    \centering
    \resizebox{0.8\textwidth}{!}{ 
    \begin{tikzpicture}
        \draw[ultra thick, -stealth, black] (-2.5,0) -- (16,0) node[right] {};

        \foreach \x/\year in {} {
            \fill[red] (\x,0) circle (3pt);
            \node[below] at (\x,-0.3) {\year}; 
        }

        \draw[BrickRed, ultra thick, -stealth] (0,0) -- (0,1) node[above] {\makecell[c]{\cite{cerrolaza20183d}} };

        \draw[RoyalBlue, ultra thick, -stealth] (1,0) -- (1,-1) node[below] {\makecell[c]{\cite{Hao2019Ventricle} } };

        \draw[orange, ultra thick, -stealth] (2,0) -- (2,2.5) node[above] {\makecell[c]{ \\ \cite{shiode20212d}}};
        
        \draw[RoyalBlue, ultra thick, -stealth] (3,0) -- (3,-2.5) node[below] {\makecell[c]{\cite{M2021Biventricular} } };
        
        \draw[ForestGreen, ultra thick, -stealth] (4,0) -- (4,1) node[above] {\makecell[c]{\cite{hu2021point}  } };

        \draw[orange, ultra thick, -stealth] (5,0) -- (5,-1) node[below] {\makecell[c]{ \\ \cite{xiong2022virtual}}};
        
        \draw[RoyalBlue, ultra thick, -stealth] (6,0) -- (6,2.5) node[above] {\makecell[c]{\cite{beetz2022point2mesh}  } };
        
        \draw[BrickRed, ultra thick, -stealth] (7,0) -- (7,-2.5) node[below] {\makecell[c]{\cite{hu2022srt} } };
        
        \draw[ForestGreen, ultra thick, -stealth] (8,0) -- (8,1) node[above] {\makecell[c]{ 2Dto3D-GAN \\ \cite{yang2022generative} }};

        \draw[ForestGreen, ultra thick, -stealth] (9,0) -- (9,-1) node[below] {\makecell[c]{ 4DHM \\ \cite{F2022Weakly} } };

        \draw[RoyalBlue, ultra thick, -stealth] (10,0) -- (10,2.5) node[above] {\makecell[c]{\cite{Zijie2023Shape} } };
        
        \draw[ForestGreen, ultra thick, -stealth] (11,0) -- (11,-2.5) node[below] {\makecell[c]{\cite{hu20233} } };
        
        \draw[RoyalBlue, ultra thick, -stealth] (12,0) -- (12,1) node[above] {\makecell[c]{PCCN \\ \cite{Beetz2023Multi}} };
        
        \draw[BrickRed, ultra thick, -stealth] (13,0) -- (13,-1) node[below] {\makecell[c]{HybridVNet \\ \cite{Gaggion2023Multi}} };

        \begin{scope}[shift={(-1,4.5)}]
            \draw[ultra thick, BrickRed] (0,0) -- (1,0) node[right] {(5) Variational Auto Encoder};
            \draw[ultra thick, RoyalBlue] (9,0) -- (10,0) node[right] {(7) Point Cloud Interpolation};
        \end{scope}

        \begin{scope}[shift={(-1,4)}]
            \draw[ultra thick, ForestGreen] (0,0) -- (1,0) node[right] {(6) Generative Adversarial Nnetwork};
            \draw[ultra thick, orange] (9,0) -- (10,0) node[right] {(8) Diffusion Model};
        \end{scope}

    \end{tikzpicture}
    } 
    \caption{\textcolor{black}{Chronological} overview of representative deep learning-based generative models for medical image-to-mesh reconstruction.}
\end{figure*}

\begin{table*}[!ht]
\centering
\caption{Generative Models for Medical Image-to-Mesh Reconstruction. 
}
\resizebox{\textwidth}{!}{%
\begin{tabular}{>{\centering\arraybackslash}m{4cm} >{\centering\arraybackslash}m{4cm} >{\centering\arraybackslash}m{3cm} >{\centering\arraybackslash}m{3cm} >{\centering\arraybackslash}m{3cm} >{\centering\arraybackslash}m{2cm} >{\centering\arraybackslash}m{6cm}}

\hline \hline

\textbf{Name} & \textbf{Method}& \textbf{Input} & \textbf{Output}& \textbf{Anatomy}& \textbf{Modality} & \textbf{Dataset} \\
\hline \hline
\makecell[c]{HybridVNet \\ \cite{Gaggion2023Multi}} & VAE & Multi-view image & Volume Mesh & Heart & CMR & UK Biobank \\ \hline
\makecell[c]{4DHM \\ \cite{F2022Weakly}}& GAN & 2D videos & Dynamic Surface Mesh & Heart & US ECGI & EchoNet-Dynamic \\ \hline

\makecell[c]{ \\ \cite{hu2021point}} & GAN & 2D Image & Point Cloud & Brain & MR & Private Data (317 AD and 723 Healthy) \\ \hline

\makecell[c]{\cite{hu20233}} & GAN & Incomplete 2D Image & Point Cloud & Brain & MR & Private Data (317 AD and 723 Healthy) \\ \hline

\makecell[c]{SRT \\ \cite{hu2022srt}} & VAE & 2D Image & Point Cloud & Brain & MR & Private Data (317 AD and 723 Healthy) \\ \hline

\makecell[c]{REC-CVAE \\ \cite{cerrolaza20183d}} & VAE & 2D Image & Surface Mesh & Brain Skull & US & Private Data (72) \\ \hline

\makecell[c]{ \\ \cite{Hao2019Ventricle}} & Point Interpolation & 2D Image & Surface mesh & Heart & MR & Cardiac SSM (1093) \\ \hline
\makecell[c]{GCN-Shape \\ \cite{Zijie2023Shape}} & Point Interpolation & Partial mesh & Complete Mesh & Liver, Stomach & CT & Private Data (124 pancreatic tumor) \\ \hline

\makecell[c]{\cite{M2021Biventricular}} & Point Interpolation & 2D Contours & Surface Mesh & \makecell[c]{Heart \\ (LV, RV)} & MR & UK Biobank, Cardiac SSM \\ \hline
\makecell[c]{\cite{beetz2022point2mesh}} & Point Interpolation & 3D Image & Surface Mesh & \makecell[c]{Heart \\ (LV)} & CMR & UK Biobank \\ \hline

\makecell[c]{PCCN \\ \cite{Beetz2023Multi}} & Point Interpolation & 3D Image & Surface Mesh & \makecell[c]{Heart \\ (LV, RV)} & CMR & Cardiac SSM, UK Biobank \\ \hline

\makecell[c]{ \\ \cite{Zijie2023Shape}}& Point Interpolation & 2D Image & Surface Mesh & Spine & CT & Private Data (1012) \\ \hline

\makecell[c]{ 2Dto3D-GAN \\ \cite{yang2022generative}}& GAN & 3D Image & Surface Mesh & Liver & CT & Private Data (124) \\ \hline

\makecell[c]{ \\ \cite{xiong2022virtual}} & GAN & 3D Image & Surface Mesh & Skull Midface & CT & Private Data (518) \\ \hline

\makecell[c]{ \\ \cite{shiode20212d}} & GAN & 2D Image & Surface Mesh & Distal Forearm Bone & X-ray & Private Data (33) \\ \hline

\makecell[c]{ \\ \cite{wodzinski2024improving}}& Diffusion & Defective Mask & Reconstructed Mask & Skull & CT & SkullFix, SkullBreak, MUG500+\\ \hline

\makecell[c]{ DMCVR \\\cite{he2023dmcvr}}& Diffusion & Sparse 2D Images & Dense 2D Images then Mesh & Heart & MR & UK Biobank \\ \hline

\makecell[c]{DISPR \\ \cite{waibel2023diffusion}}& Diffusion & Single 2D Image & 3D Voxel Grid then Mesh & Cell & Microscopy & SHAPR \\ \hline

\end{tabular}%
}

\end{table*} 

\subsection{VAE-based Generative Models}

\begin{figure}[!ht]
    \centering
    \includegraphics[width=\linewidth]{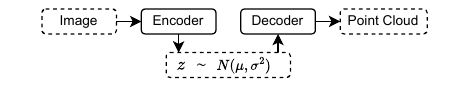}
    \caption{Schematic representation of a variational autoencoder (VAE)-based model for mesh generation. The input image is processed by the encoder to compute the mean and variance, which are used to sample a latent vector $z \sim \mathcal{N}(\mu, \sigma^2)$. The latent vector is then passed through the decoder to generate a point cloud, which is subsequently converted into a mesh.}
    \label{fig:DM5}
\end{figure}

Variational-based models have emerged as a powerful framework in generative modelling. Unlike deterministic approaches that directly map inputs to outputs, these models capture the underlying probability distributions of the data. By introducing a probabilistic perspective, variational-based models learn to parameterize distributions, typically using parameters like mean ($\mu$) and variance ($\sigma$), rather than predicting fixed values. This probabilistic nature allows them to model uncertainty and generate diverse outputs by sampling from the learned distributions. The optimization of these models often involves minimizing the Kullback-Leibler divergence between the learned and target distributions, along with task-specific reconstruction losses. This approach has proven particularly effective for complex generative tasks where capturing the full range of possible outputs is essential.

\cite{Gaggion2023Multi} developed a novel deep learning architecture called HybridVNet, combining convolutional neural networks and graph convolutions, using a multi-view approach to process cardiac MR images and directly extract 3D meshes from medical images.  

\cite{hu2022srt} proposed a variational-based transformer framework for shape reconstruction (SRT) that reconstructs 3D point cloud representations from a single 2D brain MR image. The model adopts an encoder-decoder architecture, where the encoder uses a Multi-headed Attention Module (MAM) to encode the input image into a feature vector that follows a normal distribution, and the decoder employs a unique ``up-down-up" generation system to reconstruct point clouds with 2048 points. The main innovation lies in completely avoiding CNNs and GANs, relying solely on transformer architecture and fully-connected networks, which ensures reconstruction accuracy and addresses issues like training instability and poor real-time performance that exist in GAN-based methods.

\cite{cerrolaza20183d} presented a novel approach for 3D fetal skull reconstruction from 2D ultrasound scans. It takes multiple standard 2D ultrasound planes (axial, sagittal, and coronal views) as input and produces a 3D reconstruction of the fetal skull as output. The process involves two proposed deep conditional generative networks: REC-CVAE, which uses a conditional variational autoencoder framework, and HiREC-CVAE, which implements a hierarchical structure based on the clinical relevance of each view. The main innovation lies in its hierarchical approach that can operate effectively even with missing views.

Both \cite{Gaggion2023Multi} and \cite{cerrolaza20183d} addressed multi-input VAE problems but employed different strategies: the former adopted a parallel approach, directly combining all $\mu$ and $\sigma$ generated from separate inputs to produce the latent variable $z$; the latter used a cascading approach, progressively connecting two sets of $\mu$ and $\sigma$ to generate $z$, then deriving new $\mu$ and $\sigma$ from $z$ to connect with the next input's $\mu$ and $\sigma$, thereby constructing a hierarchical information structure step by step.

\subsection{GAN-based Generative Models}

\begin{figure}[!ht]
    \centering
    \includegraphics[width=\linewidth]{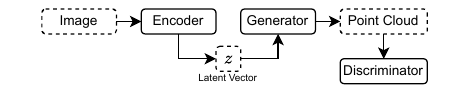}
    \caption{Schematic representation of a generative adversarial network (GAN)-based model for mesh generation. The input image is processed by the encoder to produce a latent vector. The latent vector is then passed to the generator, which outputs a point cloud. The generated point cloud is subsequently evaluated by a discriminator and converted into a mesh. }
    \label{fig:DM6}
\end{figure}

\cite{F2022Weakly} developed an automated framework using a self-supervised learning approach to infer personalized 3D+time heart meshes from 2D echocardiography videos, generating 4D heart meshes closely corresponding to input videos without requiring paired data.

Instead of using CNN to get an incomplete point contour in Fig.~\ref{fig:DM3_interpolation}, \cite{hu20233, hu2021point} used a GAN predictor to generate initial incomplete point clouds, then used a hierarchical encoder-decoder architecture and attention mechanisms as a completion network to refine and reconstruct the point clouds. It specifically designs a branching GNN generator and attention gate blocks to improve reconstruction quality.

\cite{hu2021point} proposed a GAN-based 3D brain shape reconstruction model that takes a single 2D brain MR image as input and outputs a 3D point cloud representation consisting of 2048 points. The main innovation lies in proposing a tree-structured graph convolutional generative network, which encodes the 2D image into a latent vector through an encoder, and then utilizes alternating branching block and graph convolution block to construct a tree-structured feature transmission mechanism, achieving the generation from a single vector to complex 3D point clouds. This method was the first to apply point cloud generation to brain shape reconstruction, outperforming the existing method PointOutNet \cite{fan2017point} in both qualitative and quantitative evaluations. The method is particularly suitable for 3D shape visualization needs in minimally invasive surgery and robot-assisted surgery.

\subsection{Interpolation-based Methods} 

In generative models, point cloud interpolation (upsampling/completion) models \cite{yu2018pu, qi2017pointnet} achieve mesh reconstruction by completing missing parts of the structure. In this process, the input image is passed through a CNN to extract features and generate the target contour or sparse point cloud. An interpolation network then densifies the sparse point cloud, ultimately producing the final mesh. This approach is beneficial for handling occluded or incomplete medical images, effectively reconstructing missing structural information. Interpolation-based methods emphasize data-driven completion capabilities, allowing them to generate complete and detailed meshes even in challenging imaging environments.

\begin{figure}[!ht]
    \centering
    \includegraphics[width=\linewidth]{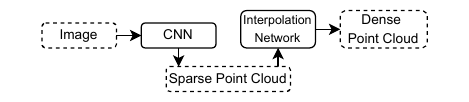}
    \caption{Schematic representation of a interpolation-based method for mesh generation. The input image is processed by a CNN to generate a sparse point cloud. This sparse point cloud is then refined by a completion network to produce a dense point cloud, which is subsequently converted into a mesh.}
    \label{fig:DM3_interpolation}
\end{figure}

Based on Fig.~\ref{fig:DM3_interpolation}, \cite{beetz2022point2mesh} proposed Point2Mesh-Net, an innovative geometric deep learning approach that transforms 2D MR slices directly into 3D cardiac surface meshes. Its key innovations have two parts. First, they used dual architecture combining a point cloud-based encoder and mesh-based decoder, which enables the network to directly process sparse MR contour point clouds from image segmentation while outputting 3D triangular surface meshes suitable for downstream tasks. Second, they used a hierarchical design with multi-scale downsampling and upsampling that enables effective feature learning and successfully addresses the two main challenges in cardiac surface reconstruction - data sparsity and slice misalignment. 

Based on their previous work \cite{beetz2022point2mesh}, \cite{Beetz2023Multi} further developed a novel multi-class Point Cloud Completion Network (PCCN) for reconstructing 3D cardiac anatomy from cardiac magnetic resonance images. Unlike previous completion methods, this approach uniquely addresses both sparsity and misalignment challenges within a single unified model. Furthermore, the network maintains both multi-class anatomical information (left ventricular cavity, left ventricular myocardium, right ventricular cavity) and bi-temporal cardiac phase data (end-diastolic and end-systolic) throughout the reconstruction process. This integrated design represents a significant advancement in cardiac anatomy reconstruction from medical imaging data.

\cite{M2021Biventricular} also proposed a deep learning method based on point completion networks to reconstruct dense 3D biventricular heart models from misaligned 2D cardiac MR contours. The method addresses both data sparsity and slice misalignment problems simultaneously.
The method first segments 2D MR images to obtain contours, converts them to 3D point clouds, then reconstructs through a point completion network consisting of two PointNet \cite{qi2017pointnet} encoder layers and a decoder with a FoldingNet \cite{yang2018foldingnet} block, finally generating meshes using the ball pivoting algorithm \cite{bernardini2002ball}.

\cite{Hao2019Ventricle} also developed a completion network, but it performs completion at the voxel level rather than at the point cloud level.
This method uses deep learning algorithms with a volumetric mapping approach to reconstruct 3D ventricular surfaces from 2D cardiac MR data and is capable of processing contours from multiple orientations and correcting motion artifacts.
This approach transforms the mesh fitting problem into a volumetric mapping problem, using 3D U-Net to generate dense volumetric predictions from sparse input, followed by isosurfacing to obtain the final 3D surface.
The method consists of three main steps: 1) generating sparse volumetric input data from contours; 2) generating dense 3D volumetric predictions of LV myocardium and LV/RV cavities using 3D U-Net; 3) generating 3D meshes from predictions using an isosurfacing algorithm.

Based on Fig.~\ref{fig:DM3_interpolation}, \cite{Zijie2023Shape} used a GNN for the completion network. Additionally, the input for this method was not a contour but two surface meshes, one from the template mesh and the other from the mesh of undetectable regions in the medical image. These two meshes are augmented with internal vertices based on the surface before being fed into the completion network.
Specifically, this is a GCN-based approach for reconstructing undetectable regions of organs in medical images. The method transforms an initial template into patient-specific organ shapes using prior information obtained from detectable organs across different patients. To address inaccurate estimation results of GNN with triangular surface meshes, the authors introduced a method of adding internal vertices and edges, improving calculation accuracy by enhancing the mesh topology.

\subsection{Diffusion-based Generative Model}

\begin{figure}[!ht]
    \centering
    \includegraphics[width=\linewidth]{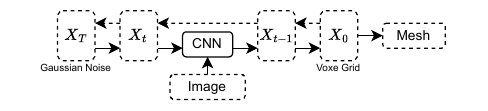}
    \caption{Schematic representation of a diffusion model for medical surface reconstruction. }
    \label{fig:DM3}
\end{figure}

Diffusion models \cite{ho2020denoising, kazerouni2023diffusion} are a class of generative models that learn to generate data through a two-step process: a forward process and a backward process. In the forward process, structured data is gradually transformed into noise by iteratively adding small amounts of Gaussian noise over multiple timesteps. The backward process, also known as denoising, aims to recover the original data by learning to iteratively remove the noise in a controlled manner. This is typically achieved using deep neural networks such as U-Net \cite{ronneberger2015u}, which predict and remove noise at each timestep to reconstruct the desired output.

One of the key advantages of diffusion models is their ability to handle reconstructions with missing information. They are particularly effective in tasks such as reconstructing 3D shapes from a single 2D image or recovering complete 3D structures from sparse 3D volumes, where information is inherently incomplete. However, diffusion models introduce diversity in the generated outputs, which can sometimes conflict with the fidelity required for high-precision reconstruction. When input information is sufficiently complete, such as in the case of converting a full 3D volume into a 3D mesh, diffusion models are rarely used. In this scenario, deterministic methods provide more accurate and reliable results.

Even for incomplete input to complete output reconstruction, there is currently no diffusion model designed for direct representation transfer in an end-to-end image-to-mesh pipeline. Instead, existing approaches leverage diffusion models to enhance image quality before applying segmentation or the marching cubes algorithm to generate the final mesh.

\cite{waibel2023diffusion} introduced DISPR, a diffusion-based model for reconstructing 3D cell shapes from 2D microscopy images. Instead of directly transferring representations from images to meshes, the model leverages diffusion processes to generate high-quality 3D volume reconstructions. DISPR is conditioned on the input 2D microscopy image and progressively refines noisy 3D volumes through multiple denoising steps, ensuring realistic shape predictions. The generated 3D structures are then evaluated using morphological features such as volume, surface area, and curvature, demonstrating superior performance over deterministic shape reconstruction methods. The model also serves as a data augmentation tool by generating additional synthetic training samples, improving the classification accuracy of single-cell morphological analysis.

\cite{he2023dmcvr} proposed DMCVR, a framework that utilizes a morphology-guided diffusion model for 3D cardiac volume reconstruction from sparse 2D cine MRI slices. Instead of directly generating meshes, the method enhances the axial resolution of 2D MR images by employing a diffusion-based generative process conditioned on both global semantic and regional morphology latent codes. The global semantic encoder captures high-level anatomical features, while the regional morphology encoder focuses on specific cardiac structures (left ventricle cavity, left ventricle myocardium, and right ventricle cavity) to improve anatomical accuracy. \cite{song2020denoising} used denoising diffusion implicit models to iteratively refine reconstructed slices, which are then stacked to form a complete 3D volume. By incorporating latent-space interpolation techniques, DMCVR ensures smooth and high-resolution 3D reconstructions, outperforming traditional interpolation and GAN-based method \cite{chang2022deeprecon} in cardiac volume generation.

\cite{wodzinski2024improving} applied latent diffusion models (LDMs) proposed by \cite{rombach2022high} to cranial defect surface reconstruction by using them as a data augmentation technique to improve deep learning-based reconstruction models. Instead of directly using diffusion models for end-to-end surface reconstruction, the approach leverages LDMs in the latent space to generate synthetic defective skulls, which are then used to enhance the training dataset for cranial defect reconstruction networks. The latent diffusion model refines the variational autoencoder (VAE) \cite{kingma2013auto} or vector quantized VAE \cite{van2017neural} representations by incrementally adding and denoising noise in the latent space, thereby producing more diverse and realistic cranial defect samples. This augmentation significantly improves the generalizability of the reconstruction network, enabling it to reconstruct real clinical cranial defects despite being trained solely on synthetic datasets.

\section{Implicit Models}

\begin{figure*}
    \centering
    \resizebox{0.8\textwidth}{!}{ 
    \begin{tikzpicture}
        \draw[ultra thick, -stealth, black] (-1,0) -- (14,0) node[right] {};

        \foreach \x/\year in {2.5/2022, 4.5/2023, 10.5/2024} {
            \fill[red] (\x,0) circle (3pt);
            \node[below] at (\x,-0.3) {\year}; 
        }

        \draw[ForestGreen, ultra thick, -stealth] (0,0) -- (0,1) node[above] {\makecell[c]{DeepCSR \\\cite{Rodrigo2020DeepCSR}}}; 
        \draw[BrickRed, ultra thick, -stealth] (1,0) -- (1,-1) node[below] {\makecell[c]{Vox2Surf \\ \cite{hong2021vox2surf}}}; 
        \draw[RoyalBlue, ultra thick, -stealth] (2,0) -- (2,2.5) node[above] {\makecell[c]{\cite{lebrat2021corticalflow} }}; 
        \draw[orange, ultra thick, -stealth] (3,0) -- (3,-2.5) node[below] {\makecell[c]{\cite{Sheng2022Development}}}; 
        \draw[RoyalBlue, ultra thick, -stealth] (4,0) -- (4,1) node[above] {\makecell[c]{\cite{Shanlin2022Topology} }}; 
        \draw[RoyalBlue, ultra thick, -stealth] (5,0) -- (5,-1) node[below] {\makecell[c]{CortexODE \\ \cite{Q2022CortexODE} }}; 
        \draw[RoyalBlue, ultra thick, -stealth] (6,0) -- (6,2.5) node[above] {\makecell[c]{NDM \\ \cite{Meng2023Neural} }}; 
        \draw[BrickRed, ultra thick, -stealth] (7,0) -- (7,-2.5) node[below] {\makecell[c]{SurfNN \\ \cite{Hao2023SurfNN} }}; 
        \draw[BrickRed, ultra thick, -stealth] (8,0) -- (8,1) node[above] {\makecell[c]{LightNeuS \\ \cite{Batlle2023LightNeuS}}}; 
        \draw[BrickRed, ultra thick, -stealth] (9,0) -- (9,-1) node[below] {\makecell[c]{Image2SSM \\ \cite{Xu2023Image2SSM} }}; 
        \draw[BrickRed, ultra thick, -stealth] (10,0) -- (10,2.5) node[above] {\makecell[c]{NIR-3DUS \\ \cite{Hongbo2023Neural} }}; 
        \draw[RoyalBlue, ultra thick, -stealth] (11,0) -- (11,-2.5) node[below] {\makecell[c]{V2C-Flow \\ \cite{Fabian2024Neural}}}; 
        \draw[ForestGreen, ultra thick, -stealth] (12,0) -- (12,1) node[above] {\makecell[c]{X2V \\ \cite{guven2024x2v} }}; 
        \draw[BrickRed, ultra thick, -stealth] (13,0) -- (13,-1) node[below] {\makecell[c]{FUNSR \\ \cite{Hongbo2024Neural} }};

        \begin{scope}[shift={(-1,5)}]
            \draw[ultra thick, BrickRed] (0,0) -- (1,0) node[right] {(9) Implicit Model - SDF};
            \draw[ultra thick, ForestGreen] (8,0) -- (9,0) node[right] {(10) Implicit Model - Occupancy Function};
        \end{scope}

        \begin{scope}[shift={(-1,4.5)}]
            \draw[ultra thick, RoyalBlue] (0,0) -- (1,0) node[right] {(11) Implicit Model - Neural ODE};
            \draw[ultra thick, orange] (8,0) -- (9,0) node[right] {(12) Implicit Model - NeRF};
        \end{scope}

    \end{tikzpicture}
    } 
    \caption{\textcolor{black}{Chronological} overview of representative deep learning-based implicit models for medical image-to-mesh reconstruction.}
    \label{table:implicit model}
\end{figure*}
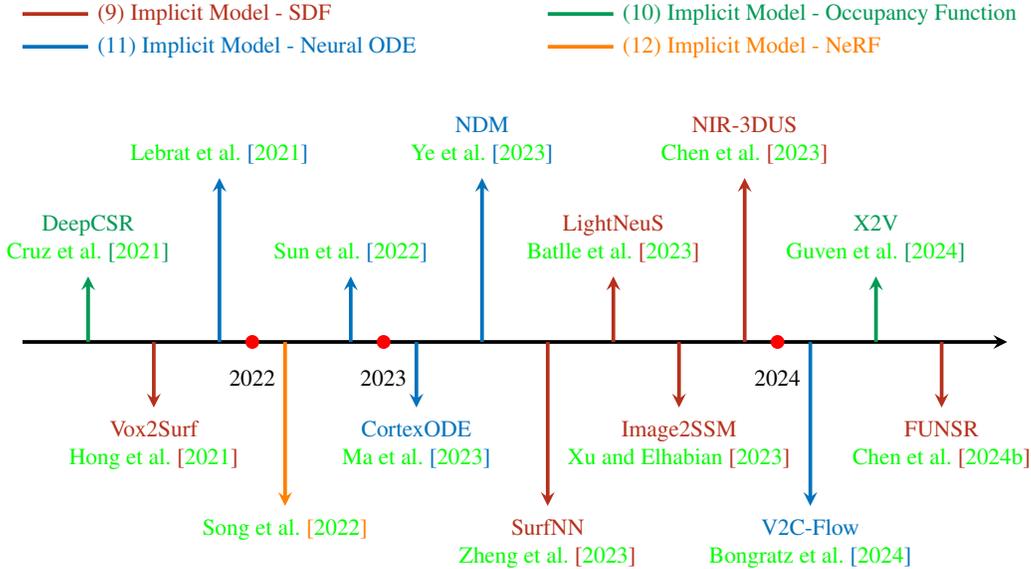

Implicit models are a class of methods that use neural networks to learn continuous functions for representing 3D shapes. Unlike traditional explicit methods such as voxel grids or point clouds, implicit models do not directly store geometric information but instead leverage multilayer perceptrons (MLPs) \cite{pinkus2020multilayer} to learn an implicit function value for each spatial point, thereby defining the 3D structure indirectly. Common types of implicit models summarized in Table \ref{table:implicit model} include signed distance functions (SDF), neural ordinary differential equations (Neural ODE), occupancy functions, and neural radiance fields (NeRF). SDF models represent shapes by learning the signed distance from a point to the surface, Neural ODEs model dynamic structures by learning the temporal evolution of shapes, occupancy function models predict whether a point belongs to an object’s interior using binary classification, and NeRF incorporates density and colour information to enable both geometric reconstruction and rendering.  

The advantage of implicit models lies in their high-resolution and continuous representation capabilities, making them independent of fixed-resolution constraints and highly efficient for handling complex geometries. This makes them particularly valuable in fields such as medical image reconstruction, computer vision, and computer graphics, especially for tasks requiring precise surface reconstruction, such as mesh generation from medical scans, scene reconstruction, and shape optimization based on physical simulations.

\begin{table*}[!ht]
\centering
\caption{Implicit Models for Medical Image-to-Mesh Reconstruction}
\resizebox{\textwidth}{!}{%
\begin{tabular}{>{\centering\arraybackslash}m{4cm} >{\centering\arraybackslash}m{4cm} >{\centering\arraybackslash}m{3cm} >{\centering\arraybackslash}m{3cm} >{\centering\arraybackslash}m{3cm} >{\centering\arraybackslash}m{2cm} >{\centering\arraybackslash}m{6cm}}
\hline
\hline
\textbf{Name} & \textbf{Method}& \textbf{Input} & \textbf{Output}& \textbf{Anatomy} & \textbf{Modality}& \textbf{Dataset} \\
\hline \hline

\makecell[c]{CortexODE \\ \cite{Q2022CortexODE}} & Neural ODE & 3D Image & Surface Meshs & Brain Cortex & MR-T1w/T2w & dHCP (874), HCP (600), ADNI (524) \\ \hline

\makecell[c]{FUNSR \\ \cite{Hongbo2024Neural}} & SDF & Segmentation masks & Surface mesh (SDFs) & Hip, carotid artery, Prostate & 3D Freehand Ultrasound & Hip phantom (2), CCA (10), CAB (77), Prostate (73) \\ \hline

\makecell[c]{Vox2Surf \\ \cite{hong2021vox2surf}} & SDF & T1w and T2w MR & Surface meshes & Brain & MR & HCP (22) \\ \hline

\makecell[c]{LightNeuS \\ \cite{Batlle2023LightNeuS}} & SDF & Monocular endoscopic video frames & Watertight Surface Mesh & Colon & Endoscopy & C3VD dataset (22) \\ \hline

\makecell[c]{NIR-3DUS \\ \cite{Hongbo2023Neural}} & UDF & 3D images & Surface mesh (MAB, plaque, LIB) & Carotid artery & 3D Ultrasound & simulated volumes (6),  Private data (5) \\ \hline

\makecell[c]{SurfNN \\ \cite{Hao2023SurfNN}} & SDF & 3D MR + mesh) & WM + pial + Midthickness surface & Brain cortical surfaces & T1 MR & ADNI (747/70/113 train/val/test) \\ \hline

\makecell[c]{DeepCSR \\ \cite{Rodrigo2020DeepCSR}} & Occupancy Function & T1w MR & Inner and outer cortical surfaces & Brain cortex & MR & ADNI (3876 MRIs), Test-Retest (120 MRIs), MALC (30 MRIs) \\ \hline

\makecell[c]{X2V \\ \cite{guven2024x2v}} & Occupancy Function & Single X-ray DRR & 3D Mesh & Lung & X-ray & NLST (6,392 CT scans from 2,560 subjects) \\ \hline

\makecell[c]{Image2SSM \\ \cite{Xu2023Image2SSM}} & SDF & Image-Segmentation pairs & PDMs/Surface mesh & Femur/Left atrium & CT/MR & 50 proximal femur CT scans \\ \hline

\makecell[c]{NDM \\ \cite{Meng2023Neural}} & Neural ODE & Sparse point cloud (2D CMR slices) & Dense mesh with deformations & Bi-ventricular heart& CMR & Cardiac SSM \\  \hline 

\makecell[c]{V2C-Flow \\ \cite{Fabian2024Neural}} & Neural ODE & 3D Image & Surfaces Mesh & Brain Cortex & MRI-T1w & ADNI, J-ADNI, OASIS, Mindboggle, TRT, JHU \\  \hline
\makecell[c]{CorticalFlow \\ \cite{lebrat2021corticalflow}} & Neural ODE & 3D Image & Surface Mesh & Brain Cortex & MR & ADNI \\ \hline

\end{tabular}%
}
\end{table*}

\subsection{Neural ODE} 

\begin{figure}[!ht]
    \centering
    \includegraphics[width=\linewidth]{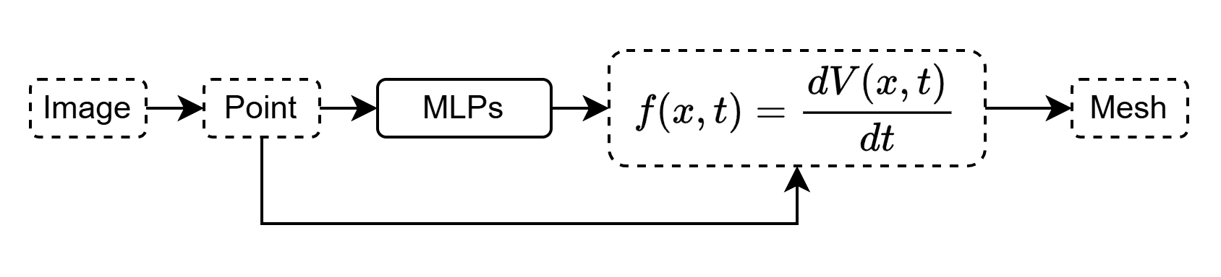}
    \caption{Schematic of Neural ODE-based methods. In Neural ODEs, \( {x} \) represents a point in 3D space, \( t \) is the time parameter that controls the dynamic evolution, \( {V}({x}, t) \) denotes the state of the point \( {x} \) at time \( t \) (such as its deformation during surface reconstruction), and \( f({x}, t) \) describes the dynamic system governing how the state evolves over time.
}
    \label{fig:Implicit Neural ODE}
\end{figure}

Neural ordinary differential equation (Neural ODE) methods are a variant of GCN-based approaches that model the deformation process as a continuous dynamic system. In this framework, features are first extracted from the input image using CNN or MLP, and a neural ODE governs the deformation of a template mesh. The neural ODE evolves the template over time to fit the target shape, described by:
\[
\frac{d {V}({x}, t)}{dt} = f({x}, t),
\]
where \( {V}({x}, t) \) represents the state of a vertex \( {x} \) at time \( t \), and \( f({x}, t) \) is the velocity field driving the deformation. This method produces smooth deformation trajectories, ensuring mesh stability and physical plausibility. Additionally, it provides not only the deformed mesh but also the velocity field or dynamic deformation information, making it particularly suitable for tasks requiring continuous and nonlinear deformations.

In contrast, the Deformation Field method applies instantaneous displacement directly to the mesh vertices, described as:
\[
{x}' = {x} + D({x}),
\]
where \( D({x}) \) is the displacement vector at point \( {x} \). While simpler and more efficient, this approach is better suited for small or localized deformations due to its discrete nature.

The primary distinction between the two methods lies in their treatment of deformation: neural ODEs model deformation as a time-dependent, continuous process, whereas the Deformation Field represents deformation as an instantaneous, discrete operation. This makes neural ODEs particularly advantageous for scenarios requiring smooth, large-scale deformations, while deformation fields excel in cases where efficiency and simplicity are priorities.

\subsection{Diffeomorphic flow-based deformations}

Based on Fig.~\ref{fig:Implicit Neural ODE}, \cite{lebrat2021corticalflow} used a differential mesh deformation module to replace the deformation GCN. 
This method introduces a novel geometric deep-learning model that utilizes a flow ordinary differential equation framework to learn a series of diffeomorphic transformations, deforming a reference template mesh towards a target object, thereby achieving efficient cortical surface reconstruction

\textbf{NODE}
\cite{Shanlin2022Topology} proposed a new model called Neural Diffeomorphic Flow (NDF), using a generative model to represent 3D shapes as deformations of implicit shape templates, preserving the topological structure of shapes, and employing an autodecoder with Neural Ordinary Differential Equation (NODE) blocks to realize diffeomorphic deformation.

Based on Fig.~\ref{fig:Implicit Neural ODE}, \cite{Meng2023Neural} used a multi-level structure to achieve coarse-to-fine mesh reconstruction.
This method proposes a neural deformable model (NDM) to reconstruct complete 3D bi-ventricular heart shapes from sparse cardiac magnetic resonance images. NDM combines traditional deformable superquadrics with modern deep learning techniques, adopting a two-stage strategy: first using parameterized functions for global deformation to capture coarse shape, then employing neural diffeomorphic flows for local deformation to recover details.

Based on Fig.~\ref{fig:Implicit Neural ODE}, \cite{Fabian2024Neural} upgraded the algorithm from \cite{Fabian2022Vox2Cortex}, using NODE to achieve cortical surface reconstruction, and further introduced separate templates for different regions.
This method (V2C-Flow) is an end-to-end deep learning method based on neural ODEs for cortical surface reconstruction from MRI. The method generates white matter and pial surfaces by deforming template meshes while maintaining vertex correspondences. V2C-Flow reconstructs all four cortical surfaces simultaneously, runs fast (1.6 seconds), and ensures geometric accuracy through a curvature-weighted Chamfer loss.

\cite{Q2022CortexODE} introduced CortexODE, a deep learning framework for cortical surface reconstruction. The method leverages neural ordinary differential equations (Neural ODEs) to learn a diffeomorphic flow that deforms an input surface into a target shape. The implementation first uses a 3D U-Net to predict white matter segmentation from brain MR scans, then generates a signed distance function and performs fast topology correction to ensure homeomorphism to a sphere. Following surface extraction, two CortexODE models are trained to deform the initial surface to white matter and pial surfaces respectively. The entire pipeline completes cortical surface reconstruction in less than 5 seconds while ensuring high geometric accuracy and minimal self-intersections.

\subsection{Signed Distance Function (SDF)}

\begin{figure}[!ht]
    \centering
    \includegraphics[width=\linewidth]{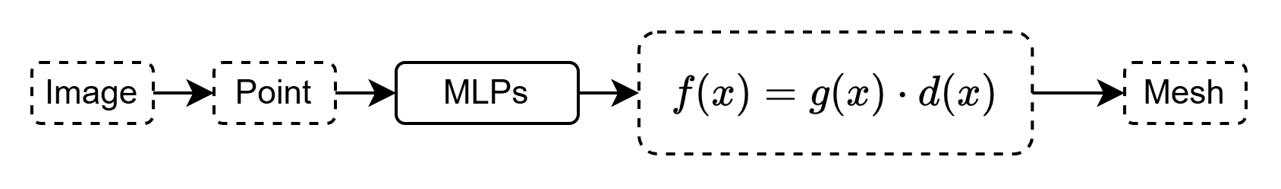}
    \caption{Schematic of SDF methods. In the signed distance function, \( {x} \) represents a point in 3D space, \( d({x}) \) is the distance from \( {x} \) to the target surface, and \( g({x}) \) indicates whether \( {x} \) is inside (negative sign) or outside (positive sign) the surface. The \( g({x}) \) is often determined based on the gradient of the distance function. Together, these elements describe the signed distance value of the point \( {x} \) with respect to the surface.
}a
    \label{fig:IM2}
\end{figure}

Signed distance function (SDF) is a fundamental implicit surface representation that defines a continuous field where each point in space is assigned a value representing the signed distance to the closest point on the surface. The sign indicates whether the point is inside (negative) or outside (positive) the surface, while the surface itself is represented by the zero-level set. Formally, for a point \( x \in \mathbb{R}^3 \), the SDF is defined as:

\begin{equation}
    f(x) = \text{sign}(x) \cdot d(x),
\end{equation}

where \( d(x) \) is the Euclidean distance from \( x \) to the closest point on the surface. The sign of \( f(x) \) can be determined based on the gradient of the distance function:
\begin{equation}
    \text{sign}(x) = \begin{cases}
        -1, & \text{if } \nabla d(x) \cdot \mathbf{n}(x) < 0, \\
        \phantom{-}1, & \text{if } \nabla d(x) \cdot \mathbf{n}(x) > 0,
    \end{cases}
\end{equation}
where \( \nabla d(x) \) is the gradient of the distance function at \( x \), and \( \mathbf{n}(x) \) is the unit normal vector pointing outward from the surface at the closest point. The zero-level set of the SDF, where \( f(x) = 0 \), defines the exact location of the surface. SDFs not only provide a compact and differentiable representation of geometry but also encode rich geometric information such as the distance to the surface, which can be utilized in various applications like surface reconstruction, simulation, and optimization tasks.

SDF provides several advantageous properties for mesh reconstruction. It offers a continuous and differentiable representation of the surface, enabling smooth optimization and gradient-based learning. The representation naturally handles topological changes and complex geometries without explicit mesh manipulation. Additionally, the gradient of the SDF directly gives the surface normal direction, facilitating the computation of geometric properties like surface curvature.
In medical image-to-mesh reconstruction, SDFs are particularly useful as they can represent complex anatomical structures while maintaining smoothness and allowing for topology changes. The implicit nature of SDFs makes them well-suited for deep learning frameworks, where networks can be trained to predict the signed distance field from medical images. 

\cite{Hongbo2024Neural} presented FUNSR, a self-supervised neural implicit surface reconstruction method for freehand 3D ultrasound data. The method directly transforms segmentation masks into volumetric point clouds as input and learns signed distance functions through neural networks to represent anatomical surfaces. Two geometric constraints are introduced: sign consistency constraint and surface constraint with adversarial learning, to improve reconstruction quality. The entire network is trained end-to-end in a self-supervised manner without requiring additional ground truth data or post-processing.

\cite{Hongbo2023Neural} presented a neural implicit representation approach for 3D ultrasound carotid surface reconstruction. The method first employs a U-Net for multi-class segmentation of ultrasound images to obtain vessel boundary point clouds, then uses a multi-layer perceptron network to learn unsigned distance functions (UDF) for reconstructing smooth and continuous vessel surfaces. Compared to the traditional iso-surface method \cite{remelli2020meshsdf}, this approach generates smoother and more continuous geometric surfaces.

\cite{hong2021vox2surf} presented Vox2Surf, a method for direct surface reconstruction of cortical and subcortical structures from volumetric MR data. The method learns a continuous-valued signed distance function (SDF) as implicit surface representation using deep learning, combining global and local features for surface reconstruction. The network can handle multiple brain structures simultaneously.

\cite{Batlle2023LightNeuS} presented LightNeuS, a neural network approach for endoscopic 3D reconstruction. The method is based on two key insights: first, endoluminal cavities are watertight, naturally enforced through signed distance function (SDF) representation; second, scene illumination varies and decays with inverse square distance from the endoscope light source to the surface. By explicitly incorporating illumination decline in the NeuS architecture and introducing a calibrated photometric model, the method achieves watertight reconstruction of complete colon sections with high accuracy on phantom data.

\cite{Hao2023SurfNN} introduced SurfNN for cortical surface reconstruction from MR images. Its novelty lies in simultaneously reconstructing both inner (white-gray matter boundary) and outer (pial) surfaces by training a single network to predict a midthickness surface that lies at the centre of both surfaces. The method takes as input a 3D MR and an initialization of the midthickness surface represented both as a distance map and triangular mesh, outputting the inner, outer, and midthickness surfaces. 

\subsection{NeRF}

\begin{figure}[!ht]
    \centering
    \includegraphics[width=\linewidth]{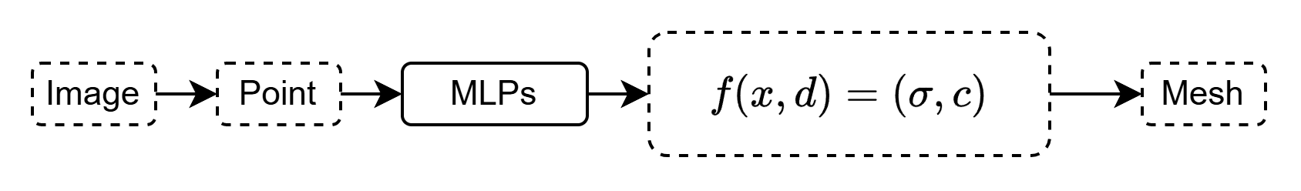}
    \caption{Schematic of NeRF methods. In neural radiance fields, \( x \) is a point in 3D space, and \( d \) is the viewing direction. The density \( \sigma \) represents the likelihood of the point \( x \) being part of the object’s surface, and the colour \( c \) (in RGB) provides information about the appearance of the surface at \( x \). For surface reconstruction, the surface geometry can be extracted as the iso-surface where the density \( \sigma \) exceeds a threshold.}
    \label{fig:DM2}
\end{figure}

Neural Radiance Field (NeRF) \cite{mildenhall2021nerf, velikova2024implicit, barron2021mip} represents 3D scenes by learning a continuous volumetric function using a neural network. While originally designed for novel view synthesis, NeRF has been adapted for medical image-to-mesh reconstruction tasks. The core idea of NeRF is to model a scene as a continuous function that maps every 3D spatial location \( x = (x, y, z) \) and viewing direction \( d = (\theta, \phi) \) to a volume density \( \sigma \) and RGB colour \( c \):
\begin{equation}
   F: (x, d) \rightarrow (\sigma, c).
\end{equation}

In medical applications, this formulation is typically modified to focus on surface reconstruction rather than radiance modelling. The viewing direction \( d \) is omitted, and the output is adapted to represent volumetric density for anatomical surface extraction:
\begin{equation}
   F: x \rightarrow \sigma.
\end{equation}
Here, \( x \) represents a point in 3D space, and \( \sigma \) represents the likelihood of \( x \) being part of the object's surface. For surface reconstruction tasks, the geometry of the surface can be extracted from the volumetric density representation using techniques like marching cubes or isosurface extraction.

The key advantage of NeRF-based approaches in medical imaging is their ability to handle multi-view consistency and incorporate 3D spatial context naturally. The continuous nature of the representation allows for high-resolution reconstruction and smooth interpolation between views. Additionally, the volumetric formulation is particularly well-suited for medical imaging modalities that inherently capture volumetric data, such as CT or MR scans. 

\cite{Sheng2022Development} presented a deep learning-based 3D ultrasound reconstruction algorithm to improve the quality of carotid artery volume reconstruction. The method uses a multi-layer perceptron (MLP) network to implicitly infer and represent both intensity and semantic probability of voxels in the image volume. By combining semantic segmentation information with volume intensity, the method produces smoother and more continuous vessel wall reconstructions compared to traditional VNN methods.

\subsection{Occupancy Function}

\begin{figure}[!ht]
    \centering
    \includegraphics[width=\linewidth]{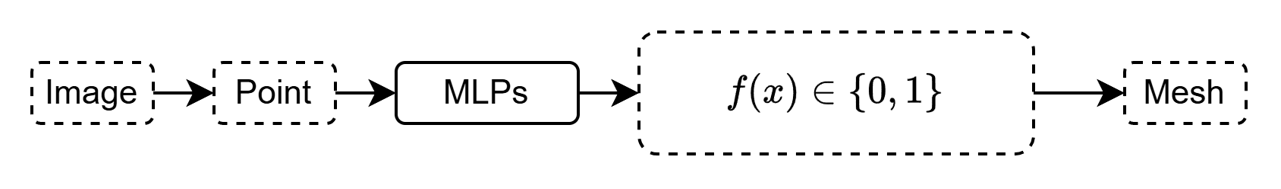}
    \caption{Schematic of occupancy function methods. In the occupancy function, \( {x} \) is a point in 3D space, and \( f({x}) \) outputs a binary value indicating whether \( {x} \) is inside or outside a target object. If \( f({x}) = 1 \), the point is inside the object; if \( f({x}) = 0 \), the point is outside.}
    \label{fig:IM4}
\end{figure}

The occupancy function provides another form of implicit surface representation where each point in space is assigned a binary value indicating whether it lies inside or outside the shape. Formally, for a surface \( \mathcal{S} \) that bounds a volume \( \mathcal{V} \), the occupancy function \( f(x) \) for a point \( x \) is defined as:
\begin{equation}
   f(x) \in \{0, 1\},
\end{equation}
where \( f(x) = 1 \) if the point \( x \) lies inside the volume \( \mathcal{V} \), and \( f(x) = 0 \) otherwise.

In practice, deep learning approaches often approximate this binary occupancy function with a continuous field that outputs values in the range \([0, 1]\), representing the probability of a point being inside the shape. The final surface is then extracted as the isosurface where \( f(x) = 0.5 \). Compared to Signed Distance Functions (SDFs), occupancy functions are simpler to represent but provide less geometric information since they do not encode distance to the surface. However, they are particularly effective for binary segmentation tasks and can be more efficient to train as they focus on inside/outside classification rather than on precise distance values.

\cite{Rodrigo2020DeepCSR} introduced DeepCSR for cortical surface reconstruction. The method first aligns input MR images with a brain template, then uses a neural network with hypercolumn features to predict occupancy representations for points in a continuous coordinate system. The cortical surface at a desired level of detail is obtained by evaluating surface representations at specific coordinates, followed by topology correction and isosurface extraction. Through its continuous nature and effective hypercolumn feature scheme, the method efficiently reconstructs high-resolution cortical surfaces capturing fine details in cortical folding.

\cite{guven2024x2v} presented X2V, an innovative approach for 3D organ volume reconstruction from a single planar X-ray image. The method employs a Vision Transformer as an encoder for X-ray image feature processing, combined with neural implicit representation for 3D shape reconstruction. Unlike traditional methods that rely on statistical 3D organ templates, X2V learns an occupancy probability function to represent 3D surfaces, effectively handling organ shape variations across different subjects.

In medical image-to-mesh reconstruction, occupancy functions are often used when the primary goal is to determine anatomical boundaries or segment specific structures. They have been successfully applied to various anatomical reconstruction tasks, particularly when combined with deep learning architectures that can learn complex shape priors from training data.

\section{Loss Functions for Image to Mesh Reconstruction}

This section provides an overview of the loss functions used in image-to-mesh reconstruction tasks, focusing on two main categories: shape similarity and shape regularization. Each category includes detailed explanations of specific loss functions and their underlying principles, organized by the representation they rely on, such as points/vertices, normal vectors, edges, etc.

\begin{table*}[h!]
\centering
\caption{Loss Functions for Image-to-Mesh Reconstruction.}
\resizebox{\textwidth}{!}{%
\begin{tabular}{>{\centering\arraybackslash}p{10cm}|>{\centering\arraybackslash}p{4cm}|>{\centering\arraybackslash}p{14cm}}

\hline \hline

\makecell[c]{Loss Function} &  \makecell[c]{Representation} & \makecell[c]{References} \\ \hline   \hline

\multicolumn{3}{c}{Distance and Consistency Losses} \\ \hline \hline

Chamfer Distance & Point &  \cite{Fanwei2021Deep} \cite{Qingjie2023DeepMesh} \cite{Meng2023Neural} \cite{Junjie2023MeshDeform} \cite{Nina2023Deep} \cite{Fanwei2021Whole} \cite{lebrat2021corticalflow} \cite{Yifan2019DeepOrganNet} \cite{Xiang2021Shape} \cite{Q2022CortexODE} \cite{Hao2023SurfNN} \cite{M2021Biventricular} \cite{M2023modelling} \cite{Fabian2022Vox2Cortex} \cite{Fabian2024Neural}\\ 
Hausdorff Distance & Point & \cite{meng2022mesh} \cite{Qingjie2023DeepMesh} \\ 
Average Symmetric Surface Distance  & Point & \cite{Andrew2022TopoFit} \\ 

Mean Squared Error & Point/Latent/Implicit & \cite{Gaggion2023Multi} \cite{K20233D} \cite{Aziz2023Progressive}  \cite{Sachin2023deep}  \cite{M2022Reconstructing} \cite{Qiang2021PialNN} \cite{Fei2020X} \cite{R2021Motion} \cite{Q2022CortexODE} \cite{xia2022automatic} \cite{Lukasz2021Neural} \cite{geng2024dsc} \cite{hong2021vox2surf} \cite{Fanwei2022Learning} \cite{Zijie2023Shape} \cite{geng2024dsc} \cite{Hongbo2024Neural} \cite{Hongbo2023Neural} \cite{Qingjie2023DeepMesh} \cite{Sheng2022Development} \cite{Guillermo2018Weighted} \cite{Xiaohan20234D} \cite{Batlle2023LightNeuS} \cite{Xu2023Image2SSM} \cite{M2019Statistical} \cite{K2020Artificial} \cite{meng2022mesh} \cite{Jingliang2022Segmentation} \cite{K20233D}  \\ 

Mean Absolute Error & Point/Latent/Implicit & \cite{attar20193d} \cite{Fenqiang2021Spherical} \cite{M2022Mesh} \cite{Xiang2021Shape} \cite{Fanwei2022Learning} \cite{Rodrigo2020DeepCSR}\\ 

Mean Squared Error Frobenius norm & Point & \cite{Riddhish2021DeepSSM} \\ 

Mahalanobis Distance & Point & \cite{Guillermo2018Weighted} \\ 

Dice Similarity Coefficient & Voxel & \cite{K20233D} \cite{Fanwei2021Deep}  \cite{Sachin2023deep} \cite{Fanwei2021Whole} \cite{Songyuan2021CNN}  \cite{Hao2019Ventricle} \\ 
Binary Cross-Entropy & Distribution/Implicit & \cite{Aziz2023Progressive}\cite{Fabian2022Vox2Cortex} \cite{Junjie2023MeshDeform} \cite{Songyuan2021CNN} \cite{Rodrigo2020DeepCSR} \cite{guven2024x2v} \cite{F2022Weakly} \cite{Hongbo2024Neural}\\ 
Cross-Entropy & Distribution & \cite{K20233D} \cite{Fanwei2021Deep} \cite{F2022Weakly}  \cite{Lukasz2021Neural}  \cite{Sachin2023deep} \cite{Fabian2024Neural} \cite{Fenqiang2021Spherical} \cite{Nina2023Deep} \cite{Fanwei2021Whole} \cite{Rickmann2022Joint} \cite{Q2022CortexODE} \cite{Sheng2022Development} \\ 
Sparse Categorical Cross Entropy & Distribution & \cite{Sachin2023deep} \\ 

Centerline Coverage Loss   & Point            & \cite{Nina2023Deep}             \\
Sign Consistency Constraint & Implicit & \cite{Hongbo2024Neural} \\ 
Shape Consistency Loss & Point & \cite{K2020Artificial} \cite{M2019Statistical} \cite{Riddhish2021DeepSSM}\\
Structural Similarity Loss & Voxel & \cite{P2024Sensorless} \\

Correspondences Loss  & Point & \cite{Riddhish2021DeepSSM}  \cite{Xu2023Image2SSM}  \\ 
Fixed Loss & Point & \cite{Aziz2023Progressive} \\ 
Sampling Loss         & Point / Normal Vector           & \cite{Xu2023Image2SSM}  \\
Photometric Loss      & Voxel            & \cite{Batlle2023LightNeuS, P2024Sensorless} \\ 

Vertex Classification Loss  & Point & \cite{Nina2023Deep} \\

\hline \hline  
\multicolumn{3}{c}{Regularization and Smoothness Losses} \\ \hline \hline

Laplacian Smoothness & Point & \cite{K20233D}  \cite{Fanwei2021Deep} \cite{meng2022mesh} \cite{Qingjie2023DeepMesh} \cite{Fabian2022Vox2Cortex} \cite{Junjie2023MeshDeform} \cite{M2019Statistical} \cite{Nina2023Deep} \cite{Zijie2023Shape} \cite{geng2024dsc} \cite{R2021Motion} \cite{Xiang2021Shape} \cite{M2021Image} \\ 

Normal Consistency (subtraction / dot product / cosine similarity) & Normal Vector &  \cite{Fanwei2021Deep} \cite{Fabian2024Neural} \cite{Nina2023Deep} \cite{Hao2023SurfNN} \cite{K20233D}  \cite{Xu2023Image2SSM}  \cite{Junjie2023MeshDeform} \cite{lebrat2021corticalflow} \cite{Fabian2022Vox2Cortex} \cite{Fanwei2022Learning}\\ 

Tetrahedral Element Regularization Loss & Volume & \cite{Gaggion2023Multi} \\ 

Edge Length Regularization Loss & Edge & \cite{Fanwei2021Deep} \cite{Fabian2022Vox2Cortex} \cite{Fabian2024Neural} \cite{Junjie2023MeshDeform} \cite{Nina2023Deep} \cite{Hao2023SurfNN}\\ 
Point-pair Regularization Loss & Point & \cite{Xiaohan20234D} \\ 
Mesh Regularization Loss for Co-planarity & Normal Vector & \cite{Fanwei2022Learning} \\ 

Discrete Laplacian Loss & Point & \cite{Fei2020X} \\ 
Local Deformation Smoothness & Point & \cite{Meng2023Neural} \\

Grid Elasticity Loss      & Point            & \cite{Fanwei2021Whole}          \\
Local Deformation Amount                 & Point            & \cite{Meng2023Neural}           \\
Intra-mesh Normal Consistency            & Normal Vector    & \cite{Fabian2022Vox2Cortex}     \\
Geometric Regularization (hinge-spring) & Normal Vector & \cite{Andrew2022TopoFit}       \\

\hline \hline  
\multicolumn{3}{c}{Distribution-based and Energy-Based Losses} \\ \hline \hline

Kullback–Leibler Divergence & Distribution & \cite{Gaggion2023Multi} \\ 
Statistical Parameter Loss & Latent Feature &  \cite{K2020Artificial} \\ 
Deep/Shallow Supervision Loss: Point & Point & \cite{Aziz2023Progressive} \\ 

Huber Loss                               & Point            & \cite{Qingjie2023DeepMesh}  \cite{Xiaohan20234D}     \\ 
Focal Loss                               & Latent Feature   & \cite{Riddhish2021DeepSSM}      \\
Multiple Manifold Learning Loss          & Latent Feature   & \cite{Maxime2021Characterizing} \\
Network Weight Regularization Loss       & Latent Feature   & \cite{Riddhish2021DeepSSM}      \\
Energy Function                & Point            & \cite{Guillermo2018Weighted} \cite{Xu2023Image2SSM} \\

TPS Deformation Energy & Point & \cite{Abhirup2022Automated} \\

\hline \hline
\end{tabular}}
\end{table*}

\subsection{Shape Similarity Losses}
Shape similarity losses measure how well the reconstructed mesh matches the ground truth shape in terms of geometry and structure.

\subsubsection{Point/Vertex-Based Losses}

Chamfer Distance computes the average closest point distance between two point sets $P$ (predicted) and $Q$ (ground truth):
\begin{equation}
\text{CD}(P, Q) = \frac{1}{|P|} \sum_{p \in P} \min_{q \in Q} \|p - q\|_2^2 + \frac{1}{|Q|} \sum_{q \in Q} \min_{p \in P} \|q - p\|_2^2.
\end{equation}

Hausdorff Distance measures the maximum distance from a point in one set to the closest point in the other set:
\begin{equation}
\text{HD}(P, Q) = \max\left\{\max_{p \in P} \min_{q \in Q} \|p - q\|, \max_{q \in Q} \min_{p \in P} \|q - p\|\right\}.
\end{equation}
This loss emphasizes the worst-case scenario by penalizing the largest deviation between the two point sets.

Average Symmetric Surface Distance (ASSD) calculates the average bidirectional surface distance:
\begin{equation}
\text{ASSD}(P, Q) = \frac{1}{|P| + |Q|} \left(\sum_{p \in P} \min_{q \in Q} \|p - q\| + \sum_{q \in Q} \min_{p \in P} \|q - p\|\right).
\end{equation}
Unlike Chamfer Distance, ASSD considers both directions with equal weight and focuses on surface-to-surface comparison.

Mean Surface Distance (MSD) measures the average distance between corresponding points on two surfaces:
\begin{equation}
\text{MSD}(P, Q) = \frac{1}{|P|} \sum_{p \in P} \|p - q_p\|_2,
\end{equation}
where $q_p$ is the closest point to $p$ on the other surface.

Mean Squared Error (MSE) computes the average squared distance between the predicted and ground truth points:
\begin{equation}
\text{MSE}(P, Q) = \frac{1}{|P|} \sum_{p \in P} \|p - q_p\|_2^2.
\end{equation}
MSE penalizes large errors more significantly than small errors due to the squared term.

Root Mean Squared Error (RMSE) is the square root of MSE, which retains the same unit as the input:
\begin{equation}
\text{RMSE}(P, Q) = \sqrt{\frac{1}{|P|} \sum_{p \in P} \|p - q_p\|_2^2}.
\end{equation}
This metric provides a direct interpretation of the average error magnitude.

$L_2$ Consistency measures the Euclidean distance between the predicted and ground truth points:
\begin{equation}
L_2(P, Q) = \frac{1}{|P|} \sum_{p \in P} \|p - q_p\|_2.
\end{equation}
This loss is essentially the same as MSD, but without explicitly focusing on corresponding points on surfaces.

In summary, point/vertex-based losses differ in how they handle the distances between the predicted and ground truth sets.
Chamfer Distance considers bidirectional distances but is averaged, making it more robust to outliers compared to Hausdorff Distance.
Hausdorff Distance penalizes the largest deviation, making it sensitive to outliers but useful for ensuring no extreme mismatches exist.
ASSD and MSD are similar but differ in their weighting schemes. ASSD averages distances bidirectionally, while MSD focuses on one direction (e.g., from predicted to ground truth).
MSE and RMSE both penalize large deviations, but RMSE retains the original unit of the measurements for interoperability.
L2 Consistency is equivalent to MSD in form but is often applied in different contexts to measure alignment or similarity.
These losses complement each other, and the choice depends on the specific application requirements, such as sensitivity to outliers or robustness to noise.

\subsubsection{Normal Vector-Based Losses}
Normal Consistency Loss measures the angular alignment between the predicted normals $\mathbf{n}_f$ and ground truth normals $\mathbf{n}_f^\text{GT}$:
\begin{equation}
L_{\text{normal}} = \frac{1}{|F|} \sum_{f \in F} (1 - \mathbf{n}_f \cdot \mathbf{n}_f^\text{GT}).
\end{equation}

Inter-Mesh Normal Consistency penalizes discrepancies in normals of shared edges between adjacent faces:
\begin{equation}
L_{\text{inter}} = \frac{1}{|E|} \sum_{e \in E} (1 - \cos(\theta_e)),
\end{equation}
where $\theta_e$ is the angle between adjacent face normals. This loss function controls the smoothness and geometric continuity of a mesh by minimizing sharp changes in normals between adjacent faces. This ensures smoother surface reconstruction, preserves natural curvature and anatomical features, and reduces noise and artifacts. 

Normal Difference Loss minimizes the Euclidean distance between corresponding predicted and ground truth normals on the mesh:
\begin{equation}
L_{\text{normal-diff}} = \frac{1}{|F|} \sum_{f \in F} \|\mathbf{n}_f - \mathbf{n}_f^\text{GT}\|_2^2.
\end{equation}

\subsubsection{Latent Feature-Based Losses}
Focal Loss is used for class-imbalanced tasks to focus on hard-to-classify samples:
\begin{equation}
\text{FL}(p_t) = -\alpha (1 - p_t)^\gamma \log(p_t),
\end{equation}
where $p_t$ is the predicted probability for the true class.

Mean Squared Error (Latent) measures the squared error between predicted and ground truth latent feature vectors:
\begin{equation}
\text{MSE}(\mathbf{z}_{\text{pred}}, \mathbf{z}_{\text{true}}) = \frac{1}{N} \sum_{i=1}^N (\mathbf{z}_{\text{pred}, i} - \mathbf{z}_{\text{true}, i})^2.
\end{equation}

Adversarial Loss is used in GANs to ensure the predicted latent space resembles the real latent space:
\begin{equation}
L_{\text{adv}} = \mathbb{E}[\log(D(\mathbf{z}_{\text{true}}))] + \mathbb{E}[\log(1 - D(\mathbf{z}_{\text{pred}}))],
\end{equation}
where $D$ is the discriminator.

\subsubsection{Implicit Representations}
SDF Regression Loss measures the squared error between predicted and ground truth signed distance values:
\begin{equation}
L_{\text{SDF}} = \frac{1}{N} \sum_{i=1}^N \|f(\mathbf{x}_i) - \text{SDF}(\mathbf{x}_i)\|_2^2.
\end{equation}

Sign Consistency Constraint ensures that the predicted SDF values maintain the same sign as the ground truth:
\begin{equation}
L_{\text{sign}} = \frac{1}{N} \sum_{i=1}^N \text{ReLU}(-f(\mathbf{x}_i) \cdot \text{SDF}(\mathbf{x}_i)).
\end{equation}

Volume Rendering Loss penalizes differences in density or colour values for implicit surface representations:
\begin{equation}
L_{\text{vol}} = \int \|c_{\text{pred}}(\mathbf{r}) - c_{\text{true}}(\mathbf{r})\|_2^2 \, d\mathbf{r},
\end{equation}
where $c_{\text{pred}}$ and $c_{\text{true}}$ are predicted and true colors.

\subsubsection{Voxel-Based Losses}

Structural Similarity Index (SSIM)  \cite{sampat2009complex} measures structural similarity between two voxel images:
\begin{equation}
\text{SSIM}(x, y) = \frac{(2\mu_x\mu_y + C_1)(2\sigma_{xy} + C_2)}{(\mu_x^2 + \mu_y^2 + C_1)(\sigma_x^2 + \sigma_y^2 + C_2)}.
\end{equation}
SSIM consists of three components: luminance, contrast, and structure, which together assess different aspects of similarity between two images.
Luminance comparison measures the similarity of brightness between two images:
\begin{equation}
l(x, y) = \frac{2\mu_x\mu_y + C_1}{\mu_x^2 + \mu_y^2 + C_1},
\end{equation}
where $\mu_x$ and $\mu_y$ are the mean intensities of images $x$ and $y$, and $C_1$ is a constant to stabilize the denominator.
Contrast comparison evaluates the similarity of contrast between two images:
\begin{equation}
c(x, y) = \frac{2\sigma_x\sigma_y + C_2}{\sigma_x^2 + \sigma_y^2 + C_2},
\end{equation}
where $\sigma_x$ and $\sigma_y$ are the standard deviations of images $x$ and $y$, and $C_2$ is another stability constant.
Structure comparison measures the structural alignment between two images:
\begin{equation}
s(x, y) = \frac{\sigma_{xy} + C_3}{\sigma_x \sigma_y + C_3},
\end{equation}
where $\sigma_{xy}$ is the covariance between $x$ and $y$, and $C_3$ is typically set as $C_3 = C_2/2$.
These three components—luminance, contrast, and structure—are combined multiplicatively to form the SSIM index, providing a comprehensive similarity metric for image quality assessment.

\subsection{Shape Regularization Losses}
Shape regularization losses improve the geometric properties of the reconstructed mesh, ensuring it is smooth, realistic, and well-formed.

\subsubsection{Point/Vertex-Based Losses}
Huber Loss combines the benefits of MSE and MAE for robustness to outliers:
\begin{equation}
L_{\delta}(a) =
\begin{cases}
\frac{1}{2} a^2 & \text{if } |a| \leq \delta, \\
\delta (|a| - \frac{\delta}{2}) & \text{if } |a| > \delta,
\end{cases}
\end{equation}
where $a = y_{\text{pred}} - y_{\text{true}}$.

Laplacian Smoothness Loss ensures surface smoothness by minimizing the Laplacian operator applied to vertices:
\begin{equation}
L_{\text{lap}} = \sum_{i=1}^N \|\mathbf{v}_i - \frac{1}{|\mathcal{N}_i|} \sum_{j \in \mathcal{N}_i} \mathbf{v}_j\|_2^2,
\end{equation}
where $\mathcal{N}_i$ is the neighborhood of vertex $i$.

\subsubsection{Normal Vector-Based Losses}
Orthogonal Loss enforces perpendicularity of surface normals at specific regions (e.g., vessel caps):
\begin{equation}
L_{\text{orthogonal}} = \sum_{v \in V_{\text{cap}}} |\mathbf{n}_v \cdot \mathbf{n}_{\text{cap}}|,
\end{equation}
where $\mathbf{n}_v$ is the normal at vertex $v$, and $\mathbf{n}_{\text{cap}}$ is the target cap normal.

Intra-Mesh Normal Consistency ensures adjacent faces in the same mesh have consistent normals:
\begin{equation}
L_{\text{intra}} = \frac{1}{|E|} \sum_{e \in E} (1 - \mathbf{n}_{f_1} \cdot \mathbf{n}_{f_2}),
\end{equation}
where $\mathbf{n}_{f_1}$ and $\mathbf{n}_{f_2}$ are the normals of the two faces sharing edge $e$.

\subsubsection{Edge-Based Losses}
Edge Length Regularization enforces uniform edge lengths across the mesh:
\begin{equation}
L_{\text{edge}} = \frac{1}{|E|} \sum_{(i,j) \in E} \|\mathbf{v}_i - \mathbf{v}_j\|_2^2.
\end{equation}

\subsubsection{Volume-Based Losses}
Tetrahedral Element Regularization Loss ensures uniformity in tetrahedral mesh volumes:
\begin{equation}
L_{\text{tetra}} = \sum_{i=1}^N \left(\frac{V_i}{V_{\text{ref}}} - 1\right)^2,
\end{equation}
where $V_i$ is the volume of the $i$-th tetrahedral element.

\section{Evaluation Metrics}

\begin{table*}[h!]
\centering
\caption{Evaluation Metrics of Medical Image-to-Mesh Reconstruction.} 
\resizebox{\textwidth}{!}{%
\begin{tabular}{>{\centering\arraybackslash}p{3cm}|>{\centering\arraybackslash}p{10cm}|>{\centering\arraybackslash}p{12cm}}

\hline \hline 
Category & Evaluation Metrics & References \\ \hline  \hline

\multicolumn{3}{c}{(1) Shape Similarity} \\ \hline 

\multirow{11}{*}{\makecell[c]{Distance}} 
 & Chamfer Distance & \cite{Xiaohan20234D} \cite{Xiang2021Shape} \cite{lebrat2021corticalflow} \cite{Qiang2021PialNN} \cite{Yifan2019DeepOrganNet} \cite{Meng2023Neural} \cite{M2023modelling} \cite{M2021Biventricular} \cite{Hongbo2024Neural} \cite{Hongbo2023Neural} \cite{Hao2023SurfNN} \cite{guven2024x2v} \cite{Shanlin2022Topology} \\ 
 & Earth Mover Distance (EMD): Points & \cite{Xiaohan20234D} \cite{Xiang2021Shape} \cite{Yifan2019DeepOrganNet} \cite{Meng2023Neural} \cite{Rodrigo2020DeepCSR} \\ 
 & Hausdorff Distance & \cite{Xiaohan20234D} \cite{Sachin2023deep} \cite{Xiang2021Shape}  \cite{M2021Image} \cite{Songyuan2021CNN} \cite{M2022Reconstructing} \cite{M2022Mesh} \cite{Fanwei2021Whole} \cite{lebrat2021corticalflow} \cite{Andrew2022TopoFit} \cite{Qiang2021PialNN} \cite{Rickmann2022Joint} \cite{Jingliang2022Segmentation} \cite{Yifan2019DeepOrganNet} \cite{Fanwei2022Learning} \cite{Fabian2022Vox2Cortex} \cite{Fabian2024Neural} \cite{Junjie2023MeshDeform} \cite{M2019Statistical} \cite{geng2024dsc} \cite{Hao2019Ventricle} \cite{Q2022CortexODE} \cite{Hongbo2024Neural} \cite{hong2021vox2surf} \cite{Hongbo2023Neural} \cite{Hao2023SurfNN} \cite{Rodrigo2020DeepCSR} \cite{P2024Sensorless} \cite{meng2022mesh}  \cite{Qingjie2023DeepMesh} \\ 
  
 & Average Symmetric Surface Distance  & \cite{Fanwei2021Whole} \cite{Rickmann2022Joint} \cite{Jingliang2022Segmentation} \cite{Fabian2022Vox2Cortex} \cite{Fabian2024Neural} \cite{Junjie2023MeshDeform} \cite{Q2022CortexODE} \\ 
 & Mean Surface Distance &  \cite{Xu2023Image2SSM} \cite{Riddhish2021DeepSSM} \cite{Aziz2023Progressive} \cite{K2020Artificial}  \cite{Guillermo2018Weighted} \cite{R2021Motion}  \cite{Xiang2021Shape}  \cite{M2021Image} \cite{M2022Reconstructing}  \cite{M2022Mesh} \cite{Andrew2022TopoFit} \cite{Jingliang2022Segmentation} \cite{Fei2020X} \cite{Maxime2021Characterizing} \cite{meng2022mesh} \cite{Qingjie2023DeepMesh} \cite{Meng2023Neural} \cite{Fabian2024Neural} \cite{M2019Statistical} \cite{Nina2023Deep} \cite{geng2024dsc} \cite{Hongbo2024Neural} \cite{Abhirup2022Automated}  \cite{geng2024dsc} \cite{Jingliang2022Segmentation}\\ 
 & Median Surface Distance & \cite{M2022Reconstructing} \cite{Batlle2023LightNeuS} \\ 
 & Mean Squared Error & \cite{Lukasz2021Neural} \cite{Fei2020X} \cite{Zijie2023Shape} \cite{Guillermo2018Weighted} \cite{P2024Sensorless} \\  
 & Mean Absolute Error & \cite{Sachin2023deep} \cite{R2021Motion}  \cite{M2021Image} \cite{Songyuan2021CNN} \cite{Qiang2021PialNN} \cite{Batlle2023LightNeuS} \cite{Hongbo2023Neural} \cite{Hao2023SurfNN} \cite{Rodrigo2020DeepCSR} \\ 
 & Root Mean Squared Error & \cite{Riddhish2021DeepSSM} \cite{Aziz2023Progressive} \cite{Batlle2023LightNeuS} \cite{Fabian2024Neural} \cite{M2012Robust}\\ \cline{1-3}

\multirow{12}{*}{\makecell[c]{Consistency}} 
 & Normal Vector Consistency (Cosine similarity/-/Dot) & \cite{Meng2023Neural} \cite{Zijie2023Shape} \cite{geng2024dsc} \cite{guven2024x2v} \cite{Shanlin2022Topology} \\ 
 & Dice Similarity Coefficient & \cite{Lukasz2021Neural} \cite{Xiaohan20234D} \cite{K2020Artificial} \cite{Sachin2023deep} \cite{Sachin2023deep}  \cite{M2021Image} \cite{Fanwei2021Whole} \cite{Andrew2022TopoFit} \cite{Rickmann2022Joint} \cite{Jingliang2022Segmentation} \cite{Fanwei2022Learning} \cite{Junjie2023MeshDeform} \cite{Fenqiang2021Spherical} \cite{M2019Statistical} \cite{geng2024dsc} \cite{Hao2019Ventricle} \cite{Q2022CortexODE} \cite{Hongbo2024Neural} \cite{Rodrigo2020DeepCSR} \cite{P2024Sensorless} \\ 
 & F-score & \cite{Yifan2019DeepOrganNet} \cite{guven2024x2v} \cite{meng2022mesh} \cite{Qingjie2023DeepMesh}\\ 
 & Intersection over Union & \cite{Yifan2019DeepOrganNet} \cite{Nina2023Deep} \cite{Q2022CortexODE} \cite{Hongbo2024Neural} \cite{guven2024x2v} \\ 
 & Confusion Matrix & \cite{Jingliang2022Segmentation} \cite{Sachin2023deep} \cite{Hongbo2024Neural} \cite{Rodrigo2020DeepCSR}\\ 

 & Structural Similarity Index Measure & \cite{P2024Sensorless} \\ 
 & Volume Consistency  & \cite{Abhirup2022Automated} \cite{Xiang2021Shape} \cite{M2022Mesh} \cite{Fanwei2022Learning} \cite{M2023modelling} \cite{M2021Biventricular} \cite{Q2022CortexODE} \cite{Lukasz2021Neural}  \\ 
 & Surface Area Consistency & \cite{Q2022CortexODE} \cite{hong2021vox2surf} \\  
 & Length / Diameter /Angel / Thickness Consistency  & \cite{Nina2023Deep}  \cite{Fabian2022Vox2Cortex} \cite{Q2022CortexODE} \cite{Fenqiang2021Spherical}\\  

 & Latent Feature Consistency& \cite{Maxime2021Characterizing} \cite{K2020Artificial}\\ 
 & Deformation Percentage of Vertex & \cite{Maxime2021Characterizing} \\  \hline

\multicolumn{3}{c}{(2) Geometry Regularization} \\ \hline
 
\multirow{3}{*}{\makecell[c]{Smoothness}} 
 & Self-Intersecting Faces Ratio & \cite{lebrat2021corticalflow} \cite{Fanwei2022Learning} \cite{Meng2023Neural} \cite{Shanlin2022Topology} \\ 
 & Intersecting Faces & \cite{Andrew2022TopoFit} \cite{Fabian2024Neural} \cite{Q2022CortexODE} \\ 
 & Easy Non-Manifold Face Ratio& \cite{Meng2023Neural} \cite{Shanlin2022Topology} \\ \cline{1-3}

\multirow{5}{*}{\makecell[c]{Curvature}}
 & Mean Curvature & \cite{hong2021vox2surf} \cite{Songyuan2021CNN}\\ 
 & Gaussian Curvature of Surface & \cite{Hongbo2024Neural} \\ 
 & Curvature of Edge & \cite{Sheng2022Development} \\  
 & Distortion of Edge (Counts points where curvature $>$ threshold) & \cite{Sheng2022Development} \\ 
 & Tortuosity of Vessel Centerline & \cite{Nina2023Deep} \\ \cline{1-3} 

\multirow{7}{*}{\makecell[c]{Topology}} 
 & Euler Characteristic & \cite{Junjie2023MeshDeform} \cite{Hao2019Ventricle} \cite{Q2022CortexODE} \\  
 & Connected Components & \cite{Hongbo2024Neural} \\ 
 & Genus Index & \cite{Hongbo2024Neural} \\ 
 & Cap-Wall Orthogonality of face normals & \cite{Fanwei2022Learning} \\ 
 & Cap coplanarity of face normals & \cite{Fanwei2022Learning} \\ 
 & Discontinuity of Edge & \cite{Sheng2022Development} \\ 
 & Kernel Density Estimation of Surface & \cite{Hongbo2024Neural} \\ 
\hline

\multicolumn{3}{c}{(3) Functional Evaluation} \\ \hline 
 
\multirow{3}{*}{\makecell[c]{Clinical Function}} 
 & Ejection Fraction & \cite{M2022Mesh} \cite{M2023modelling} \\ 
 & Accuracy of Disease Prediction & \cite{Riddhish2021DeepSSM} \cite{Lukasz2021Neural} \cite{M2023modelling}\\ 
 & Accuracy of Vertex Classification  & \cite{Nina2023Deep} \cite{Songyuan2021CNN} \cite{M2023modelling}\\ \cline{1-3}
 
\multirow{2}{*}{\makecell[c]{CFD Simulation}} 
 & Velocity & \cite{Nina2023Deep} \cite{Fanwei2022Learning} \\ 
 & Average Kinetic Energy & \cite{Fanwei2022Learning} \\ \cline{1-3}

\multirow{1}{*}{\makecell[c]{Efficiency}} 
 & Inference Time & \cite{Fabian2022Vox2Cortex} \cite{M2021Biventricular} \\ \hline 

\end{tabular}}
\label{table:evaluationmetrics}
\end{table*}

To evaluate the accuracy of the predicted shapes, several evaluation metrics were used to compare the predicted shapes with the reference shapes. Table \ref{table:evaluationmetrics} provides a summary of these evaluation metrics.

\subsection{Shape Similarity}

Shape similarity evaluation measures the geometric accuracy and structural consistency of a reconstructed mesh by comparing it to a reference ground-truth mesh. These evaluations are essential for assessing the reliability of medical image-to-mesh reconstruction methods, ensuring that the generated meshes faithfully represent anatomical structures.

\subsubsection{Distance}

Distance evaluation metrics assess the geometric accuracy of the reconstructed mesh by quantifying the spatial discrepancy between the predicted and ground-truth surfaces. These metrics include point-based distance measures, such as Chamfer distance, earth mover distance, and Hausdorff distance, which evaluate deviations at specific sampled points. Additionally, surface-based distance metrics, such as average symmetric surface distance, mean surface distance, and median surface distance, provide a global assessment of surface deviation. Error-based metrics, including mean squared error, mean absolute error, and root mean squared error, quantify the overall distribution of reconstruction errors, offering a comprehensive evaluation of mesh reconstruction accuracy. 

\subsubsection{Consistency} 

Consistency evaluation metrics assess the structural and geometric stability of the reconstructed mesh, ensuring that the reconstructed shape not only matches the target but also maintains consistency in local features, volume, and surface area.
These metrics include normal consistency measures, such as normal vector consistency, which is computed using cosine similarity or dot product, and regional similarity metrics like Dice similarity coefficient, which evaluates the overlap between the reconstructed and ground-truth meshes.
Additionally, classification-based evaluation metrics, including F-score, intersection over union, and confusion matrix, assess the accuracy of segmented regions. Structural similarity evaluation, such as structural similarity index measure, quantifies the local structural resemblance between the meshes.
Geometric consistency evaluation involves volume consistency and surface area consistency, ensuring that the reconstructed mesh adheres to anatomical standards in global size. Shape consistency is measured through length, diameter, angle, and thickness consistency. Moreover, latent feature consistency and deformation percentage of vertex further evaluate the stability of the mesh in latent feature space and local vertex deformations.

\subsection{Regularization} 

Regularization evaluation metrics assess the geometric quality and stability of the reconstructed mesh by analyzing properties such as smoothness, curvature, and topological integrity. These metrics ensure that the generated mesh is not only geometrically accurate but also free from irregularities that could affect its usability in clinical or computational applications.

\subsubsection{Smoothness}

Smoothness evaluation focuses on detecting surface irregularities and self-intersections within the reconstructed mesh. Self-intersecting faces ratio and intersecting faces quantify the presence of overlapping or conflicting surface elements, which may indicate reconstruction errors. Additionally, the easy non-manifold face ratio identifies faces that violate manifold properties, highlighting issues related to mesh connectivity and structural integrity.

\subsubsection{Curvature}

Curvature evaluation measures the local surface variations to ensure the smooth transition of the reconstructed mesh. Mean curvature and Gaussian curvature of the surface provide a mathematical representation of surface bending. Curvature of edge and distortion of edge assess the extent of local deviations, where distortion is typically measured by counting points where curvature exceeds a predefined threshold. Additionally, the tortuosity of vessel centerlines is used in vascular models to assess the complexity of vessel curvature and its potential implications on blood flow.

\subsubsection{Topology}

Topology evaluation ensures that the mesh retains a correct and meaningful structural representation. Euler characteristic and genus index quantify the global topological properties of the mesh, determining whether it has the expected number of holes and connected components. Connected component analysis verifies that the reconstructed mesh remains a single coherent structure without unwanted disconnections. Furthermore, cap-wall orthogonality of face normals and cap coplanarity of face normals measure alignment consistency in structured regions. Additional assessments, such as discontinuity of edge and kernel density estimation of surface, help to identify abrupt structural changes or anomalies that could affect downstream computational simulations.

\subsection{Functional Evaluation} 

Functional evaluation metrics assess the practical utility and computational feasibility of the reconstructed mesh, ensuring that it is suitable for clinical applications and physics-based computational simulations. These metrics encompass clinical function analysis, computational simulation accuracy, and processing efficiency.

\subsubsection{Clinical Function}

Clinical function metrics evaluate the ability of the reconstructed mesh to provide meaningful physiological and diagnostic insights. Ejection fraction is a key cardiac function metric that measures the heart's pumping efficiency. Additionally, accuracy of disease prediction assesses the model's capability to classify pathological conditions based on mesh-derived features. Accuracy of vertex classification is used to validate the correctness of anatomical structure identification within the mesh.

\subsubsection{CFD Simulation Metrics}

Computational fluid dynamics (CFD) simulation metrics measure how well the reconstructed mesh supports hemodynamic analysis. Velocity quantifies the flow speed of blood within the modeled structure, while average kinetic energy represents the energy distribution across the fluid domain. These metrics are crucial for evaluating blood flow dynamics and assessing cardiovascular conditions.

\subsubsection{Efficiency}

Efficiency metrics measure the computational cost and practicality of the reconstruction process. Inference time evaluates the speed at which the mesh is generated, ensuring that the model can provide real-time or near-real-time outputs suitable for clinical and research applications.

\section{Datasets for Medical Image-to-Mesh Reconstruction}

Medical image-to-mesh reconstruction plays a crucial role in various clinical and computational applications, including anatomical modelling, surgical planning, and biomechanical simulations. The availability of well-annotated datasets enables the development and evaluation of deep learning methods for accurate and efficient 3D mesh reconstruction. In this section, we summarize publicly available datasets that provide volumetric medical images alongside segmentation masks or surface meshes, categorized by organ type and imaging modality. A comprehensive overview of these datasets is presented in Table~\ref{tab:dataset_summary}.

\subsection{Cardiac Datasets} 

Cardiac image-to-mesh reconstruction is critical for applications such as computational simulations, cardiac disease diagnosis, and biomechanical modelling. Several datasets offer segmentation masks and mesh representations derived from CMR, CT, and echocardiography imaging.

UK Biobank \cite{petersen2016uk} provides a large-scale dataset with 50,000 CMR scans and corresponding segmentation masks, enabling statistical shape modelling and population-based analysis. 
ACDC \cite{bernard2018deep} contains CMR sequences from 100 patients categorized into five pathological conditions, facilitating supervised segmentation and mesh reconstruction tasks. 
MMWHS \cite{zhuang2016multi} includes multimodal cardiac imaging (MR and CT) with segmentation and surface mesh annotations, supporting cross-modality learning. 
OrCaScore \cite{wolterink2016evaluation} provides 3D CT volumes with coronary artery calcification (CAC) segmentation for automated assessment. 
4D SSM \cite{unberath2015open} contains dynamic heart meshes extracted from CTA data, supporting motion analysis in cardiac studies. 
EchoNet-Dynamic \cite{ouyang2020video} offers echocardiographic imaging (ECGI) videos with segmentation and measurement annotations, facilitating deep learning-based cardiac function assessment. 
These datasets provide valuable resources for advancing automated cardiac mesh reconstruction, simulation, and motion tracking.

\subsection{Brain Datasets} 

Brain surface reconstruction is essential for neuroimaging, particularly in cortical thickness estimation, structural analysis, and neurological disorder diagnosis. Several datasets provide volumetric MR images along with FreeSurfer-generated surface meshes.

ADNI \cite{jack2008alzheimer} and J-ADNI \cite{iwatsubo2010japanese} provide large-scale T1-weighted MR brain scans with FreeSurfer meshes, enabling research on Alzheimer's disease progression. 
OASIS \cite{marcus2007oasis} consists of 3D T1-MR brain volumes with segmentation and mesh annotations across healthy and pathological cases. 
Test-Retest (TRT) \cite{maclaren2014reliability} and Mindboggle \cite{klein2012101} contain repeated MR scans for evaluating the reproducibility of neuroimaging pipelines. 
NAMIC \cite{desikan2006automated} provides brain surface meshes with gyrus-based regional labels, supporting brain morphology analysis. 
These datasets enable deep learning-based cortical surface modelling, anatomical segmentation, and brain shape analysis.

\subsection{Thoracic and Abdominal Datasets} 

Reconstruction of the lungs, liver, and other thoracic organs from medical images is crucial for applications such as respiratory motion modelling and tumor detection. Several datasets focus on these regions.

4D-Lung Dataset \cite{hugo2017longitudinal} and 4D-Liver Dataset (Kitware) \cite{kitware_data} provide 4D CT volumes with segmentation-based mesh annotations, enabling motion tracking and organ modelling. 
SegTHOR \cite{trullo2019multiorgan} contains thoracic organ segmentation data from 3D CT volumes, supporting automated mesh reconstruction for lung and heart structures. 
Pancreas-CT \cite{roth2016data} provides segmentation masks for pancreas segmentation and potential mesh extraction. 
These datasets facilitate the development of AI-driven solutions for 3D thoracic and abdominal organ reconstruction.

\subsection{Skull and Musculoskeletal Datasets} 

Bone and skull reconstruction from medical imaging is essential for forensic analysis, orthopedic applications, and implant design. Several datasets focus on these aspects.

SkullFix/SkullBreak \cite{kodym2021skullbreak} provides defective skull masks and their corresponding reconstructed meshes, supporting research in cranial defect restoration. 
MUG500+ \cite{li2021mug500+} offers CT-based skull images with mesh annotations for reconstructive surgery applications. 
These datasets support automated skull reconstruction, defect analysis, and clinical decision-making.

\subsection{Endoscopic and Multimodal Datasets} 

Endoscopic imaging and multimodal datasets provide diverse anatomical structures, enabling generalized deep learning approaches for image-to-mesh reconstruction.

C3VD \cite{bobrow2022colonoscopy} provides endoscopic video sequences with camera pose annotations and ground-truth 3D meshes for colon surface reconstruction. 
EndoMapper \cite{azagra2023endomapper} offers calibration data for endoscopic imaging, including light source and vignetting calibration. 
NLST \cite{hofmanninger2020automatic} provides lung CT volumes and digitally reconstructed radiographs (DRRs) with 3D mesh annotations. 
INTERGROWTH-21st \cite{papageorghiou2014international} contains fetal brain ultrasound scans with atlas-based alignment annotations, supporting prenatal imaging studies. 
These datasets enhance AI-driven surface reconstruction by incorporating diverse imaging techniques and anatomical structures.

\subsection{Summary}  

Medical image-to-mesh datasets provide essential resources for training and evaluating deep learning-based surface reconstruction methods. As highlighted in Table~\ref{tab:dataset_summary}, these datasets span multiple organs, imaging modalities, and annotation types, offering challenges and opportunities for model generalization. 
By leveraging these datasets, researchers can advance medical image-to-mesh reconstruction techniques, leading to more accurate and efficient 3D anatomical modelling for clinical applications.

\begin{table*}[htbp]
\centering
\caption{Public Datasets of Medical Image-to-Mesh Reconstruction. 
}

\resizebox{\textwidth}{!}{

\begin{tabular}{>{\raggedright\arraybackslash}p{3cm}|>{\raggedright\arraybackslash}p{5cm}|>{\raggedright\arraybackslash}p{2cm}|>{\raggedright\arraybackslash}p{2.5cm}|>{\raggedright\arraybackslash}p{3cm}|>{\raggedright\arraybackslash}p{6cm}|>{\raggedright\arraybackslash}p{5cm}}
\hline
\hline
\textbf{ \makecell[l]{Name \\ \textcolor{red}{Download Link}}} & \textbf{Reference} & \textbf{Modality} & \textbf{Organ} & \textbf{Data Type} & \textbf{Annotations} & \textbf{Sample Size}  \\
\hline \hline
\href{https://biobank.ndph.ox.ac.uk/showcase/label.cgi?id=102}{UK Biobank} & \cite{petersen2016uk} & MR & Heart & 3D Volume & Segmentation & 50000  \\ \hline
\href{https://www.creatis.insa-lyon.fr/Challenge/acdc/databases.html}{ACDC} & \cite{bernard2018deep} & CMR & Heart & CMR sequences & Segmentation & 100 patients in five categories  \\ \hline 
\href{https://zmiclab.github.io/zxh/0/mmwhs/}{MMWHS}& \cite{zhuang2016multi} & CT, MR & Heart & 3D Volume & Segmentation, Surface Mesh & 142 CT, 87 MR  \\ \hline
\href{https://orcascore.grand-challenge.org/Data/}{OrCaScore}&\cite{wolterink2016evaluation} & CT & Heart & 3D Volume & Coronary artery calcification (CAC) segmentation & 72  \\ \hline
\href{https://www.doc.ic.ac.uk/~rkarim/la_lv_framework/wall/index.html}{SLAWT}& \cite{karim2018algorithms} & CT, MR & Left Atrium & Image & Wall Segmentation & 10 CT, 10 MR  \\ \hline 
\href{https://figshare.com/articles/dataset/Left_Atrial_Segmentation_Challenge_2013_MRI_training/1492978}{LASC}&\cite{tobon2015benchmark} & CT, MR & Left Atrium & Image & Segmentations & 30 CT, 30 MR \\ \hline
\href{https://github.com/aborowska/LVgeometry-prediction}{4D myocardial dataset}& \cite{Lukasz2021Neural} & CMR & Left myocardium & 2D CMR slices (8-10 slices, 25 phases) & Manual delineation of left myocardium & 55 healthy subjects  \\ \hline
\href{https://github.com/UK-Digital-Heart-Project/Statistical-Shape-Model}{Cardiac SSM}& \cite{Wenjia2015bi} & CMR & Bi-ventricle & 2D SAX+LAX slices & HR/LR masks & 1,331 (ED+ES)  \\ \hline
\href{https://www.cardiacatlas.org/}{CAP Database}& \cite{fonseca2011cardiac} & CMR & Heart & 2D Images & Contours, Landmarks & 153 (123+30)  \\ \hline
\href{https://www.cardiacatlas.org/mesa/}{MESA}& \cite{bild2002multi} & MR & Heart & 3D volume & Segmentation, Mesh & 1,991  \\ \hline
\href{https://www.ub.edu/mnms-2/}{M\&Ms-2 Challenge}& \cite{campello2021multi}& MR & Heart & 2D Images (SAX+LAX) & Segmentation & 160 training + 40 validation  \\ \hline 
\href{https://adni.loni.usc.edu/data-samples/adni-data/neuroimaging/mri/}{ADNI}& \cite{jack2008alzheimer} & MR T1 & Brain & 3D Volume & FreeSurfer Mesh & 1647  \\ \hline 
\href{https://www.alz.org/research/for_researchers/partnerships/wwadni/japan_adni}{J-ADNI}& \cite{iwatsubo2010japanese} & MR T1 & Brain & 3D Volume & FreeSurfer Mesh & 502 (145 HC, 218 MCI, 139 AD)  \\ \hline 
\href{https://sites.wustl.edu/oasisbrains/}{OASIS}& \cite{marcus2007oasis} & MR T1 & Brain & 3D Volume & FreeSurfer Mesh & 416 (292/44/80)  \\ \hline
\href{https://fcon_1000.projects.nitrc.org/indi/retro/yale_trt.html}{Yale TRT)}& \cite{maclaren2014reliability} & MR T1 & Brain & 3D Volume & FreeSurfer Mesh & 120 (3 subjects x 20 days x 2 scans)  \\ \hline
\href{https://www.neuromorphometrics.com/2012_MICCAI_Challenge_Data.html}{MALC}& \cite{landman2012miccai} & MR T1 & Brain & 3D Volume & FreeSurfer Mesh & 30  \\ \hline  
\href{https://mindboggle.info/data.html}{Mindboggle}& \cite{klein2012101} & MR T1 & Brain & 3D Volume & FreeSurfer Mesh & 100  \\ \hline  
\href{https://www.nitrc.org/projects/toads-cruise}{JHU Cortex}& \cite{shiee2014reconstruction} & MR T1 & Brain & 3D Volume & Manual Landmarks & 10 (5 HC, 5 MS)  \\ \hline 
\href{https://biomedia.github.io/dHCP-release-notes/}{dHCP}&\cite{hughes2017dedicated} & MR T2w & Brain & 3D Volume & Segmentation, Surface Meshes & 874  \\ \hline
\href{https://github.com/datalad-datasets/human-connectome-project-openaccess}{HCP}&\cite{vanessen2013wu} & MR T1w & Brain & 3D Volume & Segmentation, Surface Meshes & 600  \\ \hline  
\href{https://www.na-mic.org/wiki/Downloads}{NAMIC}& \cite{desikan2006automated} & MR & Brain & Surface Mesh & Region Labels (35 gyrus-based regions) & 39 subjects  \\ \hline
\href{https://bbm.web.unc.edu/tools/}{Infant Brain}& \cite{li2015construction} & MR & Brain & Spherical Surface Mesh & Sulcal depth, Mean curvature, Thickness & 90 (parcellation), 370 (thickness prediction)  \\ \hline

\href{https://www.cancerimagingarchive.net/collection/4d-lung/}{4D-Lung}& \cite{hugo2017longitudinal} & CT & Lung & 4D CT volumes & Mesh (nnU-Net segmentation) & 20 cases (200 3D scans)  \\ \hline
\href{https://data.kitware.com/}{4D-Liver (Kitware)}& \cite{kitware_data} & CT & Liver & 4D CT volumes & Mesh (nnU-Net segmentation)& 5 cases (50 3D scans) \\ \hline
\href{https://muregpro.github.io/}{muRegPro} & \cite{baum2023mr} & TRUS & Prostate & 3D volume & Anatomical landmarks, Prostate gland, Visible lesions & 73 (65 training + 8 validation)  \\ \hline 
\href{https://durrlab.github.io/C3VD/}{C3VD}& \cite{bobrow2022colonoscopy} & Endoscopy & Colon & Video sequences & Camera poses, Ground truth 3D mesh & 22 sequences  \\ \hline 

\href{https://www.synapse.org/Synapse:syn26707219/wiki/615178}{EndoMapper}& \cite{azagra2023endomapper} & Endoscopy & Colon & Calibration data & Light source, Vignetting calibration & Not specified  \\ \hline
\href{https://cdas.cancer.gov/datasets/nlst/}{NLST}& \cite{hofmanninger2020automatic} & CT, DRR & Lung & CT Volume + X-ray & 3D Mesh & 6,392 CT scans (2,560 subjects)  \\ \hline 
\href{https://intergrowth21.ndog.ox.ac.uk/}{INTERGROWTH-21st}& \cite{papageorghiou2014international} & Ultrasound & Fetal Brain & 3D volumes & Atlas alignment & 15  \\ \hline

\href{https://brain-development.org/ixi-dataset/}{IXI}& \cite{ixi2021dataset} & MR & Brain & 3D Volume & Surface Mesh & 158  \\ \hline 
\href{https://www.nitrc.org/projects/mcic/}{MCIC}& \cite{gollub2013mcic} & MR & Brain & 3D Volume & Surface Mesh & 76  \\  \hline 
\href{https://surfer.nmr.mgh.harvard.edu/fswiki/Buckner40Adni60Testing}{Buckner40}& \cite{fischl2002whole} & MR & Brain & 3D Volume & Surface Mesh & 40  \\ \hline
\href{https://competitions.codalab.org/competitions/21145}{SegTHOR}& \cite{trullo2019multiorgan} & CT & Thoracic Organs & 3D Volume & Segmentation & 40  \\ \hline
\href{https://med.emory.edu/departments/radiation-oncology/research-laboratories/deformable-image-registration/downloads-and-reference-data/index.html}{DIR-Lab}&\cite{castillo2009framework} & CT & Lungs & 4D-CT & CT scans & 10  \\ \hline
\href{https://www.kaggle.com/datasets/raddar/nodules-in-chest-xrays-jsrt}{JSRT}&\cite{shiraishi2000development} & X-ray & Chest & 2D images & X-ray images & 247  \\ \hline 

\href{https://radiopaedia.org/articles/ct-pancreas-protocol-1?lang=gb}{Pancreas-CT}& \cite{roth2016data} & CT & Pancreas & 3D volume & Segmentation & Not specified  \\ \hline 

\href{https://www5.cs.fau.de/conrad/data/heart-model/}{4D SSM}& \cite{unberath2015open} & CTA & Heart & 3D volume & Dynamic Mesh & 20  \\ \hline  

SPUM-ACS & \cite{klingenberg2015safety} & US-TTE & Heart & 2D Video & Segmentation & 314 (Acute Myocardial Infarction)  \\ \hline 

\href{https://www.cardiacatlas.org/}{CCT48} & \cite{suinesiaputra2017statistical} & CT & Heart & 3D volume & Segmentation & 48  \\ \hline

\href{https://www.cardiacatlas.org/mitea/}{MITEA}& \cite{zhao2023mitea} & US & Heart & 3D volume & Segmentation & 536  \\ \hline  

\href{https://medicaldecathlon.com/}{MSD liver}& \cite{simpson2019large} & CT & Liver & 3D volume & Segmentation & 131  \\ \hline 

\href{https://echonet.github.io/dynamic/}{EchoNet-Dynamic}& \cite{ouyang2020video}  & ECGI  videos & Heart & 3D Volume & Segmentation, Measurement & 10030 \\ \hline

\href{https://figshare.com/articles/dataset/SkullFix_-_MICCAI_AutoImplant_2020_Challenge_Dataset/14161307}{SkullFix/SkullBreak}& \cite{kodym2021skullbreak}& Binary Mask & Skull & Defective Mask & Reconstructed Mask & 670 \\ \hline

\href{https://figshare.com/articles/dataset/MUG500_Repository/9616319}{MUG500+}& \cite{li2021mug500+} & CT & Skull & 3D Volume &Surface Mesh & 529 \\ \hline

\href{https://github.com/marrlab/SHAPR}{SHAPR}& \cite{waibel2022shapr} & Microscopy & Cell & Single 2D Image & Surface Mesh & 825 \\ \hline   


\href{https://zenodo.org/records/7031924}{Red Blood Cell}& \cite{simionato2021red} & Microscopy & Cell & Single 2D Image &Surface Mesh & 2000 cells \\ \hline

\end{tabular}}

\label{tab:dataset_summary}
\end{table*}

\section{Meta Analysis}

We conducted a meta-analysis to evaluate the performance of template models, statistical shape models, generative models, and implicit models on cardiac MR and cortical MR datasets. The analysis was carried out following clinical principles \cite{julian2019alcohol}, adhering to a study-based statistical approach. Specifically, only studies targeting the same medical objective and employing the same imaging modality were included in the analysis to ensure consistency and comparability.

\begin{figure*}[!ht]
    \centering
    \includegraphics[width=\linewidth]{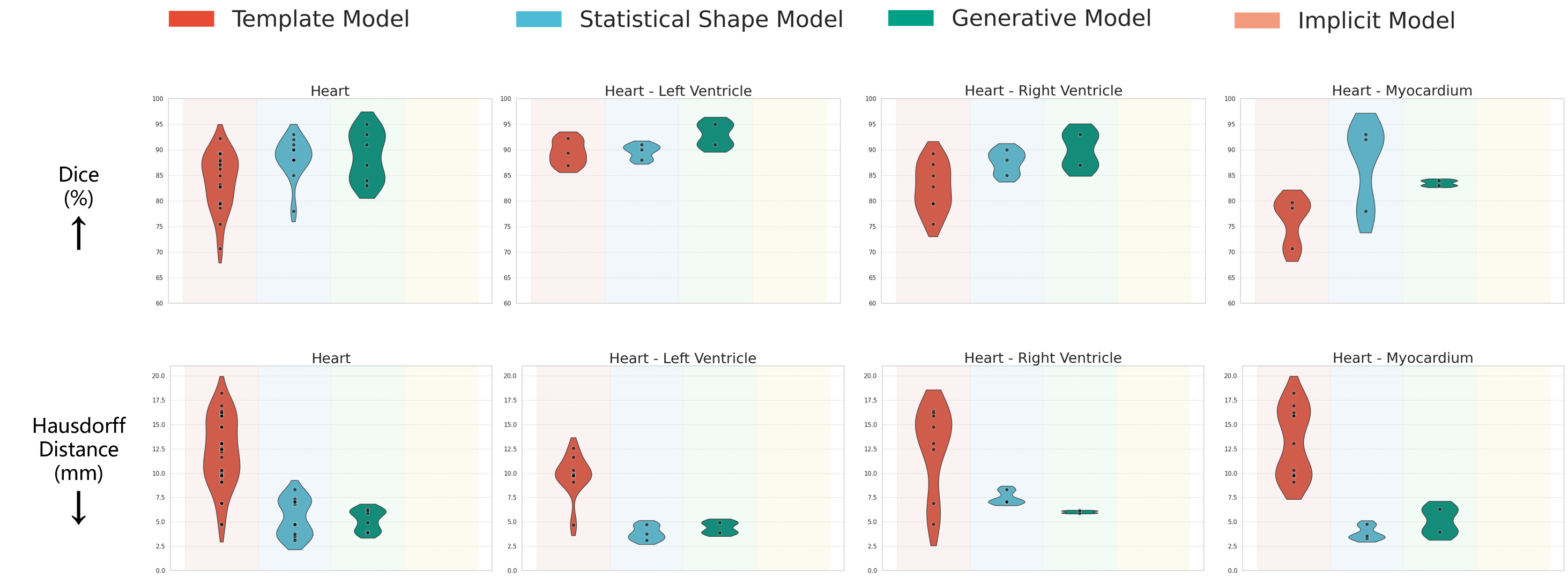}
    \caption{Quantitative results comparison on the Cardiac MR Dataset. The first column on the left displays reconstruction results on the cardiac MR dataset(whole heart, include substructures on the right). The columns to the right show reconstruction results for various heart substructures (left ventricle, right ventricle, and myocardium). Different colours represent different model categories. $\uparrow$ indicates that a higher value is better, while $\downarrow$ indicates that a lower value is better}
    \label{fig:result_cardiac}
\end{figure*}

The performance of template models, statistical shape models, and generative models on the cardiac dataset was analyzed based on Dice similarity coefficient and Hausdorff distance, as shown in Fig.~\ref{fig:result_cardiac}.
For Dice similarity (\%), higher values indicate better shape reconstruction accuracy. The generative model achieved a higher median Dice score than the statistical shape model and the template model.
For Hausdorff distance (mm), lower values indicate better geometric accuracy. The generative model demonstrated the lowest median Hausdorff distance, followed closely by the statistical shape model. The template model showed the highest Hausdorff distance.
Overall, the generative model achieved the best performance among the three, outperforming both the statistical shape model and the template model in terms of shape accuracy and geometric consistency. The statistical shape model performed better than the template model but was still inferior to the generative model.

\begin{figure*}[!ht]
    \centering
    \includegraphics[width=\linewidth]{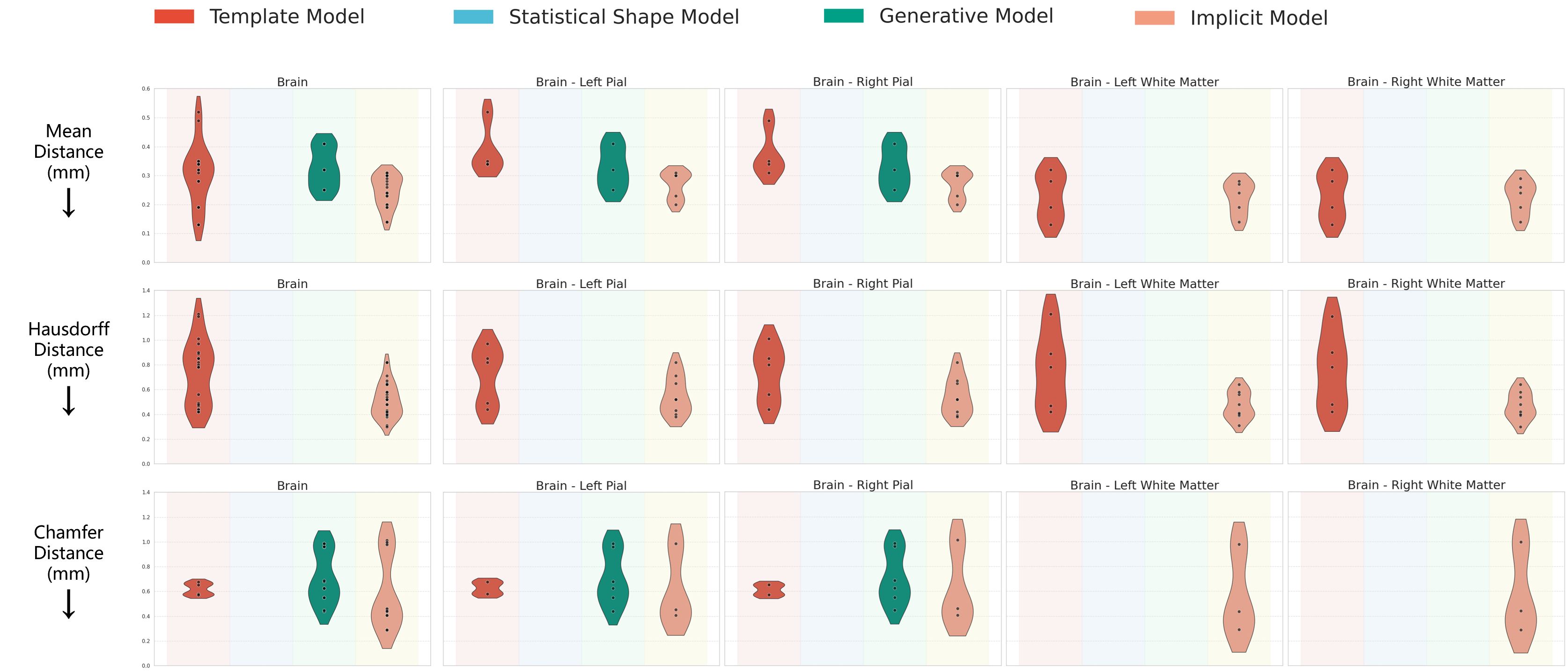}
    \caption{Quantitative results comparison on the brain MR dataset. Our data is from the same modality and the same anatomy, making it comparable across different studies according to clinical standards. The first column on the left displays reconstruction results on the brain MR dataset (whole structure). The columns to the right show reconstruction results for various brain substructures (left and right pial, left and right white matter). Different colours represent different model categories. $\uparrow$ indicates that a higher value is better, while $\downarrow$ indicates that a lower value is better. From the results, it can be seen that there is no absolute superiority or inferiority between the methods, but the overall robustness of each method can be observed in their distribution.}
    \label{fig:result_cortical}
\end{figure*}

The performance of template models, generative models, and implicit models in the cortical dataset was analysed using four evaluation metrics: Average Distance, Chamfer Distance, Hausdorff Distance, and Average Symmetric Surface Distance, as shown in Fig.~\ref{fig:result_cortical}. Lower values across all metrics indicate better reconstruction accuracy.
For average distance, the implicit model demonstrated the lowest values across both pial and white matter surfaces, indicating superior geometric accuracy. The generative model followed closely, achieving better results than the template model. 
For Chamfer distance, the implicit model outperformed the generative model and template model. The implicit model generally had the lowest values, indicating better alignment with ground truth surfaces. 
For Hausdorff distance, the implicit model achieved the lowest values across all anatomical regions, confirming its advantage in maintaining fine structural details. The template model performed moderately well.
For average symmetric surface distance, the implicit model again exhibited lower values than the template model, reinforcing its superiority in shape reconstruction. 
Overall, the implicit model demonstrated the best reconstruction accuracy across all metrics, followed by the generative model and template model.

Overall, the results indicate that the implicit model consistently outperforms the other approaches across multiple datasets and evaluation metrics. On the cortical dataset, the implicit model demonstrated superior geometric accuracy, achieving the lowest error across all four evaluation metrics. The generative model followed, showing better performance than the template model. On the cardiac dataset, where statistical shape models were included and implicit models were excluded, the generative model achieved the highest accuracy, outperforming both statistical shape models and template models.

Considering the results from both the cardiac and cortical datasets, the ranking follows the order: implicit model $>$ generative model $>$ statistical shape model $>$ template model. This ranking is relative rather than absolute, as different methods may still achieve the best performance on specific datasets or under certain conditions. The effectiveness of each approach can vary depending on factors such as anatomical structures, data quality, and task-specific requirements. This trend highlights the increasing effectiveness of data-driven approaches, with implicit and generative models consistently surpassing traditional statistical and template-based methods. The statistical shape model serves as an intermediate solution between generative and template models in certain medical applications, but its performance remains inferior to more advanced deep-learning-based approaches.

\begin{table*}[htbp]
\centering
\caption{Performance comparison of methods on Cardiac datasets.}
\resizebox{\textwidth}{!}{
\begin{tabular}{>{\centering\arraybackslash}p{3cm}>{\centering\arraybackslash}p{7cm}>{\raggedright\arraybackslash}p{3cm}>{\centering\arraybackslash}p{1cm}>{\centering\arraybackslash}p{1cm}>{\centering\arraybackslash}p{1cm}>{\centering\arraybackslash}p{1cm}>{\centering\arraybackslash}p{1cm}}

\toprule
\multirow{2}{*}{Metric} & \multirow{2}{*}{Method} & \multirow{2}{*}{\makecell[c]{Taxonomy }}  & \multicolumn{5}{c}{Cardiac MR} \\ 
\cmidrule(lr){4-8} 
 & & & Myo & LA & LV & RA & RV  \\ 
\midrule

\multirow{10}{*}{\makecell{\rotatebox{0}{Dice (\%) ↑}}} 
& MeshDeformNet \cite{Fanwei2021Deep} & \textbf{T} - Deformation  & 79.71 & 88.13 & 92.23 & 89.24 & 89.24  \\
& HeartFFDNet \cite{Fanwei2021Whole} & \textbf{T} - Deformation  & 70.67 & 83.27 & 86.92 & 82.77 & 82.77  \\
& HeartDeformNet \cite{Fanwei2022Learning} & \textbf{T} - Deformation  & 78.62 & 86.27 & 89.38 & 87.79 & 87.20  \\ 

& FFD-SEG \cite{rueckert1999nonrigid} & \textbf{T} - Registration   &  &  &  &  & 75.47 \\ 
& Dem-SEG \cite{yaniv2018simpleitk} & \textbf{T} - Registration   &  &  &  &  & 79.49   \\ 
& CNN-SEG \cite{upendra2021cnn} & \textbf{T} - Registration   &  &  &  &  & 79.51    \\ 
& Dem-CNN \cite{R2021Motion}& \textbf{T} - Registration   &  &  &  &  & 84.91    \\  \cmidrule(lr){3-8} 

& \cite{attar2019high} & \textbf{S} - Linear SSM   & 92.00 &  & 91.00 &  & 88.00 \\

& \cite{attar20193d} & \textbf{S} - Linear SSM   & 93.00 &  & 90.00 &  & 90.00 \\

& MCSI-Net \cite{xia2022automatic} & \textbf{S} - Linear SSM   & 78.00 & 88.00 & 88.00 & 88.00 & 85.00  \\  \cmidrule(lr){3-8} 

& MV-HybridVNet \cite{Gaggion2023Multi} & \textbf{G} - VAE   & 84.00  &  & 91.00  &  & 87.00   \\  
& \cite{Hao2019Ventricle} & \textbf{G} - Interpolation   & 83.00 &  & 95.00 &  & 93.00 \\ 
\midrule

\multirow{16}{*}{\makecell{\rotatebox{0}{Hausdorff (mm) ↓}}} 
& MeshDeformNet \cite{Fanwei2021Whole} & \textbf{T} - Deformation  & 18.21 & 12.43 & 12.57 & 16.36 & 16.36  \\
& HeartFFDNet \cite{Fanwei2022Learning} & \textbf{T} - Deformation & 15.96 & 10.16 & 8.97 & 12.46 & 12.46  \\
& MeshDeformNet \cite{Fanwei2021Deep} & \textbf{T} - Deformation  & 16.92 & 12.22 & 11.63 & 14.73 & 14.73  \\
& Pixel2mesh \cite{wang2020pixel2mesh} & \textbf{T} - Deformation & & 16.20  & & & 16.20   \\ 
& Mesh Deformation U-Net \cite{M2022Reconstructing} & \textbf{T} - Deformation   &  &  & 4.69 &  & 4.77   \\ 

& CPD \cite{myronenko2010point} & \textbf{T} - Registration  & & 13.05  & & & 13.05   \\ 
& GMMREG \cite{jian2010robust} & \textbf{T} - Registration  & & 15.87  & & & 15.87  \\ 
& FFD \cite{rueckert1999nonrigid} & \textbf{T} - Registration   & $10.31 $ &  & $10.31 $ &  &    \\ 
& dDemons \cite{vercauteren2007non} & \textbf{T} - Registration   & $9.71 $ &  & $9.71 $ &  &    \\ 
& MulViMotion \cite{meng2022mulvimotion} & \textbf{T} - Registration   & $9.86 $ &  & $9.86 $ &  &   \\ 
& MeshMotion \cite{meng2022mesh} & \textbf{T} - Registration   & $9.73 $ &  & $9.73 $ &  &    \\ 
& DeepMesh \cite{Qingjie2023DeepMesh}& \textbf{T} - Registration   & ${9.08 }$ &  & ${9.08 }$ &  &   \\ 
& MR-Net \cite{Xiang2021Shape} & \textbf{T} - Registration  & & {6.89 } & & & {6.89 } \\  \cmidrule(lr){3-8} 

& \cite{attar2019high} & \textbf{S} - Linear SSM   & 3.32 &  & 3.76 &  & 8.32 \\

& \cite{attar20193d} & \textbf{S} - Linear SSM   & 3.55 &  & 3.11 &  & 7.05 \\

& MCSI-Net \cite{xia2022automatic} & \textbf{S} - Linear SSM   & $4.75 $ & $6.82$ & $4.74 $ & $7.34 $ & $7.06 $  \\ \cmidrule(lr){3-8} 

& \cite{Hao2019Ventricle} & \textbf{G} - Interpolation   & $6.30 $ &  & $4.94 $ &  & 5.91    \\ 

& MV-HybridVNet \cite{Gaggion2023Multi}& \textbf{G} - VAE   & $3.96 $ &  & $3.89 $ &  & $6.13 $  \\

\midrule

\multirow{4}{*}{\makecell{\rotatebox{0}{Chamfer (mm) ↓}}} 
& MR-Net \cite{Xiang2021Shape} & \textbf{T} - Deformation  & & {4.39 } & 2.73 & & {4.39 }  \\ 
& \cite{Meng2023Neural} & \textbf{T} - Deformation  & & & 2.32 & &  \\ 

& MCSI-Net \cite{xia2022automatic} & \textbf{S} - Linear SSM   & $1.86 $ & 2.65  & $1.86 $ & 2.79  & $2.27$   \\  \cmidrule(lr){3-8} 

& MV-HybridVNet \cite{Gaggion2023Multi}& \textbf{G} - VAE   & $1.35 $ &  & $1.39 $ &  & $1.76$  \\   

\midrule
\multirow{14}{*}{\makecell{\rotatebox{0}{Mean Dist. (mm) ↓}}} 
& Mesh Deformation U-Net \cite{M2022Reconstructing} & \textbf{T} - Deformation  &  &  & ${0.98}$ &  &  1.25   \\ 
& \cite{Meng2023Neural} & \textbf{T} - Deformation & & & 1.02 & &  \\ 

& FFD \cite{rueckert1999nonrigid} & \textbf{T} - Registration  & 3.02 &  & 3.02 &  &    \\ 
& dDemons \cite{vercauteren2007non} & \textbf{T} - Registration  & 3.20 &  & 3.20 &  &    \\ 
& MulViMotion \cite{meng2022mulvimotion} & \textbf{T} - Registration  & 2.39 &  & 2.39 &  &   \\ 
& MeshMotion \cite{meng2022mesh} & \textbf{T} - Registration   & 1.98 &  & 1.98  &  &    \\ 
& DeepMesh \cite{Qingjie2023DeepMesh}& \textbf{T} - Registration  & 1.66 &  & 1.66  &  &   \\ 
& FFD-SEG \cite{rueckert1999nonrigid} & \textbf{T} - Registration   &  &  &  &  & 4.37   \\ 
& Dem-SEG \cite{yaniv2018simpleitk} & \textbf{T} - Registration   &  &  &  &  & 3.52   \\ 
& CNN-SEG \cite{upendra2021cnn} & \textbf{T} - Registration   &  &  &  &  & 3.34   \\ 
& Dem-CNN \cite{R2021Motion} & \textbf{T} - Registration   &  &  &  &  & 1.08    \\ 
& MR-Net \cite{Xiang2021Shape} & \textbf{T} - Registration  & & & 1.35 & &  \\ \cmidrule(lr){3-8} 

& \cite{attar2019high} & \textbf{S} - Linear SSM   & 1.80 &  & 1.85 &  & 2.02 \\

& \cite{attar20193d} & \textbf{S} - Linear SSM   & 1.82 &  & 1.81 &  & 2.00 \\

\bottomrule
\end{tabular}
}
\end{table*} 

\begin{table*}[htbp]
\centering
\caption{Performance comparison of image to mesh reconstruction methods on Brain MR dataset.}
\resizebox{\textwidth}{!}{

\begin{tabular}{>{\centering\arraybackslash}p{3cm}>{\centering\arraybackslash}p{8cm}>{\raggedright\arraybackslash}p{2.5cm}>{\centering\arraybackslash}p{1cm}>{\centering\arraybackslash}p{1cm}>{\centering\arraybackslash}p{1cm}>{\centering\arraybackslash}p{1cm}}

\toprule
\multirow{2}{*}{Metric} & \multirow{2}{*}{Method} & \multirow{2}{*}{Taxonomy} & \multicolumn{4}{c}{Brain MR}\\ 
\cmidrule(lr){4-7} 
 & & & Left Pial & Right Pial & Left WM & Right WM \\ 
\midrule

\multirow{13}{*}{Mean Dist. (mm) ↓} 
 & TopoFit \cite{Andrew2022TopoFit} & \textbf{T} - Deformation   & & & 0.13 & 0.13   \\ 
 & Voxel2Mesh \cite{wickramasinghe2020voxel2mesh} &  \textbf{T} - Deformation   & 0.34 & 0.31 &    \\
 & \cite{Rickmann2022Joint} & \textbf{T} - Deformation    & 0.35 & 0.35 & 0.32 & 0.32   \\
 & Vox2Cortex \cite{Fabian2022Vox2Cortex} & \textbf{T} - Deformation   & 0.34 & 0.34 & 0.28 & 0.28   \\
 & PialNN \cite{Qiang2021PialNN} & \textbf{T} - Deformation & 0.52 & 0.49 & 0.19 & 0.19  \\ \cmidrule(lr){3-7} 
 & \cite{hu2021point} & \textbf{G} - GAN & 0.32 & 0.32 &  &    \\ 
 
 & \cite{hu2022srt} & \textbf{G} - VAE & 0.25 & 0.25 &  &    \\ 
 & PointOutNet \cite{zhou2019one} & \textbf{G} - VAE & 0.41 & 0.41 &  &    \\  \cmidrule(lr){3-7} 
 
 & SurfNN \cite{Hao2023SurfNN} & \textbf{I} - SDF & 0.23 & 0.23 & 0.14 & 0.14  \\
 & DeepCSR (SDF) \cite{Rodrigo2020DeepCSR} & \textbf{I} - SDF & 0.30 & 0.29 & 0.27 & 0.26  \\

 & CortexODE \cite{Q2022CortexODE} & \textbf{I} - Neural ODE & 0.20 & 0.20 & 0.19 & 0.19  \\
 & CorticalFlow \cite{santa2022corticalflow++} & \textbf{I} - Neural ODE   & 0.30 & 0.30 & 0.24 & 0.24   \\
   
 & DeepCSR (Occ.) \cite{Rodrigo2020DeepCSR} & \textbf{I} - Occupancy & 0.31 & 0.31 & 0.28 & 0.29  \\
 
\midrule

\multirow{12}{*}{Hausdorff (mm) ↓} 
 & Voxel2Mesh \cite{wickramasinghe2020voxel2mesh} & \textbf{T} - Deformation & 0.82 & 0.80 & 1.21 & 1.19    \\
 & Vox2Cortex \cite{Fabian2022Vox2Cortex} & \textbf{T} - Deformation & 0.97 & 1.01 & 0.89 & 0.90   \\
 & \cite{Rickmann2022Joint} & \textbf{T} - Deformation    & 0.85 & 0.85 & 0.78 & 0.78   \\
& TopoFit \cite{Andrew2022TopoFit} & \textbf{T} - Deformation & 0.49 & 0.56 & 0.47 & 0.48  \\
  & PialNN \cite{Qiang2021PialNN} & \textbf{T} - Deformation & 0.44 & 0.44 & 0.42 & 0.42  \\ \cmidrule(lr){3-7} 

& V2C-Flow \cite{Fabian2024Neural} & \textbf{I} - Neural ODE & 0.40 & 0.39 & 0.39 & 0.39  \\
& CorticalFlow \cite{lebrat2021corticalflow} & \textbf{I} -  Neural ODE & 0.52 & 0.52 & 0.48 & 0.48  \\
& CF++ \cite{santa2022corticalflow++} & \textbf{I} -  Neural ODE & 0.38 & 0.38 & 0.40 & 0.40  \\
 & CortexODE \cite{Q2022CortexODE} & \textbf{I} - Neural ODE & 0.43 & 0.42 & 0.41 & 0.42  \\

 & DeepCSR (SDF) \cite{Rodrigo2020DeepCSR} & \textbf{I} - SDF & 0.65 & 0.65 & 0.56 & 0.54  \\
 & Vox2surf \cite{hong2021vox2surf} & \textbf{I} - SDF & 0.82 & 0.82 & 0.64 & 0.64  \\
 & SurfNN \cite{Hao2023SurfNN} & \textbf{I} - SDF & 0.52 & 0.52 & 0.31 & 0.30  \\ 
 & DeepCSR (Occ.) \cite{Rodrigo2020DeepCSR} &  \textbf{I} - Occupancy & 0.71 & 0.67 & 0.58 & 0.58 \\

\midrule
\multirow{8}{*}{Dice (\%) ↑} 
 & TopoFit \cite{Andrew2022TopoFit} & \textbf{T} - Deformation   & & & 97.60 & 97.60   \\ 
 & \cite{Rickmann2022Joint} & \textbf{T} - Deformation   & 87.70 & 87.70 & 90.40 & 90.40   \\
 & Vox2Cortex \cite{Fabian2022Vox2Cortex} &  \textbf{T} - Deformation   & 69.10 & 69.10 & 74.00 & 74.00   \\
 & Voxel2Mesh \cite{wickramasinghe2020voxel2mesh} & \textbf{T} - Deformation & 89.60 & 88.80 & 85.00 & 84.20  \\ \cmidrule(lr){3-7} 
 & CorticalFlow \cite{lebrat2021corticalflow} & \textbf{I} - Neural ODE & 97.70 & 97.60 & 96.20 & 96.20  \\ 
 & NMF \cite{gupta2020neural} & \textbf{I} - Neural ODE & 95.30 & 94.60 & 92.80 & 92.70  \\ 
  
 & {DeepCSR} \cite{Rodrigo2020DeepCSR} & \textbf{I} - SDF & 98.10 & 98.10 & 96.30 & 96.40  \\ 

\midrule

\multirow{10}{*}{Chamfer (mm) ↓} 
 & Voxel2Mesh \cite{wickramasinghe2020voxel2mesh} & \textbf{T} - Deformation   & 0.58 & 0.57 & &   \\
 & PialNN \cite{Qiang2021PialNN} & \textbf{T} -  Deformation & 0.68 & 0.65 & &   \\ \cmidrule(lr){3-7} 
 
 & \cite{hu2021point} & \textbf{G} - GAN &  0.96 & 0.96 &  &    \\ 

 & \cite{hu20233} & \textbf{G} - GAN & 0.44 & 0.45 &  &    \\  
 & \cite{hu2022srt} & \textbf{G} - VAE &  0.68 & 0.69 &  &    \\ 
 & TopNet \cite{tchapmi2019topnet} & \textbf{G} - VAE & 0.63 & 0.63 &  &    \\ 
 & FoldingNet \cite{yang2018foldingnet} & \textbf{G} - VAE & 0.99 & 0.99 &  &    \\ 
 & PointOutNet \cite{zhou2019one} & \textbf{G} - VAE & 0.55 & 0.55 &  &    \\  \cmidrule(lr){3-7} 

 & CortexODE \cite{Q2022CortexODE} & \textbf{I} - Neural ODE & 0.45 & 0.46 & 0.44 & 0.44   \\
 & DeepCSR \cite{Rodrigo2020DeepCSR} & \textbf{I} - SDF  & 0.99 & 1.01 & 0.98 & 1.00   \\
 & SurfNN \cite{Hao2023SurfNN} & \textbf{I} - SDF & 0.41 & 0.41 & 0.29 & 0.29   \\

\midrule

\multirow{10}{*}{ASSD (mm) ↓} 
 & TopoFit \cite{Andrew2022TopoFit} & \textbf{T} - Deformation & 0.22 & 0.25 & 0.21 & 0.21  \\
 & Voxel2Mesh \cite{wickramasinghe2020voxel2mesh} & \textbf{T} -  Deformation  & 0.49 & 0.48 & 0.53 & 0.53  \\ 

 & Vox2Cortex \cite{Fabian2022Vox2Cortex} & \textbf{T} -  Deformation    & 0.38 & 0.38 & 0.40 & 0.40   \\

 & PialNN \cite{Qiang2021PialNN} & \textbf{T} - Deformation & 0.21 & 0.21 & 0.19 & 0.19  \\\cmidrule(lr){3-7}

  & V2C-Flow \cite{Fabian2024Neural} & \textbf{I} - Neural ODE & 0.18 & 0.17 & 0.18 & 0.18  \\
& CorticalFlow \cite{lebrat2021corticalflow} & \textbf{I} - Neural ODE & 0.22 & 0.22 & 0.21 & 0.21  \\
 & CF++ \cite{santa2022corticalflow++} & \textbf{I} - Neural ODE & 0.17 & 0.17 & 0.18 & 0.18  \\
 & DeepCSR \cite{Rodrigo2020DeepCSR} & \textbf{I} - Neural ODE & 0.45 & 0.42 & 0.42 & 0.42  \\
 & CortexODE \cite{Q2022CortexODE} & \textbf{I} - Neural ODE & 0.18 & 0.19 & 0.17 & 0.17  \\

\bottomrule
\end{tabular}
}
\end{table*}

\section{Discussion}

\begin{figure*}[!ht]
    \centering
    \includegraphics[width=\linewidth]{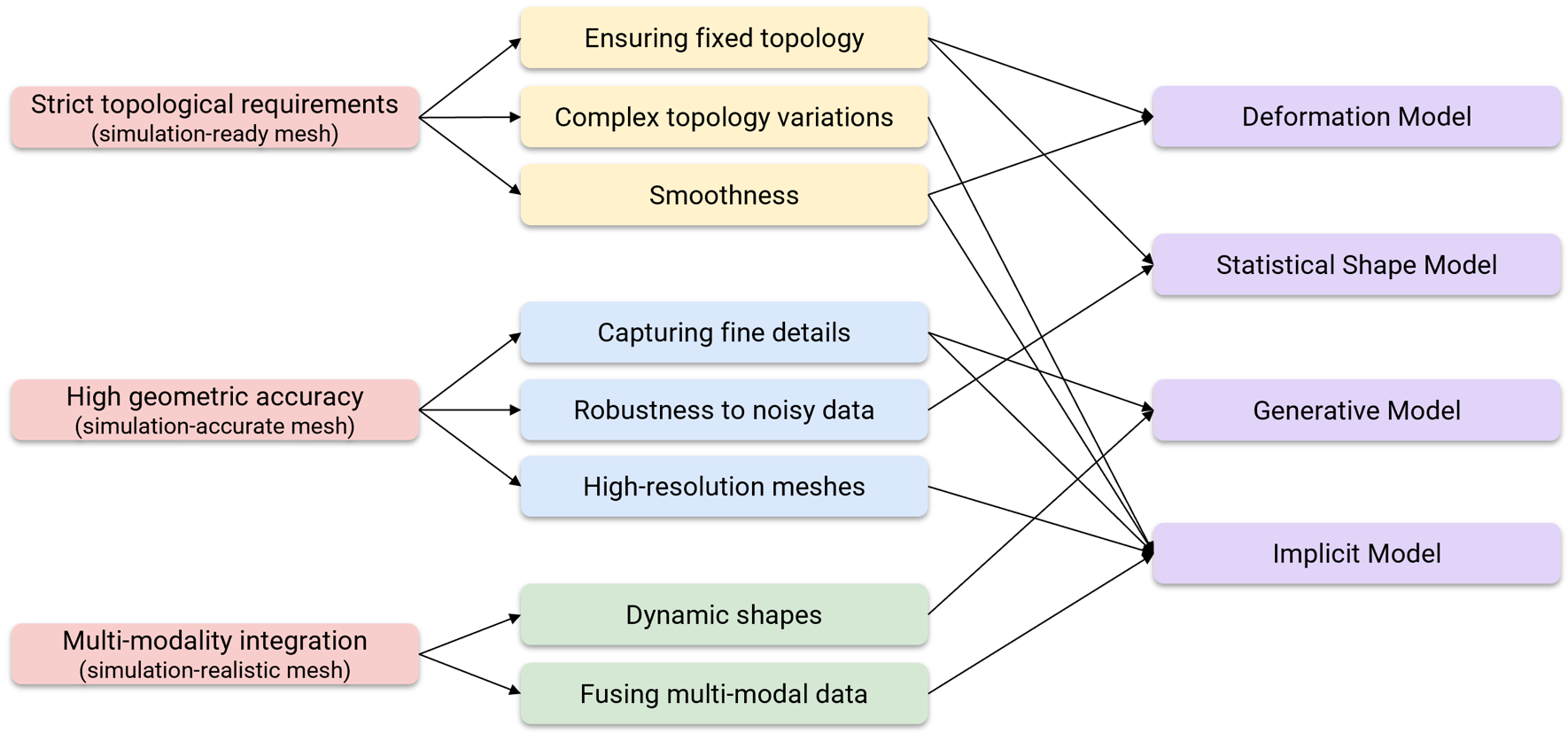}
    \caption{Challenges of Medical Image-to-Mesh Reconstruction.}
    \label{fig:challenges} 
\end{figure*}

Figure \ref{fig:challenges} systematically illustrates how different methods address key challenges in the process of medical image-to-mesh reconstruction, highlighting the suitability and strengths of various models for complex tasks. The entire framework is divided into three levels, linking challenges and requirements to core objectives and ultimately to specific method categories, creating a clear path that connects needs with solutions.  

Medical image-to-mesh reconstruction faces multiple challenges, one of which is the strict topological requirements. This often involves generating meshes with fixed topology, ensuring stability and the absence of topological errors during complex anatomical simulations. Additionally, achieving accurate simulations often necessitates high geometric accuracy, ensuring that meshes reflect anatomical details with precision, providing a solid foundation for calculations and simulations. Another critical challenge is multi-modality integration, which seeks to fuse data from different imaging modalities (such as CT, MRI, and ultrasound) to produce dynamic meshes that are more realistic and anatomically consistent.  

To address these challenges, different methods are designed to achieve various core objectives. For instance, methods aimed at meeting topological requirements focus on ensuring fixed topology or handling complex topological variations to accommodate anatomical differences across individuals. Smoothness is essential for generating continuous mesh surfaces, preventing discontinuities or surface irregularities. In pursuit of high geometric accuracy, methods must be capable of capturing fine details to preserve subtle anatomical features, while also improving robustness to noisy data to mitigate the effects of imaging artifacts or missing data. Generating high-resolution meshes is another critical goal, especially for reconstructing complex anatomical regions. In the context of multi-modality imaging, models must not only reconstruct dynamic shapes but also possess the ability to fuse multi-modal data to ensure consistent mesh reconstruction under different imaging conditions.  

Four primary methods play distinct roles in achieving these objectives, each offering unique strengths and applicability. Deformation models progressively adjust an initial template mesh to fit the target shape, making them particularly suitable for tasks with strict topological constraints, where fixed or predictable topology is essential. Statistical shape models rely on large anatomical datasets and employ statistical techniques to model shape variations, ensuring smooth surfaces and maintaining topological consistency. In contrast, generative models synthesize meshes directly from input images, capturing intricate details while maintaining robustness to noise and missing data. Implicit models, on the other hand, represent mesh surfaces through learned implicit functions, enabling the generation of high-resolution and complex topological meshes. These models excel at reconstructing dynamic shapes and integrating multi-modal data, making them ideal for highly intricate anatomical structures.  

The complementary nature of different methods in addressing various challenges and objectives allows for the selection of the most suitable reconstruction techniques based on specific applications. Combining multiple approaches can further enhance the accuracy and stability of the final mesh, providing critical support for medical research and simulations.

\begin{table*}[h!]
\centering
\caption{Comparison of different models for medical image-to-mesh reconstruction.}
\resizebox{\textwidth}{!}{%
\begin{tabular}{l|p{6cm}|p{6cm}|p{8cm}} 
\hline \hline

\textbf{Model} & \textbf{Advantages} & \textbf{Disadvantages} & \textbf{Applicable Scenarios} \\ \hline \hline
Deformation Model & \makecell[l]{Controlled topology \\ Easy to optimize} & \makecell[l]{Strong dependence on templates \\ Not suitable for complex topology} & \makecell[l]{Fixed topology shape reconstruction \\ (Example: generating normal cardiac meshes)} \\ \hline
Statistical Shape Model & \makecell[l]{Strong shape priors \\ Robust to noise} & \makecell[l]{Strong dependence on training data \\ Limited detail representation} & \makecell[l]{Scenarios requiring statistical regularities \\  (Example: reconstruction of typical cardiac anatomy)} \\ \hline
Generative Model & \makecell[l]{Adaptable to diverse shapes \\ Detailed representation } & \makecell[l]{Complex training \\ High data requirement} & \makecell[l]{Personalized mesh generation \\ (Example: pathological or abnormal shapes)} \\ \hline
Implicit Model & \makecell[l]{Flexible topology \\ High resolution} & \makecell[l]{Complex training and optimization} & \makecell[l]{Irregular surfaces or complex topologies \\ (Example: modelling fine details around cardiac valves)} \\ 

\hline

\end{tabular}}
\label{tab:comparison_models}
\end{table*}

Table \ref{tab:comparison_models} presents a comprehensive comparison of different models used in medical image-to-mesh reconstruction, outlining their respective advantages, disadvantages, and applicable scenarios. This structured overview provides valuable insights into the strengths and limitations of each approach, guiding the selection of appropriate methods based on specific reconstruction requirements.  

The deformation model stands out for its ability to control topology, making it easy to optimize. This model is particularly well-suited for tasks that require fixed topology shape reconstruction, such as generating normal cardiac meshes. However, its strong dependence on templates and limited adaptability to complex topologies can restrict its application in more intricate anatomical structures. While it provides stable and controlled results, its inability to handle highly irregular surfaces may necessitate complementary methods for complex reconstructions.  

The statistical shape model leverages strong shape priors, offering robustness to noise and ensuring consistency across different datasets. This makes it particularly useful for scenarios that demand statistical regularities, such as reconstructing typical cardiac anatomy. Nevertheless, the model's reliance on extensive training data and its limited ability to represent fine details pose notable challenges. Despite these limitations, its ability to enforce anatomical constraints makes it an essential tool for applications that prioritize structural consistency over intricate detail.  

Generative models provide a high degree of adaptability, capable of representing diverse shapes with detailed accuracy. This model excels in personalized mesh generation, particularly for pathological or abnormal shapes where standard templates may not suffice. The ability to synthesize complex geometries makes it highly valuable in cases requiring detailed, patient-specific reconstructions. However, generative models are often associated with complex training processes and significant data requirements, which can limit their accessibility and scalability in resource-constrained environments.  

Implicit models offer unmatched flexibility in topology and enable high-resolution mesh reconstruction. They are particularly advantageous in handling irregular surfaces or complex topologies, such as modelling fine anatomical details around cardiac valves. Despite their potential to generate intricate and detailed meshes, implicit models come with the challenge of complex training and optimization, which may increase computational costs and prolong development timelines. Nevertheless, their ability to capture fine-grained structures makes them indispensable in applications requiring precision and topological adaptability.  

While deformation and statistical shape models offer reliability and ease of use for standard reconstructions, generative and implicit models provide the flexibility and detail required for more challenging and patient-specific scenarios. This comparative analysis highlights the complementary nature of these models, suggesting that hybrid approaches may often yield the most robust and comprehensive reconstruction outcomes.

\section{Future Directions}

Looking back at the reviewed methods, the development trend of medical image-to-mesh reconstruction has shown a clear technological evolution path: from statistical shape models to template models, then to generative models, and finally to implicit models. Among these, implicit models have demonstrated significant advantages in both fidelity and regularization. Currently, the implicit models used for medical image reconstruction primarily include SDF and NeuralODE approaches.

Meanwhile, in the fields of computer vision and computer graphics, several surface reconstruction algorithms that outperform SDF \cite{yariv2021volume} have emerged. High-fidelity surface reconstruction using neural implicit reconstruction \cite{wang2021neus, wang2023neus2, liu2023nero, yariv2023bakedsdf} has become a hot research topic. Compared to SDF, Neural ODE and NeRF, new representation methods (such as Hash coding \cite{muller2022instant, li2023neuralangelo}, Gaussian splatting \cite{kerbl20233d, gao2025relightable, guedon2024sugar, huang20242d, yu2024gaussian, chen2024pgsr, dai2024high, wang2024gaussurf, tang2023dreamgaussian}) provide better texture preservation.

Gaussian splatting technique has demonstrated breakthrough advantages in surface reconstruction. This approach achieves superior computational efficiency with training time reduced by more than 10 times \cite{lyu20243dgsr} compared to existing neural implicit reconstruction methods. Through explicit 3D Gaussian point representation and surface alignment strategies, complex surface details can be accurately reconstructed. The flexible optimization framework allows simultaneous optimization of geometric structure and appearance features, achieving high-fidelity reconstruction. Furthermore, its use of sparse 3D Gaussian points representation significantly reduces memory overhead, while its end-to-end reconstruction capability avoids post-processing steps like depth fusion in traditional methods, thereby reducing error accumulation.
These advantages make Gaussian splatting show tremendous potential in medical image reconstruction. In the future, combining this efficient neural surface reconstruction technology with the specific requirements of medical image processing is expected to bring major breakthroughs in medical image reconstruction quality. Particularly in clinical application scenarios that require a balance of reconstruction accuracy, computational efficiency, and real-time performance, the advantages of Gaussian splatting will become especially prominent.

\section{Conclusion}

This comprehensive survey has systematically examined deep learning-based medical image-to-mesh reconstruction methods, categorizing them into four main approaches: statistical shape models, template models, generative models, and implicit models. Furthermore, we have refined this classification into twelve subcategories based on their processing pipelines and feature representations, offering a structured and in-depth taxonomy of existing methodologies. Each approach presents distinct advantages and trade-offs, making them suitable for different medical imaging applications.  

Deformation-based template models excel in preserving topological consistency and are well-suited for tasks requiring fixed mesh topology, such as cardiac mesh reconstruction. However, their dependency on predefined templates may limit their adaptability to complex anatomical variations. Statistical shape models leverage prior anatomical knowledge and demonstrate robustness in handling noisy data, making them particularly useful for standard anatomical reconstructions where statistical regularities are beneficial. Generative models introduce unprecedented flexibility in mesh synthesis, enabling personalized reconstructions that can accommodate pathological variations beyond standard anatomical templates. Implicit models offer superior adaptability in topology and allow for high-resolution reconstructions, making them particularly effective for complex anatomical structures with intricate geometric details.  

Beyond method classification, this survey has also provided a structured summary of loss functions and evaluation metrics commonly used in image-to-mesh reconstruction. Additionally, we have conducted a meta-analysis of experimental results across different anatomical structures, offering insights into the relative performance of various method categories. Furthermore, we have systematically reviewed and curated publicly available datasets relevant to this field, covering multiple anatomical structures and imaging modalities. This dataset collection serves as a valuable resource for researchers working on medical image-to-mesh reconstruction and in-silico trial simulations.  

Looking ahead, new foundational technologies are reshaping medical image-to-mesh reconstruction, enabling higher fidelity, efficiency, and adaptability in diverse clinical applications.
Emerging representation techniques such as Gaussian splatting offer new possibilities for high-fidelity mesh reconstruction with improved efficiency and scalability. Additionally, generative models like conditional diffusion models have shown significant promise in generating anatomically accurate meshes by incorporating prior constraints and uncertainty quantification. Another key direction is multi-modal fusion, where integrating diverse imaging modalities with deep learning-based reconstruction pipelines can enhance robustness and anatomical consistency. The incorporation of large language models into multi-modal frameworks also presents an exciting frontier, enabling structured prior knowledge integration, automated annotation, and context-aware reconstruction guidance. 
Furthermore, volume mesh reconstruction is gaining attention, as it is essential for biomechanical simulations and computational modelling, which require volumetric representations beyond surface meshes.
These advanced methodologies hold great potential for applications in surgical planning, disease diagnosis, and personalized treatment strategies, paving the way for the next generation of medical image-to-mesh reconstruction.

As deep learning architectures and computational capabilities continue to advance, the field of medical image-to-mesh reconstruction is evolving rapidly. The synthesis of different methodological approaches, combined with the increasing availability of high-quality medical imaging datasets, is expected to drive future innovations. With these advancements, we anticipate that medical image-to-mesh reconstruction will become an integral component of clinical workflows, bridging the gap between medical imaging and practical healthcare applications and ultimately contributing to improved patient care and clinical outcomes.

\section{Acknowledgements}

AFF acknowledges support from the Royal Academy of Engineering under the RAEng Chair in Emerging Technologies (INSILEX CiET1919\/19), ERC Advanced Grant – UKRI Frontier Research Guarantee (INSILICO EP\/Y030494/1), the UK Centre of Excellence on in-silico Regulatory Science and Innovation (UK CEiRSI) (10139527), the National Institute for Health and Care Research (NIHR) Manchester Biomedical Research Centre (BRC) (NIHR203308), the BHF Manchester Centre of Research Excellence (RE\/24\/130017), and the CRUK RadNet Manchester (C1994\/A28701).

\bibliographystyle{plainnat}
\bibliography{refs}
\end{document}